\documentclass[11pt]{report}
\usepackage[utf8]{inputenc}
\usepackage{graphicx}  
\usepackage[letterpaper, left=1in, right=1in, top=1in, bottom=1in]{geometry}
\usepackage[doublespacing]{setspace}  
\usepackage{times}  
\usepackage[explicit]{titlesec}  
\usepackage[titles]{tocloft}  
\usepackage[backend=bibtex, style=authoryear, maxbibnames=99]{biblatex}  
\renewbibmacro{in:}{%
  \ifboolexpr{%
     test {\ifentrytype{article}}%
     or
     test {\ifentrytype{inproceedings}}%
  }{}{\printtext{\bibstring{in}\intitlepunct}}%
}

\makeatletter
\newcommand\thefontsize{The current font size is: \f@size pt}
\makeatother

\usepackage[bookmarks=true, hidelinks, breaklinks]{hyperref}
\usepackage[page]{appendix}  
\usepackage{rotating} 
\usepackage{pdfpages} 

\usepackage{amsmath,amssymb}
\usepackage{subfig}
\usepackage{tikz}
\usetikzlibrary{shapes.geometric, arrows}
\tikzstyle{arrow} = [thick,->,>=stealth]

\usepackage{todonotes}

\renewcommand{\cite}[1]{\parencite{#1}}
\newcommand{\citep}[1]{\parencite{#1}}
\newcommand{\citet}[1]{\textcite{#1}}
\newcommand{\blankpage}{
}

\bibliography{CA,references,cognition,hyperbrain}
\AtEveryBibitem{\clearfield{issn}}
\AtEveryBibitem{\clearlist{issn}}

\AtEveryBibitem{\clearfield{isbn}}
\AtEveryBibitem{\clearlist{isbn}}

\AtEveryBibitem{\clearfield{city}}
\AtEveryBibitem{\clearlist{city}}

\AtEveryBibitem{\clearfield{place}}
\AtEveryBibitem{\clearlist{place}}

\AtEveryBibitem{\clearfield{eprint}}
\AtEveryBibitem{\clearlist{eprint}}

\AtEveryBibitem{\clearfield{language}}
\AtEveryBibitem{\clearlist{language}}

\AtEveryBibitem{\clearfield{url}}
\AtEveryBibitem{\clearlist{url}}

\AtEveryBibitem{%
  \ifentrytype{online}
    {}
    {\clearfield{urlyear}\clearfield{urlmonth}\clearfield{urlday}}}

\AtEveryBibitem{%
  \ifentrytype{article}
    {\clearfield{publisher}}
    {}}

\AtEveryBibitem{%
  \ifentrytype{inproceedings}
    {\clearfield{publisher}}
    {}}
    
\AtEveryBibitem{%
  \ifentrytype{incollection}
    {\clearfield{publisher}}
    {}}

\title{Topic Modeling the Reading and Writing Behavior of Information Foragers}
\author{Jaimie Murdock}
\date{\today}

\usepackage{bookmark}
\begin{document}


\newcommand{\thesisTitle}{Topic Modeling the Reading and Writing Behavior of Information Foragers}
\newcommand{\yourName}{Jaimie Murdock}
\newcommand{\yourDepartment}{School of Informatics, Computing, and Engineering and the Cognitive Science Program}
\newcommand{\yourSchool}{Indiana University}
\newcommand{\yourMonth}{June}
\newcommand{\yourYear}{2019}


\begin{titlepage}
\newgeometry{top=2in}
\begin{center}

{\large \MakeUppercase{\thesisTitle}}\\
\vspace{7\baselineskip}
\yourName\\
\vfill
Submitted to the faculty of the University Graduate School\\
in partial fulfillment of the requirements\\
for the degree\\
Doctor of Philosophy\\
in the \yourDepartment,\\
\yourSchool\\
\yourMonth{} \yourYear{}

\end{center}
\restoregeometry
\end{titlepage}

\pagenumbering{roman}
\setcounter{page}{2} 

\newcommand{\committeeChairpersonTypedName}{Colin Allen}
\newcommand{\committeeChairpersonPostNominalInitials}{Ph.D.}

\newcommand{\committeeMemberTwoTypedName}{Stasa Milojevic}
\newcommand{\committeeMemberTwoPostNominalInitials}{Ph.D.}

\newcommand{\committeeMemberThreeTypedName}{Peter Todd}
\newcommand{\committeeMemberThreePostNominalInitials}{Ph.D.}

\newcommand{\committeeMemberFourTypedName}{Michael Jones}
\newcommand{\committeeMemberFourPostNominalInitials}{Ph.D.}


\newcommand{\myRule}{\rule{0.5\textwidth}{0.4pt}}

\newcommand{\approvalDay}{30}
\newcommand{\approvalMonth}{05}
\newcommand{\approvalYear}{2019}


\newgeometry{left=1in}

\begin{center}
 
Accepted by the Graduate Faculty, Indiana University, in partial fulfillment of the requirements for the degree of Doctor of Philosophy.

\end{center}

\vspace{2\baselineskip}

\ifdefined\committeeMemberFourTypedName
Doctoral Committee\\

\null\hfill \myRule\\
\null\hfill \committeeChairpersonTypedName, \committeeChairpersonPostNominalInitials\\
\null\hfill \myRule\\
\null\hfill \committeeMemberTwoTypedName, \committeeMemberTwoPostNominalInitials\\
\null\hfill \myRule\\
\null\hfill \committeeMemberThreeTypedName, \committeeMemberThreePostNominalInitials\\
\null\hfill \myRule\\
\null\hfill \committeeMemberFourTypedName, \committeeMemberFourPostNominalInitials\\

\ifdefined\committeeMemberFiveTypedName
\null\hfill \myRule\\
\null\hfill \committeeMemberFiveTypedName, \committeeMemberFivePostNominalInitials\\
\fi

\fi
\vfill
Date of Defense: \approvalMonth/\approvalDay/\approvalYear
\restoregeometry

\clearpage
\begin{center}

\vspace*{\fill}
Copyright \copyright{} \yourYear{}\\
\yourName
\vspace*{\fill}

\end{center}
\clearpage
\newpage
\newcommand{\yourDedication}{In memory of Granddad, John Thomas Murdock.}


\begin{center}

\vspace*{\fill}
\yourDedication\\
\vspace*{\fill}

\end{center}
\newpage
\phantomsection
\addcontentsline{toc}{chapter}{Acknowledgements}

\begin{centering}
\textbf{ACKNOWLEDGEMENTS}\\
\vspace{\baselineskip}
\end{centering}

\noindent This dissertation would not be possible without the contributions of many individuals. While the overall authorship of this dissertation lists only my name, I acknowledge all the contributions throughout the work by using the inclusive ``we.'' Digital scholarship rests on the countless efforts of many archivists, digitizers, programmers, and publishers that often go unnamed.

This project is tremendously indebted to Colin Allen, my Cognitive Science co-chair and mentor. I began working with Prof. Allen my first semester of undergraduate studies at Indiana University---12 years ago! Somehow his teaching assistant, Ronaldo Vigo, discovered I knew how to program and referred me to interview with him in October 2007. I began work on the Indiana Philosophy Ontology (InPhO) Project, mostly writing SQL and PHP to advance the project. Eventually, I became the lead developer and redesigned the whole project in Python. In 2010, I became a full-time research associate, giving my first taste of a pure research job. On a personal note, his continuing patience with me throughout my late teens and twenties has been monumental, and I'm glad to be finishing this dissertation under his direction.

The HathiTrust Research Center (HTRC) has been a huge help with this dissertation. In 2012, Prof. Allen received a Digging into Data grant that changed our research trajectory dramatically towards digital history, as establishing us as the first external HTRC collaborators. Later, I would become an internal collaborator as the HTRC sponsored me on a graduate fellowship and deployed much of the code used in this dissertation to their production environments. I want to especially thank Beth Plale for authorizing my fellowship, Yu Ma for supervising me during my fellowship year, and my many co-workers and collaborators.

Sta\v{s}a Milojevi\'c, my Informatics co-chair, has my never-ending appreciation for taking me on in the last two years and teaching me how to focus a disparate collection of work. She helped me realize when to press pause and that the best dissertation was a finished dissertation. Peter Todd and Mike Jones both gave extensive comments and support to this project over the past several years and I'm incredibly glad for their participation on my committee, especially in pushing on the cognitive aspects of this work. Simon DeDeo developed the information theoretic measures throughout the project, both as a co-author on the initial Darwin studies. Both he and Larry Yaeger are responsible for my understanding of information theory. Franco Pestilli pushed the neuroscience citation study.

The staff in Informatics and Cognitive Science were both instrumental in actually seeing me to the end, especially Susan Palmer, Terri King, Linda Hostetter, Beverly Diekhoff, and Cheryl Engel. The graduate directors also incredibly helpful in navigating a dual PhD program: Marty Siegel, Selma Sabanovic, Peter Todd, and Larry Moss. At times, it felt like I was the first to tread these waters, especially as faculty turnover reshaped my committee. Without their work behind the scenes, it would've been impossible to manage.

My wife, Emily Byers, reviewed and edited chapters \ref{chapter:introduction}, \ref{chapter:readings}, and \ref{chapter:writings}. Her support reaffirmed how much this dissertation mattered to me. When we first met, I was unsure of finishing the Ph.D., and she really had the patience to let me work through that. I also can't thank her enough for the past year: taking care of our son while I wrote, plus moving and then welcoming her mother-in-law to Albuquerque, all while working on her own graduate work, starting a new job, and recovering from childbirth---she's absolutely incredible! 

Last year, our son Javier was born. His steadiness as we moved across the country was astonishing. We are so lucky to have you, Javi!

My mom, Bonnie Miller Murdock, did so much to make this dissertation a reality. When we had Javier, she moved from Kentucky to New Mexico to help with childcare as I finished writing. It was a huge risk, but seeing her find happiness with family has been a joy.

Learning to program would not have been possible without the help of my dad, Larry Murdock. As a research programmer for University of Wisconsin's soil sciences department, he had access to the birth of the World Wide Web. His friend John Faust gave me my first computer, an Amiga 1000. Mom and dad forbade video games at first, so I played around with paint programs and system utilities. Dad would teach me how to do various tasks like formatting disks, and I would spend days just formatting disks. When we got a Windows machine, he taught me how to use the command line. I learned about system internals---my first memorable debugging experience was when I renamed the \texttt{C:{\textbackslash}Windows} directory. He taught me HTML, then got me hooked on databases and ColdFusion, then applications development with C\#. I was extremely privileged to get exposure to enterprise architecture, web development, and to try my hand at production code before even starting undergrad. Computational thinking started with him.

This dissertation is dedicated to my granddad, John Thomas Murdock, who was the first Ph.D. in the family. It wasn't until I entered graduate school that we started really connecting to each other and it was great to discover this whole academic-professional aspect that I never knew. At the end of my first year in the joint program, I travelled with him to Indonesia and saw the agricultural school he helped found and walked the campus he designed. He was known there for his humility: he kept a hacksaw at his desk so he could shave ``Prof. Dr.'' and ``Ph.D.'' off his name plate every time his titles were forced on him, so he would remain approachable. He taught me the land grant creed of ``teaching, research, and service'', which guides me now. I always smile at his last card: ``Though I do not fully understand all that you are doing in your chosen field of study, the recognition of your work by your peers makes it evident that you are doing excellent work.`` He followed it with a reminder about work-life balance and finding success there as well.

My siblings have been a constant source of support and enthusiasm: thank you, Jesse, Justin, and Molly!

While working on this project, I was employed by the arXiv at Cornell University. Thanks to my supervisors, Martin Lessmeister and Erick Peirson, for extraordinary empathy as I navigated this process alongside a multitude of life changes.

I'm grateful to the help of Endrina Tay, Jack Robertson, and J. Jefferson Looney at the International Center for Jefferson Studies (ICJS) at Monticello, Charlottesville, Virginia for their assistance curating the collections used in chapter~\ref{chapter:jefferson}. Remarkably, my ICJS Fellowship started on the 200th anniversary of Jefferson's sale of his personal library to the Library of Congress. I'm also thankful for the labor of Aish Thamba and Ram Iyer for curating a HathiTrust collection of Jefferson's Retirement Library. Tom Murphy assisted with corpus curation for the first study of Darwin's reading notebooks and Justin Stamets and Ram Iyer assisted with curation of Darwin's Library.

In addition to Colin Allen, Cameron Buckner and Mathias Niepert were my first mentors at the InPhO Project. I am fortunate to have had so many opportunity through their project. Robert Rose, Doori Rose, and Jun Otsuka did incredible work on the Digging into Data grant, and it's their initial curiosity and implementation of topic modeling that led to this work. I was fortunate to have the opportunity to pass on their good will by mentoring undergraduate and graduate students through InPhO, as acknowledged in my CV.

\begin{centering}
\vspace{\baselineskip}
\textbf{PRIOR WORK}\\
\end{centering}

\noindent This dissertation builds upon the following previously-published studies:

\nocite{murdock_quantitative_2018,murdock_exploration_2017}
\printbibliography[keyword=ackpub,heading=none]{}

\noindent Thanks to each of the conference venues, journal editors, anonymous reviewers, and especially the later-unmasked reviewers, for their dialog on this work. It has greatly improved the quality of my research.  \newpage

\begin{center}

\yourName\\
\MakeUppercase{\thesisTitle}

\end{center}

\vspace{1.5\baselineskip}

How do individuals create a knowledge base over a lifetime? Charles Darwin left detailed records of every book he read from The Voyage of the Beagle to just after publication of The Origin of Species. Additionally, he left copies of his drafts before publication. I use these records to build a case study of how reading and writing interact to create conceptual novelties, such as the theory of natural selection and modification by descent. The model is extended to cover entire disciplines by bootstrapping reading and writing histories from bibliographies in scientific publications, scaling the model to address the question of how we move from an individual psychology to society?

There are two central components from cognitive science that impact the proposed models. The first is bounded cognition. People have limited attention, and that attention is further limited by an individual’s information processing ability. Information foraging is a framework for managing the trade-off between exploration of new information and exploitation of existing knowledge when searching for information. Most existing work on information foraging and bounded cognition examine short-term information foraging problems, such as formulating web search queries in a laboratory setting with a known information goal. Through the case study of Charles Darwin, we use real-world datasets to explore this problem at a timescale of decades with unknown information goals.

The base of the reading model is topic modeling with Latent Dirichlet Allocation (LDA). This method reduces the dimensionality of text by reducing each document to a topic distribution, where each topic is defined as a probability distribution over the words in the collection. With these probability distributions, we are able to apply information theoretic measures to calculate the divergence between texts. These divergences characterize a particular reading decision as exploiting the topics exposed by previously read texts or exploring new topics. I train these topic models not on the records, but identify each volume in the Hathi Trust Digital Library and train the topic model on the full text of the books.

While Darwin’s reading notebooks and manuscript drafts provide relatively precise information on reading and writing behaviors at a day-level granularity, that type of data is rare. I explore three extensions of the models, dealing with progressively more “fuzzy” data. First, I look at the contents of Darwin’s Library at the time of his death to infer readings 1860-1882. These readings are used to provide a preliminary analysis of his work on The Descent of Man and the latter editions of the Origin of Species. Then, I look at another historical figure: Thomas Jefferson, whose working library formed the basis of the Library of Congress. We examine the bibliography of his retirement library and tie it into his correspondence to find possible evidence for when certain volumes were read. Finally, I scale the model up to the discipline of neuroscience. I extract citation graphs from the Web of Science to infer reading histories for neuroscientists based on the articles they cited. I use the text of the abstracts of these articles to perform a similar analysis to the Darwin case study on readings and writings. These extensions of the model highlight the potential to work with less precise data and illuminate future problems.

Throughout the work, I emphasize the notion of multiple realizability and interpretive pluralism. Each model is itself a population of models, and while simpler term-frequency-based models may show many of the same effects as the topic models, an argument is made for the explanatory power of the topic model with respect to causality.

\ifdefined\committeeMemberFourTypedName

\null\hfill \myRule\\
\null\hfill \committeeChairpersonTypedName, \committeeChairpersonPostNominalInitials\\
\null\hfill \myRule\\
\null\hfill \committeeMemberTwoTypedName, \committeeMemberTwoPostNominalInitials\\
\null\hfill \myRule\\
\null\hfill \committeeMemberThreeTypedName, \committeeMemberThreePostNominalInitials\\
\null\hfill \myRule\\
\null\hfill \committeeMemberFourTypedName, \committeeMemberFourPostNominalInitials\\

\ifdefined\committeeMemberFiveTypedName
\null\hfill \myRule\\
\null\hfill \committeeMemberFiveTypedName, \committeeMemberFivePostNominalInitials\\
\fi

\fi
\restoregeometry

\renewcommand{\cftchapdotsep}{\cftdotsep}  
\renewcommand{\cftchapfont}{\bfseries}  
\renewcommand{\cftchappagefont}{}  
\renewcommand{\cftchappresnum}{Chapter }
\renewcommand{\cftchapaftersnum}{:}
\renewcommand{\cftchapnumwidth}{5em}
\renewcommand{\cftchapafterpnum}{\vskip\baselineskip} 
\renewcommand{\cftsecafterpnum}{\vskip\baselineskip}  
\renewcommand{\cftsubsecafterpnum}{\vskip\baselineskip} 
\renewcommand{\cftsubsubsecafterpnum}{\vskip\baselineskip} 

\titleformat{\chapter}[display]
{\normalfont\bfseries\filcenter}{\chaptertitlename\ \thechapter}{0pt}{\MakeUppercase{#1}}

\renewcommand\contentsname{Table of Contents}
\currentpdfbookmark{Table of Contents}{TOC}
\begin{singlespace}
\tableofcontents
\end{singlespace}

\clearpage

\phantomsection
\addcontentsline{toc}{chapter}{List of Tables}
\begin{singlespace}
\setlength\cftbeforetabskip{\baselineskip}  
\listoftables
\end{singlespace}

\clearpage

\phantomsection
\addcontentsline{toc}{chapter}{List of Figures}
\begin{singlespace}
\setlength\cftbeforefigskip{\baselineskip}  
\listoffigures
\end{singlespace}

\clearpage

\clearpage

\blankpage


\clearpage
\pagenumbering{arabic}
\setcounter{page}{1} 

\titleformat{\chapter}[display]
{\normalfont\bfseries\filcenter}{\MakeUppercase\chaptertitlename\ \thechapter}{0pt}{\MakeUppercase{#1}}  
\titlespacing*{\chapter}
  {0pt}{0pt}{30pt}	
  
\titleformat{\section}{\normalfont\bfseries}{\thesection}{1em}{#1}

\titleformat{\subsection}{\normalfont\bfseries}{\thesubsection}{1em}{#1}

\titleformat{\subsubsection}{\normalfont\itshape}{\thesubsubsection}{1em}{#1}

\chapter{Introduction}
\label{chapter:introduction}

Scientific discovery, and knowledge acquisition more broadly, can be understood as two separate processes: \emph{the context of discovery} details the conception of a theory, while \emph{the context of justification} details the validation of a theory \citep{Reichenbach1938}. Both of these processes can be empirically observed. However, accounts of discovery often involve internal and underspecified cognitive processes (e.g., the ``Eureka!'' moment). This led philosophers of science to neglect the context of discovery, and focus instead on justification, which was more amenable to rational analysis \citep{sep-scientific-discovery}. To avoid these ``black boxes'' in the context of discovery, \citet{simon_does_1973} outlines that discovery is not a problem of induction: a normative logic of discovery measures the efficiency of methods for finding patterns in observations. This changes the problem in a key way: rather than finding a set of rules \emph{guaranteed} to find a discovery, a logic of discovery defines a procedure for discovery that can be compared with other procedures, similar to evaluating computational complexity of an algorithm. Rather than examining concepts or proofs, as in the context of justification, we evaluate the efficiency of a particular scheme at postulating new concepts. This conception of a normative logic of discovery is inspired by \citet{Hanson1958}'s \emph{Patterns of Discovery}. This style of reasoning is also known as ``abductive'' or ``retroductive'' reasoning. In abductive reasoning, the perceptual process of observation leads to the discovery of patterns. If these patterns could plausibly be explained by a hypothesis or scientific law, then that law is proposed (the abduction). Finally, the law can be verified through the usual (inductive) logic of justification \citep{Magnani2009,Mabsout2015}.

Moreover, discovery is more than acquiring new conceptual connections. Scientific discovery is inherently a social process \citep{Knorr1999}. Reading is one method of gathering knowledge from the broader culture, while writing is a way to contribute back to that culture. By examining how an author's writings relate to other works published at the same time, we can move from individual behavior to collective behavior in constructing an information environment. Together, these two behaviors reveal ``the interplay between individual and collective phenomena where innovation takes place'' \citep{Tria2014}.

In this dissertation, I operationalize scientific discovery through several case studies of reading and writing behaviors. These behaviors implicitly reflect knowledge acquisition and knowledge creation. To measure changes in knowledge structures, I use LDA topic modeling (Chapter~\ref{chapter:topic-modeling}) to represent the information environment of the texts each author read and wrote. The differences between these texts can then be quantified using information-theoretic measures of semantic similarity (Chapter~\ref{chapter:info-theory}). In contrast to earlier studies of how individuals navigate information environments, I study them at the time scale of decades and years, rather than immediate information retrieval needs at the scale of minutes and seconds. These methods implicitly define a logic of discovery through the relation of readings in the information space to writings, which are the products of knowledge creation or scientific discovery.

In particular, I examine five case studies. Three case studies revolve around Charles Darwin. Darwin left detailed records of every book he read for 23 years, from disembarking from the H.M.S. Beagle to just after publication of \emph{The Origin of Species}. Additionally, he left copies of his drafts before publication. I characterize his reading behavior (Chapter~\ref{chapter:readings}), then show how that reading behavior interacted with the drafts and subsequent revisions of \emph{The Origin of Species}  (Chapter~\ref{chapter:writings}), and expand the dataset to include later readings and writings (Chapter~\ref{chapter:extensions}). Then, through a study of Thomas Jefferson’s correspondence, I expand the study to non-book data. Finally, through an examination of neuroscience citation data, I start an investigation at the level of collective phenomena, examining discipline-wide production as opposed to individual readings.

\section{Bounded Rationality}
People are limited in how much information they can process by the problem structure and by cognitive, temporal, and physical constraints. In response to these constraints, \citet{Simon1956} proposed that decision-makers are ``satisficers,'' searching for a satisfactory solution rather than an optimal solution. These satisfactory solutions are known as heuristics. While many classes of heuristics have been investigated for systematic errors and biases \citep{Kahneman1982}, these heuristics often lead to better decisions than optimal procedures \citep{Gigerenzer1999}. This is often due to reduced processing cost increasing the output rate. These ``fast and frugal'' methods have less computational complexity.




Information processing is bound by physical constraints of energy, time, and thermodynamics.  \citet{merkle_energy_1989} estimates that the human brain is capable of $10^{12}$ to $10^{15}$ logical operations per second, based on \citet{von_neumann_computer_2012}'s estimates of neural density and firing rates. However, these computational estimates do little to improve our understanding of cognition on the level of problem-solving behaviors. In an early seminal work on cognitive load, \citet{miller_magic_1956} declared working memory capacity to be ``the magical number 7, plus or minus 2.'' Work on \emph{schema construction} explored what kind of thing was being held in memory through ``chunking'' \citep{chase_perception_1973}, discovering differences in chunk size with expertise. Experts still held 7 plus-or-minus 2 chunks, but each expert's chunk consisted of more chess moves than a novice's chunk. Experimental paradigms on the timescale of microseconds to minutes have found multiple perceptual bottlenecks in information processing \citep{marois_capacity_2005,tombu_unified_2011}. These include recognition, retention, and response bottlenecks. 
Attention management is a significant concern for human-machine interaction, reviewed in \citet{wickens_engineering_2015}.

In this dissertation, I am interested in cognitive timescales of years and decades. I introduce the notion of ``biographically-plausible datasets,'' data sets of a scale that could be reasonably encountered by a single individual in a lifetime. Biographically-plausible datasets are a complement to big data approaches in cognitive science, discovering what reasonable bounds are for having ``enough'' data bootstrapped from a naturally-occurring data source, such as a digital library or Twitter feed \citep{jones_big_2016}. For reading data, we estimate reasonable bounds as follows: a person reading two books a week for 70 years (from 8 to the United States life expectancy of 78) would have read 7,280 books. This estimate is validated by the size of Thomas Jefferson's Library. Thomas Jefferson had 6,487 books in his early library that was donated to the Library of Congress, and 939 in his retirement library that was curated over 12 years. While it is not likely that Jefferson read all of these books, and there may be overlap in collections, the total of 7,426 is very close to our estimate of 7,280 books in a lifetime of reading. Darwin's reading list consisted of 1,315 titles over 23 years of records, so approximately 1.1 book/week, which is below the estimated data size. The most extreme example we have found is British Prime Minister William Gladstone, who claimed to have read 22,000 books, which is an outlier in what is biographically-plausible. These estimates are an aid in bootstrapping big data into reasonable samples of individual behavior, and expand our understanding of bounded cognition to long-term tasks.

\section{Case Studies}
While the bulk of empirical studies in cognitive science are concerned with measuring population-level effects due to experimental manipulations, case studies play an important role in driving cognitive theorizing, experimentation, and modeling. For example, the case of the memory-impaired patient H.M. drove many advances in cognitive neuroscience and computational models of memory (reviewed in \citet{SquireWixted2011}).  Other case studies, such as that of the frontal-lobe injury in Phineas Gage, provide important contrasts for later studies (reviewed in \citet{Macmillan2000}). 

Detailed longitudinal investigations of a single individual may reveal changing strategies that cannot be observed in laboratory experiments involving a single trial. Cognitive science could be enriched by using longitudinal studies such as ours to design laboratory studies with higher ecological validity. However, it is a challenging task to design laboratory studies of (for example) reading choices that take into account a subject’s extensive prior history of reading decisions.

The case studies of Darwin, Jefferson, and citation networks presented in this dissertation utilize what \citet{goldstone_discovering_2016} call ``naturally-occurring data sets.'' These are the incidental records that are gathered from everyday interactions, such as census records, dictionaries, encyclopedias, financial transactions, tweets, patents, and citations. With the advent of ``big data'' in the broader scientific community, cognitive science is also leveraging these records for behavioral studies \citep{jones_big_2016}. Case studies provide an opportunity to validate big data studies, seeing if individual behavior matches aggregate behavior, potentially highlighting behavioral outliers and opportunities for model revision. While this dissertation focuses on individuals, Chapter~\ref{chapter:future} details some preliminary work to expand our studies to entire scientific disciplines.

\section{Search and Information Foraging}
Search is ``the behavior of seeking resources or goals under conditions of uncertainty.'' This broad definition begins the 2012 volume \emph{Cognitive Search: Evolution, Alogrithms, and the Brain} \citep{todd_cognitive_2012}. Covering much more than use of a search engine, cognitive search is a fundamental component of human and non-human animal behavior. In the strongest view, all human problem solving can be understood as defining, refining, and resolving a search problem \citep{Newell1972}. Even without such a strong claim, search governs memory retrieval, spatial foraging, problem solving, scientific discovery, creativity, and perception from individual to group to societal cognition \citep{Hills2015}.

Individual scientists and scholars can be viewed as conducting a cognitive search in which they must balance \emph{exploration} of ideas that are novel to them against \emph{exploitation} of knowledge in domains in which they are already expert \citep{Berger-Tal2014}. This exploration-exploitation trade-off is a strategy for managing cognitive load. Researchers have studied the exploration-exploitation trade-off in cognitive search at timescales of minutes up to years and decades. Laboratory experiments on visual attention are one example of this balancing act on short timescales \citep{chun1996just}, while studies of the recombination of patented technologies demonstrate long-term group behavior \citep{Youn2014}. Human subject results on the exploration-exploitation trade-off are reviewed in \citet{Cohen2007}. These studies find that there is no general-purpose optimal solution to this trade-off. Rather than finding a single optimal solution with a stationary strategy, \citet{Berger-Tal2014} proposed that the exploration-exploitation trade-off is a dynamic system, varying with time to optimize towards a particular goal. This framework lists four major phases of exploration-exploitation trade-off over the course of an entire lifetime: \begin{enumerate}
\item \emph{Knowledge acquisition}, in which all energy resources are devoted to exploration;
\item \emph{knowledge accumulation}, in which the initial search landscape is exploited;
\item \emph{knowledge maintenance}, in which an satisfactory level of knowledge is maintained --- in other words, a satisficing stage that balances exploration with exploitation;
\item \emph{knowledge exploitation}, in which no new exploration occurs. 
\end{enumerate}
\noindent 


\subsection{Information Foraging Theory}
\emph{Information foraging theory} \citep{Pirolli1999} expands cognitive search to include an information environment about which agents have incomplete information. IFT ``aims to explain and predict how people will best shape themselves for their information environments and how information environments can best be shaped for people'' \citep{Pirolli2009}. The general problem of information foraging has been explored in many fields, including cognitive psychology \citep{todd_cognitive_2012,Hills2015}, neuroscience \citep{Cohen2007}, economics \citep{March1991,Azoulay-Schwartz2004}, finance \citep{Uotila2009}, ecology \citep{Stephens1986,Eliassen2007}, and computer science \citep{Sutton1998}. In all of these areas, the searcher aims to enhance future performance by surveying enough of existing knowledge to orient themselves in the information space.

IFT draws upon the framework, theories, and models of optimal foraging theory (OFT) in biology \citep{Stephens1986}. OFT aims to explain how animals search for food, taking into consideration an organism's metabolism and the environmental structure. There are two primary variables optimized in OFT: $time$ and $energy$. In information foraging, $energy$ is analogous to $attention$:
\begin{quote}
When information consumes is rather obvious: it consumes the attention of its recipients. Hence a wealth of information creates a poverty of attention, and a need to allocate that efficiently among the overabundance of information sources that might consume it. \citep{Simon1971b}
\end{quote}

IFT has been developed through the rational analysis of the information foraging problem space, operationalizing foraging behaviors through use of the scatter-gather experimental model. This led to the creation of several models, including ACT-IF \citep{Pirolli1999}, MESA \citep{Miller2004}, CoLiDeS \citep{Kitajima2005}, and SNIF-ACT \citep{Pirolli2003,Fu2007}. Most of these models have studied how humans search for information on the World Wide Web, with positive results from applying OFT to model human behavior \cite{Pirolli2009,Fu2012}. This dissertation pushes these models beyond laboratory studies, using naturally-occurring, biographically-plausible datasets to quantify changes in foraging behavior between exploitation and exploration regimes.

Information foragers decide what information to allocate attention to through \emph{information scent}, which is a general-purpose concept that refers to various indicators of information quality \citep{chi_using_2001}. Examples of information scent include summaries, such as those presented in web search results, in scientific abstracts, or on the jacket of a book; structured data, such as computer folders or cataloging information; and social cues, such as recommendations \citep{Pirolli2009}. Most measures of information scent are presented as goal-oriented. In the historical studies presented in this dissertation, we defer studies of information scent to future work.

\section{Operationalization}
Our primary interest is in the texts that a particular author read and wrote. We modeled these texts using using Latent Dirichlet Allocation (LDA; \citet{Blei2003,Blei2012}), a type of probabilistic topic model. LDA is a generative model that represents each document as a bag of words generated by a mixture of topics. Furthermore, topic models predict the behavior of human subjects in a variety of word association and disambiguation tasks \citep{Griffiths2007}. Topic modelling goes beyond simple word-counting methods, because it captures the ways in which the meaning of words can shift given the contexts in which they occur.

LDA topic models have been extensively used to analyze human-generated text across many domains.  Our approach contrasts with previous uses of topic modeling to analyze the large-scale structure of scientific disciplines \citep{Griffiths2004,Hall2008,Blei2007,Cohen2015} and the humanities \citep{Poetics41, JDH, Jockers2013}, which are each created through the collective effects of individual-level behavior. 
Previous models of historical records have focused on language use as an indication of larger shifts in style \citep{Hughes2012,Underwood2012}, learnability \citep{hills2015recent}, or content \citep{Michel2011,Goldstone2014,Klingenstein2014} of significant portions of publications in a field, including a study of the journal \emph{Cognition} \citep{Cohen2015}. Our application of topic modeling to the works read by a single person is a novel way to quantify an individual's information environment.

Rather than proposing that an individual's exploration-exploitation behaviors are in response to an overall goal, behavior changes relative to the information environment. People make decisions about which text to read next and we quantify the relative surprise of these decisions, independent of their goal, using information theoretic measures. Work on quantifying semantic relations stretches back to \citet{Osgood1967}'s semantic differential. The development of the vector space model, which represents documents were as weighted term frequency vectors, introduced the use of cosine distance for semantic similarity \citep{Salton1975}. As LDA topic models are a generative model, each document-topic distribution can be viewed as a source generating words for the document. This signal (the words) can be analyzed by information theoretic measures.

The first measure is the \emph{Kullback-Liebler divergence} (KL divergence) \citep{Kullback1951}. KL divergence measures the inefficiency of using one distribution to encode signals coming from a different distribution.
KL divergence is asymmetric, a fact we leverage through a notion of \emph{enclosure} that determines whether one text is more comprehensive than another. In situations where a symmetric measure is preferred, for example in mutli-document comparisons or situations where temporal order is not a factor, we use the symmetrized KL divergence, known as the \emph{Jensen-Shannon divergence} (JSD) \citep{Lin1991}. The square root of the JSD is a true metric, known as the Jensen-Shannon distance \citep{Endres2003}. The JS divergence does not satisfy the triangle inequality, while the square root does \citep{fuglede2004jensen}. By analogy, KL divergence and JSD can be thought of as different answers to the question: ``how far away is the ski lodge?'' Answering with JSD is akin to answering with the distance in miles. Answering with KL divergence is akin to answering with the travel time, which may be different depending on whether you are going up or down the mountain. This dissertation formalizes these measures in Chapter~\ref{chapter:info-theory}.

The dissertation uses KL divergence as the basis for many different analyses between texts. When examining Darwin's reading history, we use the past-to-text and text-to-text divergence to quantify the amount of ``surprise'' in each reading decision, based on either the cumulative or immediate information environment. In the case of writing comparisons, we introduce the notion of \emph{enclosure}, which uses the asymmetries of the KL divergence to determine whether one text summarizes or is more comprehensive than another.

Our approach also views the exploration-exploitation trade-off as a dynamic process, but does not claim there is any goal-directed behavior driving these behavioral changes. Rather, exploration and exploitation behaviors change in response to the information environment intrinsic to the texts, and changing with the addition of new publications.


\section{Null Models for Case Studies}
As we utilize case studies, careful consideration must be given to a null model for significance testing. In our studies of reading behavior, we permute the order of each set of readings to generate possible reading orders. The selection at each reading date is constrained by the titles published at that date, as a book could not be read before its publication. In contrast to a null that includes permutations that neglect publication date, this restricted null captures the dynamics of publication in which a new work can unexpectedly change the information space. The process also avoids directly including a temporal variable in the textual model. In addition, using a sampling process over the same data set allows us to use the same topic model for each null, which in turn allows us to avoid making further decisions about topic model comparison. This model of reading permutations is used in Chapters~\ref{chapter:readings}, \ref{chapter:writings}, and \ref{chapter:extensions}.

In the case of writing behavior, we utilize a method known as ``query sampling'' to generate multiple topic distributions for each writing, based on the topics previously learned from a set of readings. This population of possible distributions is then compared to the reading topic distributions and to the distributions of other writings. The replications allow us to calculate a confidence interval for each distance computation, which also helps us assess fitness of the topic distributions for these out-of-sample texts \citep{Trosset2008}.

Cluster analysis of the topic distributions for each text reveal that query sampling generates stable ``intepretations'' governed by their dominant topic. For example, \emph{The Origin of Species} can be understood in terms of case studies (dominant topic about ``pigeons''), geography (dominant topic about various locations), or theory (dominant topic on ``development'', ``experiment'', ``geological''). Rather than being an indicator of the model's instability, this variability of outputs supports the historical interpretation of our case studies. In the humanities, the interpretation of texts is a central (if not \emph{the}) scholarly endeavor. Our approach does not claim a single ``correct'' interpretation, but rather a set of interpretations in dialog with one another. Digital methods augment existing debates in the humanities by providing quantitative evidence for different ways of examining a text \citep{Rockwell2016,Allen2017}. We explore these multiple interpretations in Chapter~\ref{chapter:writings}.

It is important to emphasize again that we make no teleological claims about the foraging behavior. Our null models do not inject any assumptions that Darwin was reading in order to write \emph{The Origin}. When speaking of an optimal information forager, we mean the shortest path within the information environment, not the shortest path in relation to another text, as measured by semantic similarity measures.

Finally, these null models are not common in cultural analytics, and stand as a contribution to the current state of the art. When looking at randomized samples from large datasets, the correlations in each null sample are distorted. By using randomized samples from a biographically-plausible data set, we maintain some of the correlations in the texts, meaning that differences from random selection must also overcome the correlations in the selections of the data set itself. This is clearly seen in the case of Darwin's readings, where the books were selected to investigate particular phenomena, so have intrinsic correlations, yet the path Darwin chose is less surprising than random selections among those books.

\section{Limitations}
Given the theoretical constructs of a logic of discovery, bounded cognition, and information foraging theory, this dissertation defers investigation of several key concepts to future work. One limitation of our methods is that bounded cognition is not explicitly treated in these case studies. There is no investigation of how attention was managed, only the assumption that it was and that we observe the consequences of that attention management in biographically-plausible datasets. In Chapter~\ref{chapter:extensions}, we speculate on which books Darwin could have read by examining the content of his retirement library. Future work could also examine Darwin's ``books to be read'' list contained in the same notebooks we analyzed in Chapter~\ref{chapter:readings}.

While these case studies are adjacent to the study of innovation and knowledge creation, we do not touch on the creative process---how ideas are combined to form new ones. While we affirm that works like \emph{The Origin of Species} are novelties, and show how they relate to the readings, we don't explain the writing process. Despite examining questions of cultural production, we don't draw distinctions between novelties and innovations, understood through \citet{boden2004creative}'s notion of \emph{p-creativity} and \emph{h-creativity}. In \emph{p-creativity} or personal creativity, a concept is new to an individual. In \emph{h-creativity} or historical creativity, a concept is new to an entire society. We only briefly touch on issues of precedence in an analysis of Alfred Russell Wallace's co-discovery of natural selection (Chapter~\ref{chapter:writings}).

Another limit is that while these case studies describe and operationalize different portions of a logic of discovery, we do not define a generative model that could itself stand as a logic of discovery. Exploratory work through a case study of over 100,000 authors in neuroscience and their publication abstracts is found in Chapter~\ref{chapter:future}. Similar to astronomy before the development of optics, long-term information foraging is in an era of theory-making and discovering what relevant phenomena should be observed. This dissertation is a step in that direction.


\chapter{Methods: Topic Modeling}
\label{chapter:topic-modeling}

Probabilistic topic models represent documents as mixtures of ``topics'' that correspond to common semantic themes in a collection \citep{Blei2012}. The specific method that we use---topic modeling via latent Dirichlet allocation (LDA; \citet{Blei2003})---applies a Bayesian learning algorithm to identify these topics as patterns of co-occurring words. In probabilistic topic models, each document is represented as a probability distribution over topics, and each topic is itself a probability distribution over words. It is a generative model of language, rather than a descriptive model, and we interpret it here as a theory about how documents are written. It posits that each document has a distribution of contexts, or topics, that it is composed of. The document is created by selecting first a context and then selecting terms from that context. For example, \emph{The Origin of Species} is a mixture of concepts from geology, animal husbandry, zoology, natural philosophy, and geography. The words used in each of those contexts will differ, but may have some overlap (for example, geography and geology). The entire document is a mixture of words from this mixture of contexts. 

\begin{figure}
    \centering
    \includegraphics[width=.5\textwidth]{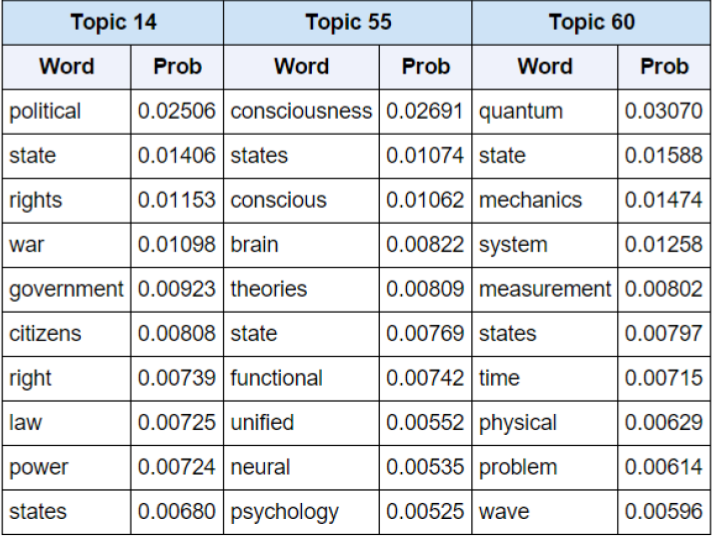}
    \caption{\textbf{The term ``state'' is a highly probable term for several topics inferred from the Stanford Encyclopedia of Philosophy.} It appears in topic 14, indicating political states; topic 55, indicating psychological states; and topic 60, indicating physical states. Notice that both state and states appear in each topics' top 10 terms. This is because we do not employ stemming in our corpus preparation steps. Terms that appear in other topics' top 10 terms are not absent from the other topics, but rather of a low probability in the topic distribution.}
    \label{fig:word-contexts}
\end{figure}

Topic modelling goes beyond simple word-counting methods, because it captures the ways in which the meaning of words can shift given the contexts in which they occur. For example, Figure~\ref{fig:word-contexts} shows how the word ``state'' appears as highly probable in multiple distinct topics inferred from the Stanford Encyclopedia of Philosophy. Furthermore, topic models predict the behavior of human subjects in a variety of word association and disambiguation tasks \citep{Griffiths2007}.

{}

The power of topic models lies in their ability to summarize themes from a large collection of texts. These themes are learned in an unsupervised manner from only a single observed variable: the words on a page. The uncanny nature of topics to be readily interpreted has led to high adoption by the digital humanities community \citep{Poetics41, JDH, Jockers2013}. In the humanities, the interpretation of texts is a central (if not \emph{the} central) scholarly endeavor. Digital methods augment existing debates in the humanities by providing quantitative evidence for different ways of interpreting a text \citep{Rockwell2016}. 
\section{Topic Model Evaluation and Interpretation}
The dialog between humanists and computer scientists has led to a different evaluation paradigm for LDA topic models than for most machine learning algorithms. Rather than optimizing models for mathematical likelihood, topic models are frequently selected based on qualitative fitness. As there is no gold standard list of topics for every corpus, \citet{chang2009reading} performed large-scale user studies for evaluating topic models and introduced several measures of evaluation. These measures draw inspiration from the psychological testing that was used to verify latent semantic analysis (LSA), a precursor to LDA \citep{landauer_computational_2002}. The evaluation tasks were:
\begin{enumerate}
    \item The \emph{word intrusion} task, in which users are presented with six terms, including the five most frequent terms in the target topic and asked to pick the outlier, which is a low-frequency term in the target topic, but a high-frequency term in another topic. 
    \item The \emph{topic intrusion} task, in which users are shown a title and excerpt form a document. They are then presented 4 topics, represented by the 8 most probable words in each topic. Three of the four are the most likely for the document, while one is an intruder that they must be identify.
\end{enumerate}
\noindent Comparison of human performance on these tasks to traditional measures of mathematical likelihood showed that topic coherence actually \emph{decreased} in the models that had higher likelihoods! \citet{lau_machine_2014} automated these tasks, achieving near-human performance by sampling lexical probabilities from a large reference corpus (for example, Wikipedia or the New York Times archives) and calculating Pointwise Mutual Information (PMI; \citet{church_word_1990}) between terms to determine the outlier. They were able to match the performance of an individual human evaluator, but not enough to overcome the wisdom of the crowd with a single reference corpus \citet{yi_wisdom_2012}. Human performance was matched with multiple automated evaluations against multiple reference corpora.

Visualization is another qualitative way to explore a topic space. Several systems display topic-document probabilities with bar graphs, while representing the topic space using dimensionality reduction on the word-topic matrix \citep{Chuang2013,Sievert2014,Murdock2015}. In this dissertation, models were spot checked with the InPhO Topic Explorer to ensure both topic quality and source-document quality \citep{Murdock2015}.

Two quantitative approaches are held-out document testing and document completion tasks, evaluated in \citet{Wallach:2009:EMT:1553374.1553515}. In these, the generative model is used to model unseen text --- either held-out data or the second half of a document --- then evaluated for log-likelihood given the resulting topic-word assignments. These evaluation methods are only relevant to the new documents though, and say nothing about the latent variables that form the model \citep{chang2009reading}.

One common error in interpreting topics is the conflation of LDA with a cluster analysis. \emph{Words do not ``belong'' to a single topic.} LDA is a mixed-membership model, in which words are shared across all topics in different proportions. The conflation of document-topic membership is a less frequent error, but of the same class. This error is compounded by the way topics are often presented as their top 10 most-probable words. These top-10 words do not fully represent the topic \citep{Schmidt2012}. In Chapter~\ref{chapter:writings}, we discover that in a 200-topic model, the most likely topic requires 530 words to comprise 50 percent of the probability mass. 

Finally, it is worth emphasizing that evaluation is only a guide to selecting a better model. Our approach does not claim a single ``correct'' topic model, but rather that each model provides its own interpretation of the collection. Strong hypotheses about the relations between documents should hold among a population of topic models and with differing number of topics. We cross-validate across number of topics in Chapter~\ref{chapter:readings} and further address the multiple interpretation of texts in Chapter~\ref{chapter:writings}.

\section{Geometric Interpretation}
Topic models define a sophisticated geometry of meaning for comparison of texts. We build these intuitions from the ground-up, starting with the \emph{vector space model} (VSM; \citet{Salton1975}). In the VSM, each document is represented as a vector of weighted term frequencies. Term weights are determined by calculating \emph{term frequency-inverse document frequency} (TF-IDF). For purposes of similarity jugements, terms that occur frequently across many documents or rarely among only a few documents are not as important as terms that occur frequently across a few documents. To reduce the effects of the curse of dimensionality, both frequent-pervasive terms and rare-sparse terms may be filtered from the term frequency vectors. Terms in a VSM are assumed to be orthogonal, as they form the axes of the vector space. This present challenges for information retrieval, as a term must be present in a document for it to have a non-0 similarity to the query vector.

Latent semantic analysis (LSA; \citet{Dumais2004}) addressed the orthogonality of terms by performing a singular value decomposition (SVD) on the term-document frequency matrix. This accounts for correlations between co-occurring terms and for correlations between documents which share terms. Documents and terms are both represented as vectors in a shared vector space, meaning that document-document, term-term, and term-document similarities can all be easily computed.

Probabilistic latent semantic indexing (pLSI; \citet{Hofmann1999}) was a significant advance in semantic modeling because it introduced a generative model of texts. Each word in a document is sampled from a topic distribution, the components of which are a word distribution. This is very similar to LDA, except that there is no probabilistic model at the document-level. This causes the number of parameters to grow linearly with the size of the corpus, as each document's mixture must be determined a priori. increasing the likelihood of overfitting. It also prevents the model from handling out-of-sample documents. However, pLSI also shifted the geometry significantly from vector spaces to simplexes. Rather than representing terms or documents as vectors, documents become points within a topic simplex. A topic is a point embedded in a word simplex. For our purposes, a simplex is a geometric representation of a probability space.

LDA expanded on pLSI by positing that the topic distribution for each document is drawn from a distribution with a randomly chosen parameter. This parameter is sampled once for each document from a smoothed distribution over the topic simplex. Another advantage of introducing this distribution is that it allows distances between documents to be measured with information theoretic measures. The LDA generative process posits each document as a source, producing a stream of tokens (i.e., a signal), that can be compared to the signal of other documents.

\begin{figure}
    \centering
    \includegraphics[width=0.75\textwidth]{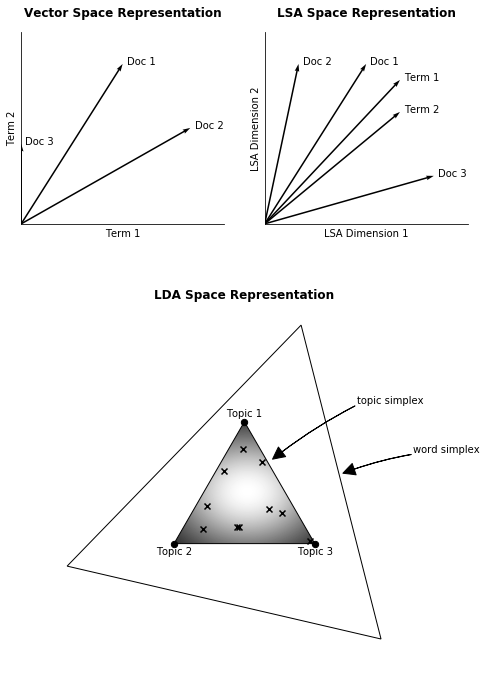}
    \caption{\textbf{Geometries of Various Semantic Models}. On the top row, the vector space and latent semantic analysis models are shown using the representation of Figure 4.1 of \citet{Dumais2004}. In the VSM, each axis is defined by a term. In LSA, each axis is defined by a latent factor, enabling comparison of term-term similarities. On the bottom row, the word and topic simplexes of probabilistic latent semantic indexing and latent Dirichlet allocation using the representation of Figure 4 of \citet{Blei2003}. In pLSI, each document-topic distribution is represented by an \textbf{x}, showing that the distribution is generated external to the topic simplex. In LDA, possible distributions are represented by the gradient, showing the generative process drawn over the topic simplex itself.}
    \label{fig:model-geometries}
\end{figure}

Figure~\ref{fig:model-geometries} summarizes the various geometries defined by the VSM, LSA, pLSI, and LDA semantic models.

\section{Formal Definition}
From this point on, I move to formal discussion of the algorithm, parameter selection, model comparison techniques, and how to handle out-of-sample documents.

The generative model of LDA is:
\begin{enumerate}
    \item For each document $d$, generate a topic distribution: $\theta \sim \operatorname{Dirichlet}(\alpha)$.
    \item For each topic $t$, generate a word distribution: $\phi \sim \operatorname{Dirichlet}(\beta)$.
    \item For each word position $i$ in a document $d$:
    \begin{enumerate}
        \item Given a document's topic distribution, select a topic assignment: $z_{di}|\theta_d \sim \operatorname{Discrete}(\theta_d)$.
        \item Given that topic's word distribution, select a word: $w_{di}|\phi_t : t = z_{di} \sim \operatorname{Discrete}(\phi_t)$
    \end{enumerate}
\end{enumerate}

\noindent Each distribution in $\theta$ and $\phi$ is drawn from a Dirichlet process, which is a distribution of distributions. $\theta$ is the topic-document matrix, with dimensions $k \times D$, where $k$ is the number of topics and $D$ is the number of documents. $\phi$ is the word-topic matrix with dimensions $V \times k$, where $V$ is the number of unique terms in the corpus (the vocabulary). The Dirichlet process for each matrix is governed by a single concentration parameter, the Dirichlet prior. $\theta$ is governed by $\alpha$ and $\phi$ by $\beta$. Each topic and word assignment is drawn from the $\theta$ and $\phi$ distributions. The joint distribution for the generative model is:
\[
p(\mathbf{w},\mathbf{z},\theta,\phi|\alpha,\beta) = p(\theta|\alpha) p(\phi|\beta) \prod_i p(z_i|\theta) p(w_i|z_i,\phi) 
\]

\subsection{Inferring Topics}
Topics are not pre-specified, but learned through an inference process over texts to determine the posterior distribution of three latent variables: the word-topic distributions ($\phi$), the topic-document distributions ($\theta$), and topic assignments ($\textbf{z}$). This inference process relies on a single observed variable, the words in the documents ($\textbf{w}$), and two pre-specified hyperparameters, known as Dirichlet priors ($\alpha$ and $\beta$). Formally, the learning process must evaluate the posterior distribution:

\[
p(\mathbf{z},\theta,\phi|\mathbf{w},\alpha,\beta) = \frac{p(\mathbf{w},\mathbf{z},\theta,\phi|\alpha,\beta}{p(\mathbf{w}|\alpha,\beta)}
\]

\noindent  This is a computationally intractable task without sampling methods \citep{Blei2003}. The original LDA paper used a method known as \emph{variational inference} that inferred distributions via expectation maximization. 

Our implementation uses a collapsed Gibbs sampler (CGS; \citet{geman_stochastic_1984}) to estimate $\theta$ and $\phi$ from the topic assignments $\textbf{z}$. Using the derivation of \citet{Griffiths2004}, our sampling process integrates out $\theta$ and $\phi$, which in turn removes their Dirichlet priors, $\alpha$ and $\beta$.\footnote{Most implementations of the Gibbs sampler use the Dirichlet priors as a smoothing factor in the calculations, including \citet{Griffiths2004}. We defer discussion of priors for simplicity of presentation.} This simplifies the posterior:

\[
p(\mathbf{z}|\mathbf{w}) = \frac{p(\mathbf{w}, \mathbf{z})}{\sum_z p(\mathbf{w},\mathbf{z})}
\]

\noindent However, this distribution is also computationally intractable, as the sum in the denominator does not factorize and involves $k^V$ terms. The CGS is a Markov chain Monte Carlo (MCMC) method, which sequentially samples from the distribution of $\mathbf{z}$ \citep{andrieu_introduction_2003}. First, we randomly initialize each word's topic assignment, $z_{di}$. Then, for each word, we update the assignments based on the current state of all but the current $z_{di}$. The number of times a word in document $d$ has been assigned to topic $t$ is given by $N_{td}$. The number of times a word $w$ has been assigned to topic $t$ is given by $N_{wt}$.

\[
p(z_{di} = k | \mathbf{z}^{\neg di}, \mathbf{w}) = \frac{1}{Z} N^{\neg di}_{wk} N^{\neg di}_{kd}
\]

\noindent where $Z$ is the normalization constant:

\[
Z = \sum_t N^{\neg di}_{wt} N^{\neg di}_{td}
\]

\noindent Given the value sampled for $z_{di}$, the counts $N_{td}$ and $N_{wt}$ are updated. 

After obtaining values for $\mathbf{z}$ we can then estimate $\hat\theta$ and $\hat\phi$:

\[
\hat\theta_{dt} = \frac{N_{dt}}{N_d}
\]
\[
\hat\phi_{wt} = \frac{N_{wt}}{N_t}
\]

\noindent The MCMC method is guaranteed to converge to the true posterior for $\theta$ and $\phi$ \citep{gordin_central_1978,Griffiths2004}. However, determining how many iterations are needed in order to converge within an acceptable error is a challenge, with diagnostic methods reviewed in \citet{cowles_markov_1996}.

\section{Parameter Selection}
LDA relies on a priori parameter selection for the number of topics and the Dirichlet priors. There are methods for estimating each based on an existing corpus. Results in later chapters were robust to variation in the number of topics, which is one of the primary reasons to explore optimization of the Dirichlet priors. In short, parameter selection does impact the model and there are better ways than ``pick a number,'' but if strong experimental design picks up an effect, it is likely that parameter optimization results only in signal boosting.  This section is presented mostly for situational awareness.

\subsection{Number of Topics ($k$)}
In general, in a model with too few topics, each topic becomes very general and hard to interpret. With too many topics, some of the topics are specialized on just a few documents, making them less useful for finding common themes. Selection of the number of topics for an LDA topic model is often seen as an opportunity for ensemble methods to determine the best model out of a population of models with varying values of $k$. \citet{greene_how_2014} proposed a novel stability method for comparing results across different numbers of topics to aid in this selection. Previously, \citet{Wallach2009} found that the Dirichlet priors impacted stability of results on a measure of topic coherence. \citep{tang_understanding_2014} relate the selection of number of topics to document length and number of documents.

We take a different approach. Our general philosophy about topic interpretation (above) highlights that different models may be better suited for different interpretive purposes. This human component of topic model evaluation can be dismissed as ``reading the tea leaves,'' as \citet{chang2009reading} highlight before validating that approach with human subjects. For the study of Darwin's reading lists (Chapter~\ref{chapter:readings}), we use a low number of topics ($k=80$), coarse graining his path through the information space. In the study of how the drafts of \emph{The Origin} relate to the reading lists, we wanted more specialized topics to distinguish between particular concepts that entered and exited each draft, so increased the number of topics to $k=200$.

\subsection{Dirichlet Priors ($\alpha,\beta$)}
\begin{figure}
    \centering
    \includegraphics[width=\textwidth]{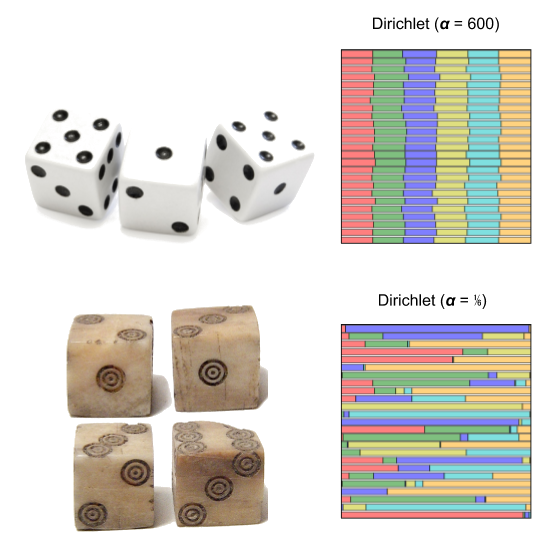}
    \caption{\textbf{Examples of Dirichlet priors}. In this example we compare two dice factories. On the top is a modern factory, that uses plastics and laser cutting to ensure consistency across dice. Each row on the right represents the probability distribution of a single die made in this factory, showing there are still some variations among the dice with a prior of $\alpha = 600$. This is in contrast to the bottom, which represents an artisinal shop that makes dice out of bones. The carvings are not consistent across dice. There is a huge amount of variation among the dice with a prior of $\alpha = \frac{1}{6}$. }
    \label{fig:dirichlet-priors}
\end{figure}

A Dirichlet process has a single (hyper)parameter, the Dirichlet prior, which controls the uniformity of distributions generated by the process. In LDA, $\theta$ is governed by the $\alpha$ prior and $\phi$ is governed by $\beta$. One way to think of the Dirichlet priors is as a measure of quality control. Say you have a factory that makes dice. A high-quality factory with laser-cutting machinery and plastic molds will make very uniform dice, therefore the Dirichlet prior will be high. By contrast, an artisinal dice manufacturer that hand-carves dice out of bone, a non-uniform material will have a much lower Dirichlet prior. This example is illustrated in Figure~\ref{fig:dirichlet-priors}.

Priors can be symmetric or asymmetric across document-topic or word-topic distributions. \citet{Wallach2009} examined the effect of prior selection on model robustness to changes in the number of topics and to highly-skewed word distributions. They found that changes to $\beta$, the word-topic Dirichlet prior, had no significant effect. However, changes to $\alpha$, the document-topic Dirichlet prior did impact results. Specifically, they found that an asymmetric prior $\alpha$ ensures commonalities across documents are preserved, while the symmetric prior over $\phi$ ensures that topics do not overlap. Our implementation uses symmetric priors of $\alpha=0.1$ and $\beta=0.01$ throughout the dissertation for ease of implementation. As our results are cross-validated across different numbers of topics, improvements from an asymmetric $\alpha$ would serve only to make results more significant, based on Wallach et al.'s findings.

\section{Corpus Preparation}
More important than parameter selection is corpus selection and preparation \citep{tang_understanding_2014}. \citet{Boyd-Graber2014} detail the initial text preparation steps necessary to generate robust topic models. Below, I review the literature on variations in topic model performance with respect to document length, stoplisting, and stemming, justifying some of the decisions we made in modeling.

\subsection{Document Length}
Throughout the Darwin case studies, we are modeling book-length documents. In the Jefferson study, we are modeling both letter-length and book-length documents. In the exploratory citation network study, we are modeling abstract-length documents. Practitioners have also modeled a wide variety of document lengths, from tweets \citep{tang_understanding_2014,cheng_btm:_2014} to abstracts \citep{Griffiths2004} to news articles \citep{Blei2003} to books \citep{mimno_organizing_2007,Newman:2010:ETM:1816123.1816156}. Heuristics have been applied to shorten overly-long documents, such as books, by reducing them to chapters or pages, as in \citet{mimno_organizing_2007}.

\citet{tang_understanding_2014} carried out a study examining how different variables affect the performance of topic models. First, they discovered that topic quality is governed primarily by the number of documents. Of secondary importance is document length, where they found that sampling a fraction of each document yields comparable topics to modeling the entire document. Poor performance is found when documents are too short.

\subsection{Stoplisting}
Stopword removal plays a critical role in semantic models and information retrieval. In the formulation of the vector space model, frequently-occurring, widely-distributed terms were excluded from the term frequency calculations, as were rarely-occurring, document-specific terms \citep{Salton1975}. These greatly improved the quality of document similarity judgments. In many information retrieval systems, stop lists are not based off corpus characteristics, but a stop word list for a particular language. Stop listing creates issues for information retrieval tasks, as search queries may depend on frequently-removed words \citep{manning2008introduction}. In practice, stoplisting removes all \emph{tokens} (occurrences) of a particular \emph{type} (term).

Specifically for stop listing LDA, \citet{schofield_pulling_2017} demonstrated that very high-frequency term removal did improve topic quality, but removing very low frequency terms had negligible impact on topic quality. The authors propose removing these words \emph{after} inference, instead of before inference. For most of our experiments, we removed types accounting for the top 50\% and the bottom 10\% of the corpus's tokens \emph{before} inference. This decision was made because while topic quality may not have been significantly impacted, there are undeniable impacts on memory and processing requirements. In the expanded Darwin corpus we reduced the number of types from 3,293,885 to 9,123 by filtering the top 75\% and bottom 5\% of tokens. This may seem like a high number, but 3.1 million of the removed terms, accounting for only 2.5\% of tokens in the corpus, appeared less than 23 times across the volumes. Corpus-specific stop listing was very helpful for our data source---OCR errors resulted in many low-frequency ``junk'' tokens.

\subsection{Stemming}
Stemming is another frequently applied data normalization method in information retrieval \citep{manning2008introduction}. Rather than having separate terms for ``apples'' and ``apple'', stemming collapses the terms to a single term: ``apple.'' This intuition is even more helpful for verbs: ``advance'', ``advancing'', ``advanced'', and ``advancement'' would all be collapsed to ``advance''. However, stemming has been shown to be detrimental to topic quality  \citep{Boyd-Graber2014,schofield_comparing_2016}. We did not implement stemming in our datasets.

\section{Model Training Scenarios}
We now examine three scenarios for training, and then comparing, models over multiple corproa:

\begin{itemize}
\item \textbf{Corpus merging} -- You have a corpus from one source and a corpus from another source. The goal is to merge them into one corpus \emph{before} training a model(s) over the combined corpus.
\item \textbf{Model comparison} -- You have two models, trained on the same or different corpora. The goal is to measure the distance between models.
\item \textbf{Query sampling} -- You have a corpus from one source and have previously trained models on it. The goal is to fit new documents into the topic space defined by only the first-source documents..
\end{itemize}



\subsection{Notation}
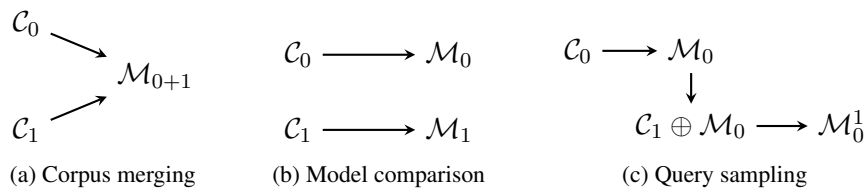
\begin{figure}[h!]
    \centering
    \subfloat[Corpus merging]{
    \begin{tikzpicture}[node distance=1cm]
    \node (Ma) [] {$\mathcal{M}_{0 + 1}$};
    \node (C0a) [above left of=Ma, xshift=-1cm] {$\mathcal{C}_0$};
    \node (C1a) [below left of=Ma, xshift=-1cm] {$\mathcal{C}_1$};
    
    \draw [arrow] (C0a) -- (Ma);
    \draw [arrow] (C1a) -- (Ma);
    \end{tikzpicture}
    }
    \subfloat[Model comparison]{
    \qquad
    \begin{tikzpicture}[node distance=1cm]
    \node (M0) [] {$\mathcal{M}_{0}$};
    \node (C0) [left of=M0, xshift=-1cm] {$\mathcal{C}_0$};
    \node (C1) [below of=C0] {$\mathcal{C}_1$};
    \node (M1) [below of=M0] {$\mathcal{M}_1$};
    
    \draw [arrow] (C0) -- (M0);
    \draw [arrow] (C1) -- (M1);
    \end{tikzpicture}
    \qquad
    }
    \subfloat[Query sampling]{
    \begin{tikzpicture}[node distance=1cm]
    \node (M0) [] {$\mathcal{M}_{0}$};
    \node (C0) [left of=M0, xshift=-0.5cm] {$\mathcal{C}_0$};
    \node (QS) [below of=M0] {$\mathcal{C}_1 \oplus \mathcal{M}_0$};
    \node (M1) [right of=QS, xshift=1cm] {$\mathcal{M}_0^1$};
    
    \draw [arrow] (C0) -- (M0);
    \draw [arrow] (M0) -- (QS);
    \draw [arrow] (QS) -- (M1);
    \end{tikzpicture}
    }

    \caption{\textbf{Three methods for working with multiple corpora.} a) \emph{Corpus merging} takes two corpora and trains a single model over their joint vocabulary. b) \emph{Model comparison} takes two corpora and trains two separate models, with independent vocabularies. c) \emph{Query sampling} takes a model trained on an initial corpus and projects a new document into that model space, using only the vocabulary from the initial corpus.}
    \label{fig:training-scenarios}
\end{figure}

Throughout this section $\mathcal{C}$ refers to a corpus, $\mathcal{M}$ refers to a topic model. $\mathcal{C}_0$ is the initial corpus, $\mathcal{C}_i$ is the $i^\mathrm{th}$ corpus. $\mathcal{M}_i$ is the model trained over $\mathcal{C}_i$. When a superscript is used with a model variable, it indicates the documents included in the model, while the subscript indicates the vocabulary of the underlying model. For example, $\mathcal{M}^1_0$, is a model trained on $\mathcal{C}_1$, using only the vocabulary found in $\mathcal{C}_0$. The $+$ indicates the union of two corpora. The three model training scenarios are detailed using this notaiton in Figure~\ref{fig:training-scenarios}.

\subsection{Corpus Merging}
Corpus merging is the process of collating data from multiple sources before training a model. Many times, topic models are trained off a single source: texts from a single digital library, online reviews, etc.

Corpus merging is not a ``hard'' modeling problem -- simply extract the text from multiple sources and train a model. However, it is worth making the corpus merge an explicit consideration when modeling multiple corpora. Different pre-processing and normalization steps must be applied to different data sources. Awareness of these corpus-specific properties is necessary to make sure we extract the proper \emph{content} from a document for modeling \citep{Boyd-Graber2014}. For example, using OCR text from the HathiTrust often benefits from using dictionary or fequency-based filtering to remove erroneous scans such as ``ert+ y76p''. There are also certain classes of errors - in some 17th century texts ``some'' may be OCR'd to ``fome''. Another example of non-content material is publication information on th title page of a text. In \cite{Murdock2017b}, topic 8 contains ``8vo'', ``vols'' and other terms from the opening covers of each book. This creates topics correlated with particular publication years correlated due to publishing types, rather than the content. If these text files originate from multiple edited sources, such as from Project Gutenberg and JSTOR, some topics may become indicative of the source repository, rather than the semantics of the texts themselves. 

\subsection{Model Comparison}
Model comparison and model selection can be conflated. Model selection presumes a selection of a ``better'' model, and often uses quantitative measures such as perplexity and log likelihood to pick the ``correct'' model. Model comparison contrasts two models without any judgment on which is the ``better'' model, rather focusing on how two alternatives relate to each other. Quantitative measures for model comparison are dependent upon the two models being compared. In contrast, model selection measures can be independently computed for each model.

Typically, two models are compared by applying a distance metric to corresponding rows in the word-topic ($\phi$) and/or topic-document ($\theta$) matricies. Jenson-Shannon distance is the most frequently used distance measure (Section~\ref{sec:jsd}). The problem of discovering corresponding rows (topics) is known as \emph{topic alignment}. Topic alignment consists of two problems: vocabulary merging and pair alignment. Topics can be aligned across models of different numbers of topics ($k$), but these alignments are not guaranteed to be symmetrical. Usually, topic alignment is constrained such that $k_0 \geq k_1$. Depending on the strategy used, not every topic in $\mathcal{M}_1$ will be used in the alignment of $\mathcal{M}_0$ (i.e., the alignment function is injective).

Once an alignment is in hand, the model distance can be calculated as the average or cumulative distance of each alignment pair.

\subsubsection{Vocabulary Merging}
\label{sec:vocabulary-merging}
We start with comparisons of the word-topic matrices. First, a vocabulary merging strategy must be selected. Some approaches are:

\begin{itemize}
\item Reduce the matrix to the terms shared by both models. Do not re-normalize probability distributions.
\item Reduce the matrix to the terms shared by both models. Re-normalize word-topic probability distributions.
\item Expand each matrix such that all terms are shared by both models. Introduce $\epsilon \leq \frac{1}{\mathrm{len}(V_0 \cup V_1)}$, a small value to initialize in the new word columns for each matrix. Re-normalize word-topic probability distributions. (This avoids asymptotic behavior in several multi-dimensional distance measures when 0s occur in the vector.)
\end{itemize}

\subsubsection{Topic Alignment}
Then, select a strategy for topic alignment.

\begin{itemize}
\item \textbf{Naive Alignment} --- Each topic in $\mathcal{M}_0$ is paired with the closest topic in $\mathcal{M}_1$. This strategy may lead to multiple topics in $\mathcal{M}_0$ mapping to the same topic in $\mathcal{M}_1$. Due to this non-injectivity, naive alignment is not guaranteed to be symmetric: $\mathrm{align}(\mathcal{M}_0, \mathcal{M}_1) \neq \mathrm{align}(\mathcal{M}_1, \mathcal{M}_0)$.
\item \textbf{Basic Alignment} --- To ensure each topic matches only one topic in the corresponding model, the number of topics in the model being mapped to must be greater than equal to the number of topics in the first model. We use a round-robin strategy: the distance from each topic to all other topics is stored. The closest topic in $\mathcal{M}_0$ for topic 1 in ${M}_1$ is selected and removed from all other distance lists. Topics are assigned in this manner until all topics are assigned.
\item \textbf{Adversarial Alignment} --- Topics can be aligned using an evolutionary strategy. Topics are aligned randomly, with alignments having the smallest model distance persisting to the next round of comparison. Iterate until convergence.
\end{itemize}

\subsection{Query Sampling}
\label{sec:query-sampling}
Introducing new documents to an already existing topic model is not as straightforward as fitting data to other types of models. There are many concerns, including endemic feature mismatch between the old corpus and the new --- the likelihood that two models use exactly the same terms is unlikely without stoplisting. This problem is particularly tricky if the whole corpus is not known ahead of time, so the vocabulary cannot be pre-determined. This problem generalizes to the out-of-sample problem in dimensionality reduction: how do you take an object that was not in the initial training set and project it into the model space \citep{Trosset2008}.

Query sampling is one way to find a topic distribution for new documents under an existing model. $\mathcal{C}_1 \oplus \mathcal{M}_0$ is the application of query sampling based on $\mathcal{M}_0$ to the corpus $\mathcal{C}_1$.

\begin{enumerate}
    \item Remove all terms from $\mathcal{C}_1$ not in $\mathcal{C}_0$ so the documents share a vocabulary.
    \item Initalize a new model, preserving the word-topic matrix ($\phi$), but with a new topic-document matrix ($\theta$) containing only the new documents.
    \item Train the model, using the standard LDA algorithm, yielding $\mathcal{M}_0^1$.
\end{enumerate}

Query sampling has some peculiarities. First, the word-topic matrix ($\phi$) is allowed to change on each iteration as the documents are fit to the terms. This means that the more iterations are run, the more likely topics will diverge from their initial definition. This \emph{topic drift} can be measured through model distance, in the same way that two separate models would be compared. Second, the topic assignments for $\mathcal{C}_0$ will be from different topics than those for $\mathcal{C}_1$. This is because in query sampling, only the new documents are fit to the model. The topic assignments of old documents do not change in reaction to the changes in $\phi$.

There are three potential solutions to whether new documents should alter the representation of the old ones. The first is to reset $\phi$ on each iteration to the initial value from $\mathcal{M}_0$. The primary disadvantage of locking $\phi$ is that it breaks the joint inference of topics and documents, so the model is no longer an LDA topic model. The advantage of this method is that it further increases the computational speed by not recomputing $\phi$ at each step. The second solution is to include the original $\theta$ and just extend it for the new documents, first initializing just the new rows using the initial $\phi$. The primary disadvantage to this method is that it increases computational time. The primary advantage is that it preserves the joint inference of documents and topics for both matrices. However, the third solution is to embrace the change in $\phi$ as a part of the experimental design. Under this design, the change moving from one set of documents to another can be measured. Further exploration of whether models trained on opposite portions, then query sampled on the remaining portion, could be explored for symmetry (i.e., $\mathcal{C}_1 \oplus \mathcal{M}_0 \stackrel{?}{=} \mathcal{C}_0 \oplus \mathcal{M}_1$). These findings could have important consequences for belief revision based on initial exposure.

\section{Summary}
In this chapter, we introduced LDA topic modeling, which is how we operationalize an information environment of textual data for our studies. The LDA model has a geometric interpretation compatible with information theoretic measures. Each document is a source, drawing tokens from a topic distribution to create a signal that can be analyzed with information theoretic measures, which we explore in the next chapter. We highlighted the emphasis on qualitative evaluation strategies for LDA models, and discuss validation through user studies. Then, we explored parameter selection and corpus preparation techniques. Finally, several experimental designs and model training scenarios involving multi-source documents were proposed. This included query sampling, a novel technique for training new documents, and the impact of various modeling decisions regarding the word-topic distributions when adding an out-of-sample document. 

\chapter{Methods: Information Theory}
\label{chapter:info-theory}

This chapter presents a brief introduction to information theory, beginning with the communication model and entropy measure. It then presents the Kullback-Leibler (KL) divergence, which is used in most of our case studies to measure surprise in reading a new text. We also present the Jensen-Shannon distance, which is used to measure distance between pairs of texts. 

\citet{Shannon1949}'s seminal paper ``A Mathematical Theory of Communication'' forms the basis of information theory. It directly addresses the engineering concerns, rather than the semantic concerns, by simplifying the notion of a message. A message is selected from a set of possible messages. The communicative system is designed to operate based on each message, and both senders and receivers are aware of the same set of messages. For example, in written English, messages are communicated through the alphabet, which defines the symbol space. Both the writer and the reader only expect characters drawn from the 26 letters (plus, perhaps, some punctuation). How this message of symbols is encoded may differ based on the medium used: morse code in a telegraph, ASCII or Unicode in a modern computational system. These encoded messages are the signal. It is then decoded by the receiver and reconstructed into the original message by the destination. Information theoretic measures quantify the amount of uncertainty in a signal (entropy) and can be used to optimize the encoding of a particular signal (coding theory).

LDA operates over probability distributions amenable to analysis by information theory, and we join the bandwagon of practitioners possibly abusing the communication model, but making strides nevertheless \citep{shannon_bandwagon_1956}.



\section{Entropy}
Information entropy is the central measure in information theory. It quantifies the amount of uncertainty in a random variable, known as the signal. Entropy is measured in bits, and given by the function:

\[
H(p) = -\sum_i p_i\log_2 p_i
\]

\noindent where $p_i$ is the probability of the $i$-th symbol. For example, a simple coin flip with 50-50 odds of heads-or-tails has an entropy of 1 bit. If a weighted coin with 100\% odds of being tails is given, and the weighted nature is known by the observer, the entropy would be 0 bits---there is no uncertainty involved in the flip. For each distribution, the maximum bits of entropy are $\log_2 |\bar p|$ bits, where $|\bar p|$ is the length of the vector.

One way to conceptualize of entropy is the average number of yes-or-no questions it would take to determine the outcome of a random process. Fractions of a bit come from these averages. Another conception of entropy is as the absolute limit on the encoding efficiency for a particular distribution.

In an LDA model, entropy for a given document is calculated at the topic level, in determining the uncertainty about which topic a term came from. This quantity is:

\[
H(\theta_d) = \sum_t p(\theta_{dt}) \log_2 p(\theta_{dt})
\]

\section{Kullback-Leibler Divergence}
\label{sec:kl-divergence}
For most of our purposes, we wish to compare two documents. In LDA, each document is represented as a probability distribution over topics. The reader is modeled in an information-theoretic fashion as building efficient mental representations of these distributions. Our task is then to quantify his exploratory behavior as he moves from one distribution to another through reading.

The Kullback-Leibler (KL) divergence addresses the problem of discriminating between two statistical populations \citep{Kullback1951}.  An agent expecting observations to be drawn from probability distribution $\vec{p}$ will have those expectations violated if it arrives according to a different distribution $\vec{q}$. It is defined by
\[
D_\mathrm{KL}(\vec{q} | \vec{p}) = \sum_{i=1}^k q_i \log_2 \frac{q_i}{p_i}
\]

\noindent where $\vec{p}$ is the distribution over topics that the reader has encountered before, and $\vec{q}$ the new distribution of topics that the reader encounters next.

\subsection{Understanding KL Divergence through ``Twenty Questions''}
To understand KL divergence, consider the children's game ``Twenty Questions''\footnote{This example owes much credit to the informal explanation of \citet{dedeo}.}: one player (the ``parent'') chooses a noun at random. The other player (the ``child'') then attempts to guess the noun by asking the parent yes-or-no questions. Success is reached when the child gets the answer right in less than 20 questions. Getting the answer in as few questions as possible is considered a better outcome.

One strategy is for the child to guess nouns at random: ``is it a car?'' ``is it a tree?'' ``is it a frog?'' But there are better strategies: one can ask discriminatory questions that reduce the possible answers quickly: ``is it something that can be dead or alive?'' ``is it an animal?'' ``is it a vehicle?'' This strategy is a much more efficient encoding of the problem space.

However, parents are not random-word generators, and they may be somewhat biased to particular terms. For example, a parent may choose the word ``car'' 90\% of the time, so despite not narrowing down the possibilities in general, a child playing with this parent could be well-served by asking ``is it a car?'' first. This distribution of word choices can be represented as a probability distribution $\vec{p}$. The information entropy, defined above, then reveals the average number of yes/no questions the child must ask using the best possible script. In particular, the average number of questions $Q$ that the child has to ask, if she is using an optimal script, lies somewhere between
\[
H(\vec{p}) \leq Q \leq H(\vec{p})+1,
\]
\noindent a result known as Shannon's source coding theorem \citep{Shannon1949}.

Let us now imagine that the child has developed an optimal script for one of her parents. What happens when she uses this script playing against the other?

If the two patterns have similar patterns of noun choice, the child may do well. However, if the patterns differ, then the script could fail very badly. In this conception, the KL divergence measures the number of ``excess questions'' asked of the second parent, when the child has optimized for the word choices of the first parent. An example is developed in Figure~\ref{fig:KL-example}.

\begin{figure}
    \centering
    \subfloat[Optimal script (encoding) for parent 1]{
    \qquad
    \begin{tikzpicture}[node distance=2cm]
    \node (car-q) [] {\textbf{Is it a car?}};
    \node (car) [below left of=car-q] {Yes (car)};
    \node (boat-q) [below right of=car-q] {\textbf{Is it a boat?}};
    \node (boat) [below left of=boat-q] {Yes (boat)};
    \node (duck) [below right of=boat-q] {No (duck)};

    \draw [arrow] (car-q) -- (car);
    \draw [arrow] (car-q) -- (boat-q);
    \draw [arrow] (boat-q) -- (boat);
    \draw [arrow] (boat-q) -- (duck);
    \end{tikzpicture}
    \qquad
    }
    
    \subfloat[Parent word probabilities]{
    \qquad
    \begin{tabular}{l|ccc}
                 & p(car) & p(duck) & p(boat) \\
        Parent 1 & 1/2    & 1/4     & 1/4     \\
        Parent 2 & 1/4    & 1/2     & 1/4     \\
    \end{tabular}{}
    \qquad
    }
    
    \caption{\textbf{An example of KL divergence}. The table above shows the word distributions for two parents in a game of 20 Questions, limited to a very simple vocabulary of 3 words. Each parent's distribution has an entropy of 1.5 bits, meaning on average, it will take a child with an optimal script 1.5 questions to get to the noun. Using the optimal encoding for parent 1 with parent 2's distribution will result in an average of 1.75 questions. The KL divergence in this case is .25 bits (or ``excess questions'').}
    \label{fig:KL-example}
\end{figure}

\subsection{KL and Belief Revision (or ``Cognitive Surprise'')}
A second interpretation of KL divergence comes from Bayesian statistics. Imagine an agent observing a process and trying to decide whether her observations are drawn from $\vec{p}$ or $\vec{q}$, when her priors are equally split between the two possibilities. Given a single observation, of type $i$, the relative log-likelihood of distribution $\vec{q}$ compared to distribution $\vec{p}$ is just
\begin{equation}
\Delta\mathcal{L}=\log{\frac{q_i}{p_i}}.
\label{stepone}
\end{equation}
If the true distribution is $\vec{q}$, the average rate at which this relative log-likelihood increases is simply
\begin{equation}
\sum_{i=1}^N q_i \Delta\mathcal{L}.
\label{steptwo}
\end{equation}
On substituting Eq.~\ref{stepone} into Eq.~\ref{steptwo}, we recover the KL divergence equation. This provides a second interpretation for KL: the rate at which log-evidence for the true distribution accumulates over time. When the new distribution is very different from the one that came before (``more surprising''), evidence for that difference will accumulate quickly.

KL divergence has been experimentally validated as a measure of cognitive surprise in many contexts. Vision researchers have used KL to track surprise in a visual scene: the places on the screen where new events most violate the viewer's previous assumptions; KL then accurately captures attention attractors in visual search tasks \citep{Demberg2008193,itti2009bayesian}. More generally, KL has seen wide use in the cognitive sciences; \citet{resnik1993selection}, for example, proposed the KL divergence as a measure of selectional preferences in language (reviewed in \citet{light2002statistical}). It has found use in many successful models of linguistic discrimination, including syntactic comprehension \citep{hale2001probabilistic,Levy20081126}, speech recognition \citep{COGS:COGS12167,COGS:COGS1267} and word sense disambiguation \citep{Resnik1997}. In computational social science, it captures dynamics previously explained through qualitative discourse analysis \citep{barron2018individuals}.

\subsection{Text-to-text and Text-to-past Surprise}
We use KL in two distinct ways for this study. We measure the text-to-text surprise: given a distribution over topics for the text a reader just read, how surprised are they upon encountering the next volume's topic distribution? Text-to-text surprise is a \emph{local} measure. We also measure the text-to-past surprise: given all of the volumes that a reader has encountered so far, how surprised are they by the text that comes next? Text-to-past surprise is a \emph{global} measure. 

Formally, these correspond to two choices for $\vec{p}$: 1) the distribution over topics for the just previous book, and 2) the average over all books in the reader's past. Meanwhile, $\vec{q}$ will always refer to a distribution over topics for the next book a reader encounters. If we define $\theta_i$ as the topic distribution for document $i$, then text-to-text ($T2T$) and text-to-past ($T2P$) surprise are defined as
\[
T2T(i) = D_{KL}(\theta_{i}|\theta_{i-1}),
\]
\[
T2P(i) = D_{KL}\left(\theta_i\left| \frac{\sum_{j=0}^{i-1} \theta_j}{i}\right.\right).
\]

Interpretively, text-to-text surprise and text-to-past surprise provide complementary windows onto the decision-making process of what to read next. Local decision-making, meaning the choice of the next text to read given the current one, is captured by text-to-text surprise. Global decision-making, the choice of which text to read given the entire history of reading to date, is captured by text-to-past surprise. Low surprise, in either case, is a signal of \emph{exploitation}, while high surprise indicates larger jumps to lesser-known topics, and thus of \emph{exploration}. These measures can be easily generalized to arbitrary text-to-$N$ surprise measures, representing the choice of the next reading given the history of readings within the past $N$ volumes or time periods. 

\subsection{Asymmetry and Enclosure}
Sometimes, two documents are related in the way that an abstract is related to a full article. Typically, the content of an abstract should be more predictable (less surprising) given the article, than the full document will be given its abstract. This kind of relationship can also hold more generally between two documents when one is more comprehensive than the other, with the first document $A$ covering all or most of the topics covered in $B$, but not vice versa.

KL divergence can be used to expose this measure due to its asymmetry. It is worth emphasizing that KL divergence is \emph{not} a measure of distance, but rather a way to quantify the ordered processing that happens as learning unfolds over time. Asymmetric measures are useful in many contexts. For example, travel time is an asymmetric measure while distance in miles is a symmetric measure. In the United States flight times west-to-east are almost always shorter than flight times east-to-west for the same routes, due to increased jet stream headwinds in east-to-west travel.

We propose this asymmetry measure as \emph{enclosure}, and use it extensively in Chapter~\ref{chapter:writings}. For two topic distributions $\vec{q}$ and $\vec{p}$, if $\operatorname{KL}(\vec{q}|\vec{p}) < \operatorname{KL}(\vec{p}|\vec{q})$, then $\vec{p}$ \emph{encloses} $\vec{q}$. This builds on the intuition that if one builds a script for a $\vec{p}$ biased towards some subset of topics, but encounters a more uniform $\vec{q}$, the coding will fail worse than if $\vec{p}$ were uniform and $\vec{q}$ highly biased.

\section{Jenson-Shannon Distance}
\label{sec:jsd}

For comparisons where the order of the texts or the relative comprehensiveness of texts is not relevant to the analysis, a symmetric measure of the distance between  documents  is more appropriate. For such purposes, we use the Jensen-Shannon distance, a symmetrized and normalized version of KL divergence that respects all the mathematical properties of a proper distance metric \autocite{Lin1991,nielsen2010family,fuglede2004jensen}. 

This measure is first derived as the Jensen-Shanon \emph{divergence}, which is the symmetrized and smoothed version of the KL divergence:
\[
\operatorname{JS}_{div}(\vec{p}, \vec{q}) = \frac{1}{2}\operatorname{KL}(\vec{p}|M) + \frac{1}{2}\operatorname{KL}(\vec{q}|M)
\]
\noindent where $M = \frac{1}{2}(\vec{p} + \vec{q})$.

This quantity does not respect the triangle inequality, meaning that it is not a proper metric \citep{Endres2003}. By taking the square root of the JS divergence, we derive the JS \emph{distance} (hereafter, JSD), which is a true metric \citep{fuglede2004jensen}:

\[
\operatorname{JSD}(\vec{p}, \vec{q}) = \sqrt{\operatorname{JS}_{div}(\vec{p}, \vec{q})}
\]

\section{Bayesian Epoch Estimation}
In the foraging literature, individuals are often assumed to persist in sustained periods of either exploration or exploitation. We call this an \emph{epoch}. We are particularly interested in whether or not these epochs align with important events in a person's life. Our model, Bayesian epoch estimation (BEE), infers epoch breaks from measurements of KL divergence in an ordered dataset. With these automatically-discovered epoch breaks, we can determine whether the data support a qualitative interpretation of the quantitative model.

BEE models an epoch as a Gaussian distribution of KL divergence, in either the text-to-text or text-to-past case, with fixed mean and variance. Each epoch is defined by a beginning point (which is also either the end of the previous epoch, or the start of the data), an average level of surprise, and the variance around that average. For each time-series, the model then contains $3n-1$ parameters, where $n$ is the number of epochs.

Epoch switches are independently selected for the text-to-text and text-to-past measures. Each transition can then be interpreted as a change in the reader's exploration and exploitation behavior at the local or global level. When a new epoch has higher average surprise than the one before, for example, we can understand the reader as moving to a more exploration-based strategy.

Our one externally-set parameter is the total number of epochs, $n$. As $n$ rises, it becomes easier and easier to fit the data; at some point, we encounter the over-fitting problem, and in the extreme case we have as many parameters as we have data points. For each choice of $n$, our model returns a likelihood: the probability that the observed data were generated by the (best fitting) choice of parameters for that model. As $n$ rises, so does the log-likelihood.

Our generative model for Bayesian epoch estimation has $3n - 1$ parameters. There are $(n-1)$ parameters to describe the end points of the first $n-1$ epochs, and $2n$ parameters to describe the mean and variance of the text-to-text (or text-to-past) surprise within each epoch. We estimate these parameters using an approximate maximum-likelihood procedure. Within each epoch $i$, we assume the surprise is constant and Gaussian distributed with a particular mean $\mu_i$ and variance $\sigma^2_i$. We write the $3n-1$ parameters as a vector $\vec{v}$; then the distribution over $\vec{v}$ given the data $s$, equal to a list of surprises, $\{s_i\}$, is
\begin{equation}
\log P(\vec{v}|s) = \log P(s|\vec{v}) + C = -\sum^n_{i=1} \frac{(e_{i+1}-e_i-1)}{2}\left(1+\ln(2\pi \hat{\sigma^2}_i) \right) + C,
\end{equation}
where $e_i$ is the start point of epoch $i$ and $C$ depends on the prior. The start point of the first epoch, $e_1$, is fixed to be volume zero; given our conventions, $e_{n+1}$ is fixed to be the final volume plus one. The sigma estimator, $\hat{\sigma^2}_i$, is the standard maximum likelihood estimator of the variance,
\begin{equation}
\hat{\sigma^2}_i = \frac{1}{e_{i+1}-e_i-1}\sum^{e_{i+1}-1}_{k=e_i} \left(s_k-\hat{\mu}_i\right)^2,
\end{equation}
and $\hat{\mu}_i$ is defined as
\begin{equation}
\hat{\mu}_i =  \frac{1}{e_{i+1}-e_i-1}\sum^{e_{i+1}-1}_{k=e_i} s_k.
\end{equation}
To do Fisher maximum-likelihood estimation, we ignore the effect of the prior $P(\vec{v})$ on the maximum; equivalently, we do maximum a posteriori estimation and assume that $P(\vec{v})$ is flat over the region of interest.

\subsection{Epoch Model Selection}
\label{sec:aic}

To determine the best-fit number of epochs, we use a simple model-complexity penalty, Akaike Information Criterion (AIC) \citep{Akaike1974}, to verify that the selected model is preferred, despite the addition of new parameters. The AIC penalizes a model's log-likelihood by the total number of parameters in the model; adding more epochs, in other words, must justify itself by a sufficiently large increase in goodness of fit. The AIC can be understood as an information-theoretic and Bayesian version of the chi-squared test, which attempts to maximize a model's predictive power \citep{burnham2003model}.

For the two- and three-epoch models we compare this likelihood to the a single-epoch null model of 2 parameters - mean and variance for text-to-text or text-to-past surprise over the whole data-set. Our 2-epoch model has 5 parameters. A 3-epoch model has 8 parameters.

\section{Summary}
In this chapter, we linked the topic models introduced in the previous chapter to the Shannon communication model, theoretically justifying their analysis using information theoretic measures. After a brief primer on information entropy, we examined the Kullback-Liebler divergence, which is a cognitively-validated measure of surprise. The asymmetries of KL divergence allow us to make claims about \emph{enclosure}, a test for relative comprehensiveness of two texts. The use of KL asymmetries for document comparison in this manner is also a novel contribution. In situations where a symmetric measure is necessary, we use the Jensen-Shannon distance. Finally, we introduced the Bayesian epoch estimation (BEE) model as a novel method to partition ordered data, such as sequences of KL divergence.

\blankpage
\chapter{Case Study: Darwin's Reading Behavior}
\label{chapter:readings}

As one of the most successful and celebrated scientists of the modern era, Charles Darwin's scientific creativity has been the subject of numerous narrative and qualitative studies \citep{Gruber1974,Johnson2010,VanHulle2014}. In part, these studies are possible because Darwin left his biographers careful records of his intellectual and personal life. These include records of the books he read from 1837 to 1860, a critical period which culminated in the publication of \emph{The Origin of Species}. Table~\ref{table-timeline} summarizes key events in Darwin's life.

\begin{table}[t]
\begin{center}
\textbf{Major Events in Charles Darwin's Life (1809-1882)}
\begin{tabular}{rl}
\hline 12 Feb 1809 & Born in Shrewsbury, England \\
22 Oct 1825 & Matriculates at University of Edinburgh \\
15 Oct 1827 & Admitted to Christ's College, Cambridge \\
27 Dec 1831 & Departs England aboard the \emph{HMS Beagle} \\
2 Oct 1836 & Return to England aboard the \emph{HMS Beagle} \\ \hline
\textbf{July 1837} & \textbf{First entries in reading notebooks} \\
Aug 1839 & Publication of \emph{The Voyage of the Beagle} (1st edition) \\
May 1842 & Writes the 1st Essay on Species \\
4 July 1844 & Writes the 2nd Essay on Species \\
Aug 1845 & Publication of \emph{The Voyage of the Beagle} (2nd edition) \\
1 Oct 1846 & Begins barnacle project \\
19 Feb 1851 & Publishes first volume of barnacle work \\
9 Sep 1854 & Begins sorting notes on natural selection \\
14 May 1856 & Starts writing ``large work'' on species \\
24 Nov 1859 & Publication of \emph{The Origin of Species} (1st edition)\\
\textbf{13 May 1860} & \textbf{Last entry in reading notebooks} \\ \hline
24 Feb 1871 & Publication of \emph{The Descent of Man} \\
19 Feb 1872 & Publication of \emph{The Origin of Species} (6th and final edition) \\
21 Apr 1882 & Dies at Down House in Kent, England \\ \hline
\end{tabular}
\end{center}
\caption{{\bf Timeline of Major events in Charles Darwin's life}. This includes those marked on Figure~\ref{fig:kl_over_time}. This paper focuses on the critical period of his work from 1837 to 1860, leading to the publication of \emph{The Origin of Species}. See \citet{berra2009charles} for an expanded chronology.}
\label{table-timeline}
\end{table}

This chapter presents the first quantitative analysis of an important scientist's reading diaries, tracking how Darwin navigated the exploration-exploitation trade-off in choosing what to read\footnote{These results were first presented in \citet{murdock_exploration_2017} and are expanded here with material on corpus curation and a revised discussion.}. We link Darwin's reading records with the full text of the original volumes. We then use probabilistic topic models (see Chapter~\ref{chapter:topic-modeling}) to represent the original text of each book Darwin read as a mixture of topics. We use information theory to measure the surprise, or unpredictability, of the next book that Darwin chose to read, compared to his past history of reading.

We present three key findings: 
\begin{enumerate}
\item Darwin's reading patterns switch between both exploitation and exploration throughout his career. This is in contrast to a pure surprise-minimization strategy that consistently exploits content within a local region before moving on. The general trend, as Darwin's career develops, is towards increasing exploration.
\item In comparison to the publication order of the texts Darwin read, Darwin's reading order shows higher average surprise. This indicates that the order in which the books were written by the scientific community is less surprising than the order in which Darwin read them.
\item Darwin's strategies fall into three long-term epochs, or behavioral modes characterized by distinct patterns of surprise-seeking. These epochs correspond to three biographically significant periods: Darwin's post-\emph{Beagle} studies, his extensive work on barnacles, and a final period leading to his synthesis of natural selection in the \emph{Origin of Species}.
\end{enumerate}

Our case study approach contrasts with previous uses of topic modeling to analyze the large-scale structure of scientific disciplines \citep{Griffiths2004,Hall2008,Blei2007,Cohen2015} and the humanities \citep{Poetics41, JDH, Jockers2013}. Previous models of historical records have focused on language use as an indication of larger shifts in style \citep{Hughes2012,Underwood2012}, learnability \citep{hills2015recent}, or content \citep{Michel2011,Goldstone2014,Klingenstein2014} of significant portions of publications in a field, including a study of \emph{Cognition} itself \citep{Cohen2015}. 

These works model the collective state of all published works at a particular date, but obscure the role of individual foraging behavior. By focusing on a single individual for whom ample records exist, we gain access to what \citet{Tria2014} describe as ``the interplay between individual and collective phenomena where innovation takes place''.

\section{Dataset Curation}
\label{sec:notebook-corpus}
Darwin was a meticulous record-keeper---starting in April 1838, he kept a notebook of ``books to be read'' and ``books read''. These records span the 23 years from 1837 to 1860, tracking his reading choices from just after his return to England aboard the \emph{HMS Beagle} to just after the publication of \emph{The Origin of Species}. We located the full-text of 665 of the 687 (96.7\%) English non-fiction titles mentioned in these reading notebooks. These full-text versions were found in a variety of digital libraries: Project Gutenberg, HathiTrust, and the Internet Archive.

\begin{figure}
    \centering
    \includegraphics[width=0.5\textwidth]{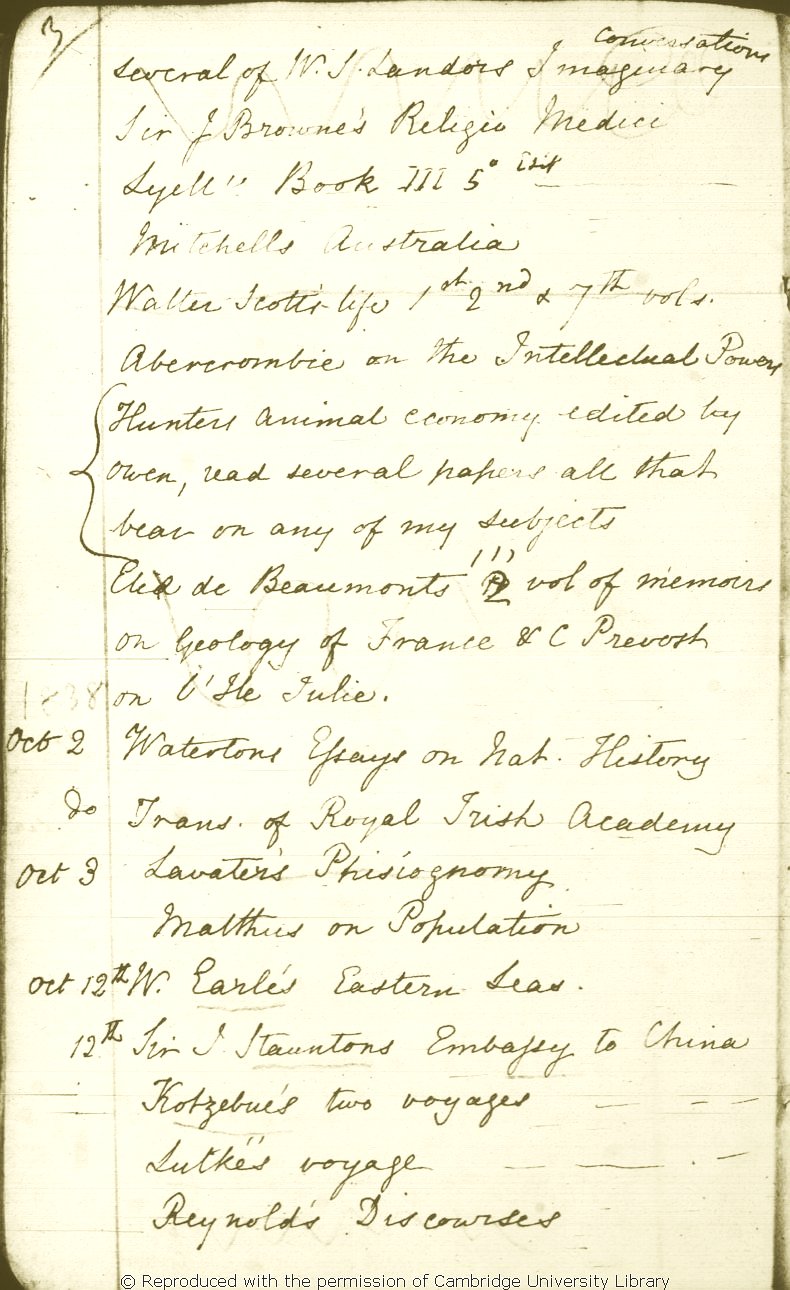}
    \caption{{\bf Page from Darwin's reading notebooks}. Page 3a of Darwin's first notebook (DAR 119), during which he began to track the exact reading dates. Note the reading of Malthus's \emph{On Population} on October 3, 1838. Photo courtesy of Cambridge University Libraries.}
    \label{fig:notebook-page}
\end{figure}

Despite our use of digital libraries, the notebooks and books they refer to are originally physical artifacts (see Figure~\ref{fig:notebook-page} for a notebook page). We use transcriptions of the reading notebooks and automatically-generated transcriptions of the books that utilize optical character recognition (OCR) technologies, both of which may introduce information loss. Additionally, Victorian publishing practices often spread a single title over multiple volumes for portability and ease of use. In this paper, we use \emph{volume} to refer to each physical artifact. Each individual entry of Darwin's notebooks is referred to as a \emph{title}. In the case of books, a \emph{title} gathers together one or more volumes. A \emph{title} is roughly equivalent to a \emph{catalog record} in a traditional library.

{}

While Darwin also read French, German, and Latin texts, we reduced the corpus to English-only to reduce cross-linguistic effects in the model \citep{BoydGraber2009}. Additionally, we focused only on non-fiction texts. An examination of the influence of fiction on Darwin is a topic for further exploration.

There are 647 catalog records and 1057 volumes corresponding to the 665 titles modeled in this study. Some volumes in the corpus alignment were unable to be matched to the exact edition listed by the Darwin Correspondence Project, and thus there is occasionally a difference between the volume Darwin read and the volume whose text we use for topic modeling. Table~\ref{table-corpus-stats} shows the summary of the items which were located and remain missing.

\begin{table}[h!]
\begin{center}
\begin{tabular}{l|rr|r}
& Located & Non-located & Total \\ \hline
\textbf{Total} & \textbf{811} & \textbf{104} & \textbf{915} \\
- Fiction & - 79 & - 1 & - 80  \\
- Non-English & - 63 & - 85 & - 148 \\ \hline
\textbf{English Non-fiction} & \underline{\textbf{665}} & \textbf{22} & \textbf{687} \\ \hline
\end{tabular}
\end{center}
\caption{{\bf Corpus Composition}. Composition of the Reading List in terms of fiction, non-fiction, English, and non-English texts. Located titles refers to the number identified in the HathiTrust (\url{http://hathitrust.org/}), Internet Archive (\url{http://archive.org/}), and Project Gutenberg  (\url{http://gutenberg.org/}). Non-located texts were unavailable in the HathiTrust, Internet Archive, or Project Gutenberg as of December 1, 2015.}
\label{table-corpus-stats}
\end{table}

Our publication dates are those listed by the Darwin Correspondence Project (DCP); the DCP uses the publication date of the volume, if found in Darwin's library, otherwise the date of first publication. The reading order is determined by dates listed in the reading notebook. When multiple titles are listed at a particular date, we use their natural ordering in the notebooks --- titles written at the top of the page are assumed to be read before those at the bottom.

We use the InPhO Topic Explorer \citep{Murdock2015} for tokenization and modeling of texts. First, plain-text editions downloaded from the HathiTrust, the Internet Archive, and Project Gutenberg are normalized by merging cross-line hyphens into single words, normalizing into ASCII using the Python library Unidecode, removing all words containing punctuation and numerals (often due to OCR errors), and lower-casing all words. Then, words appearing in the English stopwords corpus from the Natural Language Toolkit (NLTK) \citep{nltk} are removed. Finally, words occurring less than 30 and more than 15,000 times were excluded from the corpus. After this pre-processing, the corpus consisted of 40,822,136 tokens drawn from 77,611 unique tokens. We made no attempt to apply stemming or clean up OCR errors, other than the filtering of words occurring fewer than 30 times. Further justification of our decision not to use stemming and to use frequency-based stoplists can be found in Chapter~\ref{chapter:topic-modeling}.

\begin{figure}
\begin{center}
\includegraphics[width=0.95\textwidth]{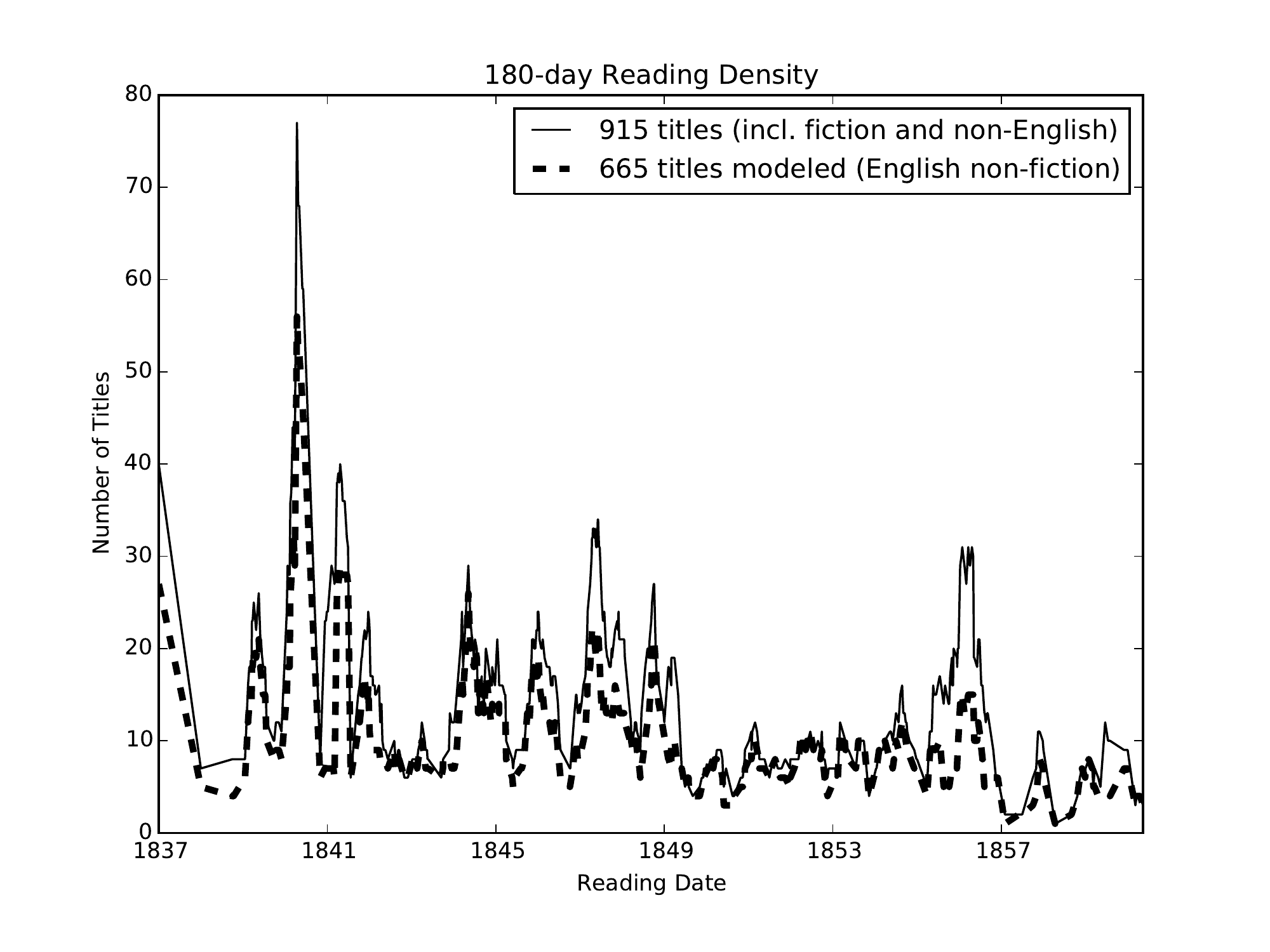}
\end{center}
\caption{\textbf{Reading density of Darwin's reading notebooks, smoothed over a 6-month window}. The dashed line shows the 665 titles here modeled, while the thin solid line represents all 915 titles in the reading notebooks.}
\label{fig:reading-density}
\end{figure}

Figure~\ref{fig:reading-density} shows the density of Darwin's readings modeled. Notice the large jump in 1840 corresponds to a period when he was reading entire series of journals, each article of which was a separate title in his notebook. Also, note that Figure~\ref{fig:reading-density} shows both the density of the selection modeled and the entire reading notebook list.

\begin{figure}
\begin{center}
\includegraphics[width=\textwidth]{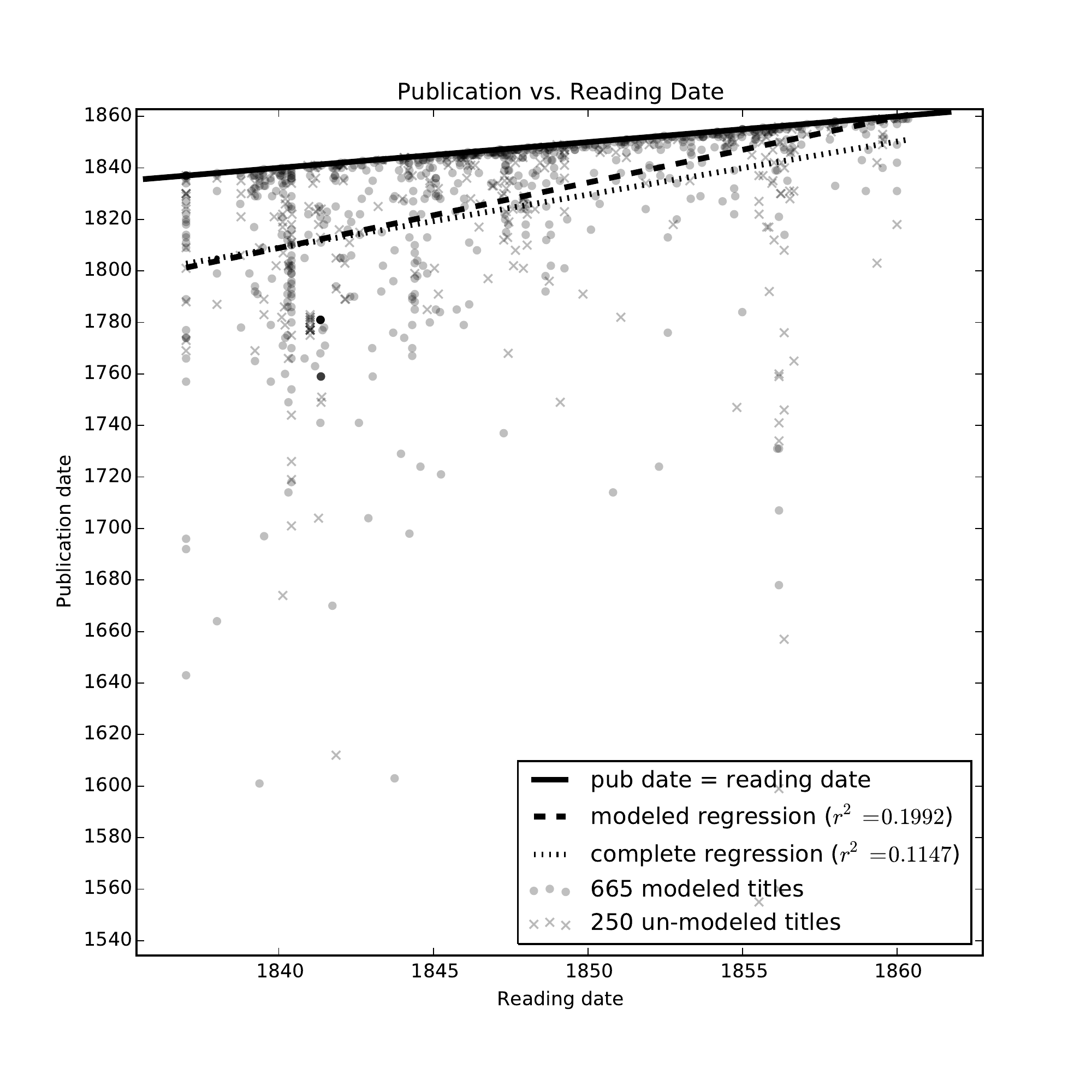}
\end{center}

\caption{\textbf{Publication and reading dates}. Scatter plot of the publication and reading dates of the titles in Darwin's reading list. The 665 modeled titles are shown with dots, while the remaining 250 titles are shown as \texttt{x}s.  The solid line indicates when the reading date and publication date are equal. The dashed line indicates a linear regression over the dots ($r^2 = 0.1992$), and the dotted line indicates a linear regression over the dots and \texttt{x}s combined ($r^2 = 0.1147$). The appearance of older materials in 1856-57 corresponds to Darwin's literature review of pigeon breeding, conducted as a case study in artificial selection and included in \emph{The Origin of Species}.}
\label{fig:read-pub-dates}
\end{figure}

Figure~\ref{fig:read-pub-dates} indicates that as Darwin's readings progress he begins reading more recently published, contemporary sources. We also show a linear regression for the un-modeled texts, showing that his total reading also progressed toward contemporary sources, although at a slower rate.

\section{Methods}
While the methods are summarized in earlier chapters, a brief review here is helpful. We quantify the texts probabilistic topic modeling (see Chapter~\ref{chapter:topic-modeling}) for 500 iterations per model. To test the robustness of our results, we vary the number of topics, $k$. In this chapter, we report for $k=80$ because we found that our results were robust to alternative values of $k$ and to differing random seeds.

In order to analyze Darwin's path through the texts we quantify the divergence between topic distributions using Kullback-Leibler divergence (KL; see Chapter~\ref{sec:kl-divergence}). We use KL in two ways: the text-to-text divergence and the past-to-text divergence. In the text-to-text divergence, we quantify the divergence between two successive texts. In the past-to-text divergence, we quantify the divergence between a text and all texts that came before it. Increasing divergence indicates a period of exploration. Decreasing divergences indicates a period of exploitation. We refer to trends in text-to-text divergence as \emph{local} exploration/exploitation and trends in past-to-text divergence as \emph{global} exploration/exploitation. Throughout the chapter, we plot the cumulative divergence to emphasize the slope of the data and highlight periods of exploitation/exploration.

We also examine how Darwin selectively re-orders the products of his culture as he selects which book to read. We compare the order in which Darwin read the books to the order in which they were published. We can then ask whether Darwin's readings were more or less exploratory than the order in which they were produced.  Does Darwin's reading order reduce the surprise relative to the publication order, or does it increase it?

\subsection{Null Reading Models}
All results are relative to a null reading model that holds Darwin's original reading dates fixed and re-samples without replacement from his original reading list. The title selection at each reading date is constrained to those titles published before that date. In contrast to a null that includes permutations that neglect publication date, this restricted null captures the dynamics of publication in which a new work can unexpectedly change the information space.

We compare the production and consumption of texts by considering only the texts Darwin recorded himself as reading. This allows us to make a direct comparison between the average surprise of the publication order and reading order within that set. But the production of these texts, of course, occurs in a much larger context, and it is reasonable to ask about the books that Darwin considered reading but did not, or an even more complete representation of the state of Victorian science constructed using all the scientific books available to him in Kent and London during these years. Here we focus on the books he chose to read, and the space he actually explored, partly for practical reasons and partly because we are interested in the decisions he made to order the texts, rather than the decision of whether to read them at all. That question could be explored through investigation of his ``books to be read'' notebooks, but is left for Future Work (Chapter~\ref{chapter:future}).

\section{Text-to-text and text-to-past surprise}

We characterize Darwin's decision process by the combination of text-to-text and text-to-past surprise. Exploration, indicated by high surprise, happens when a searcher is moving across a space not previously explored. Exploitation, indicated by low surprise, happens when a searcher has a sustained focus on material they are already familiar with.

These local and global behaviors do not have to align. For example, text-to-text surprise may be high (local exploration) at the same time that text-to-past surprise is low (global exploitation). This can happen if Darwin's readings interleave different topics that he has already seen. If, for example, Darwin alternates between readings in philosophy with readings in travel narratives, then each local jump will have high KL (a travel narrative is dominated by topics that are rare in a philosophical text, and vice versa) and Darwin's readers will appear as a local exploration. However, once this pattern of alternation has been established, the average over past texts will include both philosophical and travel narrative topics, lowering the text-to-past surprise, and driving the system back towards global exploitation.
  
Conversely, text-to-text surprise can also be low while text-to-past surprise is high. This can happen if Darwin has recently begun a novel, but focused, investigation. In this situation, he focuses on a particular subset of topics that are under-represented in his overall history. If Darwin begins by focusing on philosophical texts, and then switches to travel narratives, his second (and subsequent) travel narrative readings will have low text-to-text surprise; but the average over past texts will be dominated by a long history of philosophical readings, leaving the text-to-past surprise high until he has accumulated so many readings of the latter type that they dominate the past average.

\section{Results}
\subsection{Exploration and Exploitation}
Over the 647 records in our corpus, Darwin's reading order led to a below-null average surprise, where the null is the average surprise of 1,000 permutations of Darwin's reading order, constrained by each book's publication date.

On average, the KL divergence from text to text in the corpus is 10.78 bits compared to a null expectation of 11.41 bits ($p \ll 10^{-3}$). Meanwhile, Darwin's text-to-past average surprise is 2.96 bits in the data versus 2.98 bits in the null ($p = 0.02$). Darwin's average surprise, in both text-to-text and text-to-past, is lower than expected from a null model. While our rejection of this simple null model provides little new insight into the cognitive process of a reader's decision-making---which we expect to have some correlation from book to book---it is a crucial test of the sensitivity of our methods themselves.

A surprise-minimizing path is one that orders the texts so as to minimize the total sum of text-to-text, or text-to-past, surprise. Finding this shortest path amounts to a variant of the ``traveling salesman problem'', which is famously difficult to solve (reviewed in \citet{cook2011}). The greedy shortest path algorithm attempts to approximate the surprise-minimizing path by starting with the first text that Darwin read, and choosing as the next one to read the one with smallest KL divergence from the first, and so on, minimizing at each step either the text-to-text, or text-to-past KL depending on which quantity one is interested in.

While Darwin's path is lower in surprise than the null, it is far larger than many paths that can be found: the greedy shortest-path algorithm, for example, can reduce the text-to-text average surprise to 2.11 bits and text-to-past average surprise to 2.97 bits. Table~\ref{table-steps} shows the raw local text-to-text and global text-to-past KL divergence data, along with the greedy shortest path single-visit traversals of the KL distance matrix. 

The null, actual, and greedy shortest-path results show that Darwin has a focused reading strategy despite not following a pattern of pure surprise-minimization. Interestingly, the greedy shortest path is slightly longer than the path Darwin took in the global measure. This highlights how preference for exploration at the local scale---\emph{i.e.}, not taking the closest book in topic space at each step---can lead to an unexpectedly efficient path in the global measure.

\begin{table}[t]
\begin{center}
\begin{tabular}{l|c|c}
& Local & Global \\
& text-to-text & text-to-past \\
& (bits/step) & (bits/step) \\ \hline
Darwin's order & 10.78 & 2.96 \\ \hline
Null (1,000 permutations) & $11.41\pm 0.28$ & $2.98^{+0.04}_{-0.02}$ \\
($p$-value) & $\ll 10^{-3}$ & 0.02 \\ \hline
Greedy shortest path & 2.11 & 2.97\\
\end{tabular}
\end{center}
 \caption{\textbf{Exploration habits.} Average text-to-text (local) and text-to-past (global) KL Divergence (bits/step) over the reading path. Text-to-past KL is lower, as Darwin's reading spreads out to cover the topic space and lowers the information-theoretic surprise of subsequent books. Darwin's reading strategy is simultaneously more exploitative than would be expected of a random reader while also not following a strategy of pure surprise-minimization. Note that Darwin's order displays lower global surprise than the greedy shortest path, which demonstrates that selecting the next most similar book is not the best overall strategy for minimizing average global surprise over time. \label{table-steps}}
\end{table}

\subsection{Readings over Time}

\begin{figure}[p]
\includegraphics[width=\textwidth]{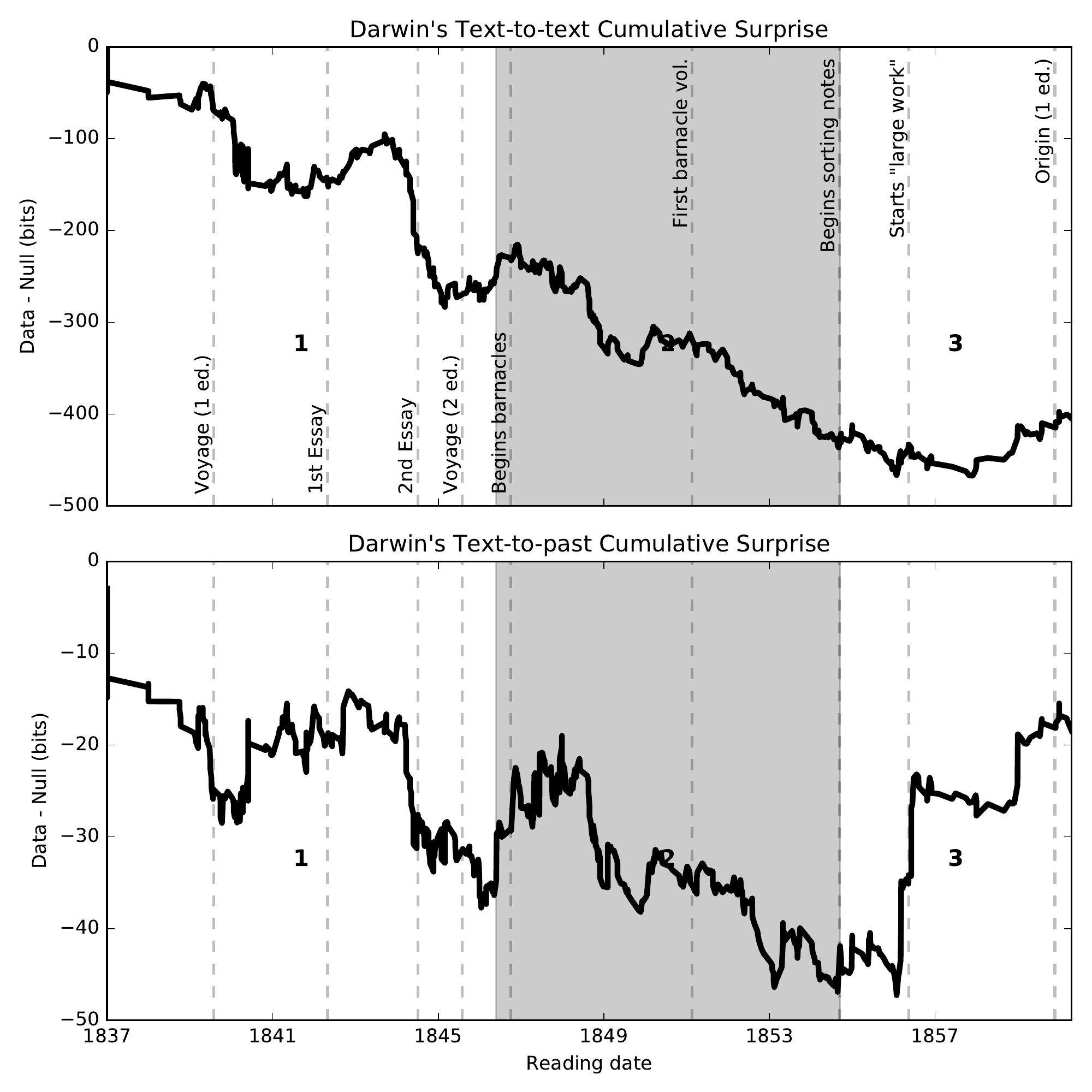}
\caption{{\bf Epochs of exploration and exploitation in Darwin's reading choices}. Text-to-text (top) and text-to-past (bottom) cumulative surprise over the reading path, in bits. More negative (downward) slope indicates lower surprise  (exploitation); more positive (upward) slope indicates greater surprise (exploration). The three epochs, identified by an unsupervised Bayesian model, are marked as alternating shaded regions with key biographical events marked as dashed lines and labeled in the top graph. The first epoch shows global and local exploitation (lower surprise). The second epoch shows local exploitation and global exploration (increased surprise, in text-to-past only). The third epoch shows local and global exploration (higher surprise in both cases).}
\label{fig:kl_over_time}
\end{figure}

While Darwin is on average more exploitative, this is not necessarily true at any particular reading date. Darwin's surprise accumulates at different rates depending on time, as can be seen in Figure~\ref{fig:kl_over_time} for the text-to-text case (top panel) and the text-to-past case (bottom panel). These figures plot the cumulative surprise relative to the null, so that a negative (downward) slope indicates reading decisions by Darwin that produce below-null instantaneous surprise (exploitation). Conversely, a positive (upward) slope indicates decisions that are more surprising than the null (exploration). 

Over the entire corpus, as we know from the previous section, Darwin's cumulative surprise is below the null expectation, showing an overall bias towards both local and global exploitation. Tracking the slopes in these charts over time, however, allows us to see how Darwin moves between low-surprise and high-surprise choices on a range of timescales. The interaction of these decision rules at the text-to-text and text-to-past levels characterize Darwin's behavior. 

\subsection{Rank Distribution}
\begin{figure}[t]
\includegraphics[width=\textwidth]{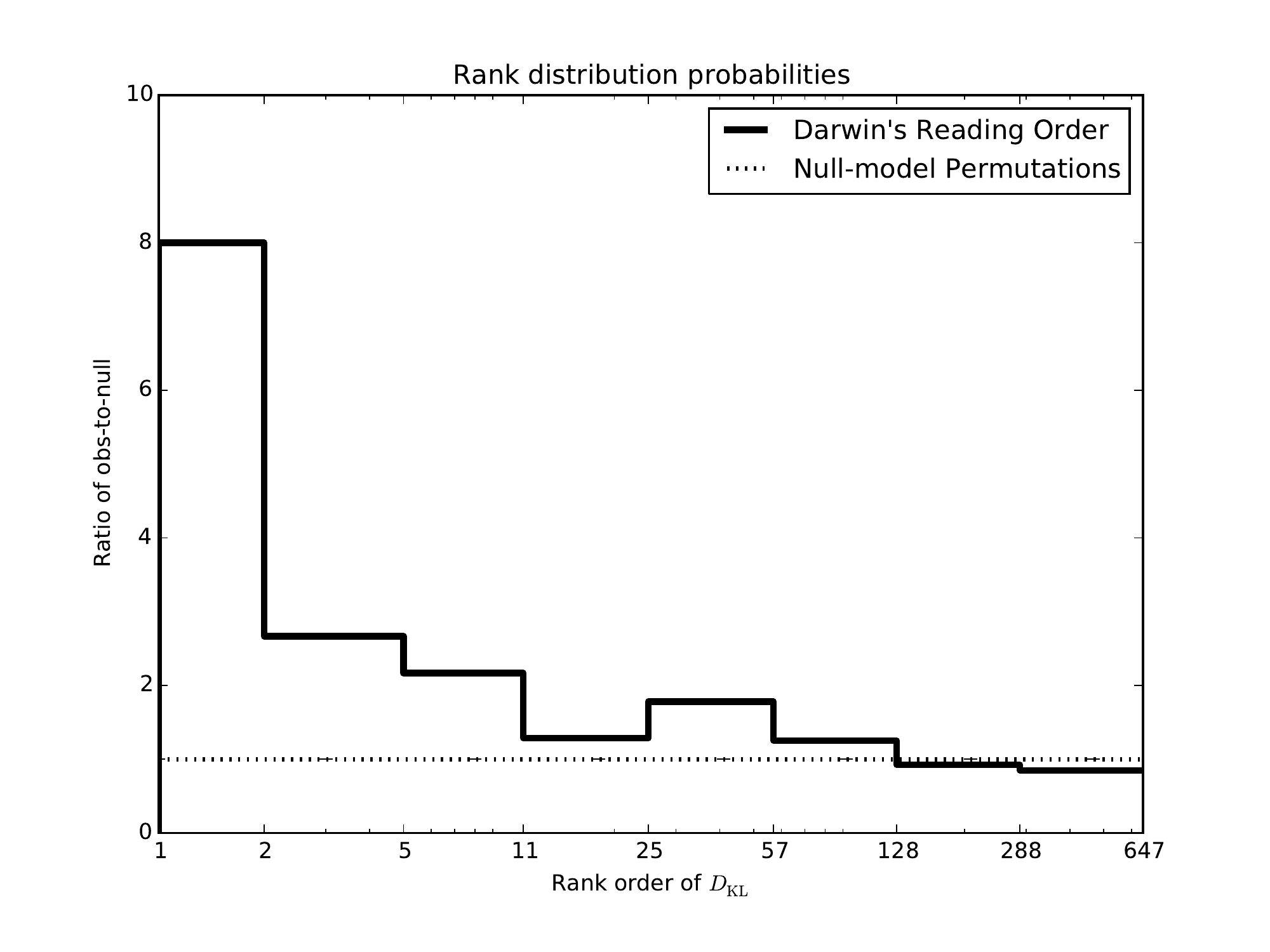}
\caption{\textbf{Rank distribution of text-to-text KL divergences}. Rank distribution of $D_{\mathrm{KL}}(\theta_i, \theta_{i+1})$ for Darwin's reading notebooks relative to a null-model permutation of his reading order, as indicated by the dashed line, with 95\% confidence intervals shown. The lines are logarithmically binned, showing clearly that Darwin is 8 times more likely to select the nearest KL neighbor, as opposed to volumes further away, which are selected 0.85 times as likely than the null.}
\label{fig:rankdistribution}
\end{figure}

In addition to the information-theoretic measures used above, descriptive statistics also capture Darwin's explore-exploit behavior. For each volume, we look at the rank of the KL divergence to the next volume by reading order compared to all other volumes in the corpus, as shown in Figure~\ref{fig:rankdistribution}. We can compare this to a null model, as described in the Methods.

Interestingly, Darwin is 8 times more likely than the null model to pick the nearest neighbor, indicating that explorations are overall rarer than exploitation choices, and emphasizing that exploitations do indeed occur on a text-to-text basis.

\subsection{Individual and Collective}
\begin{figure}[p]
\includegraphics[width=\textwidth]{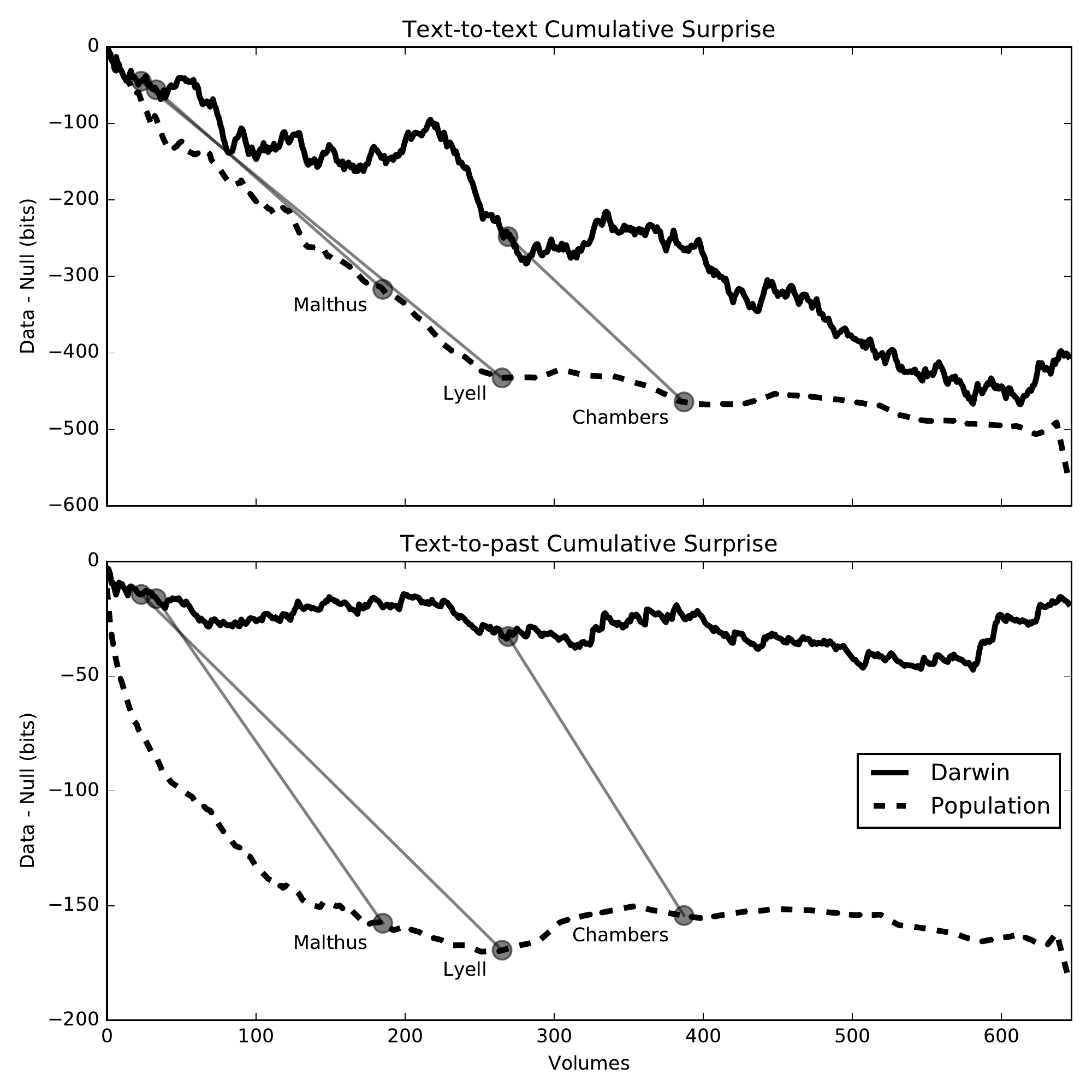}
\caption{{\bf Darwin's reading order more exploratory than the culture's production}. Text-to-text (top) and text-to-past (bottom) cumulative surprise over the reading order (solid) and the publication order (dashed). More negative (downward) slope indicates lower surprise (exploitation); more positive (upward) slope indicates greater surprise (exploration). In both cases, Darwin's cumulative surprise is higher than the publication order; in the second case, very significantly so. We mark the positions of three biographically significant books: Charles Lyell's \textit{Principles of Geology} (3rd ed., 1837; read in 1837), Thomas Malthus's \textit{An Essay on the Principle of Population} (1803; read on October 3, 1838), and Robert Chambers's \textit{Vestiges of the Natural History of Creation} (1844; read on November 20, 1844). Darwin's juxtaposition of Lyell and Malthus, for example, is characteristic of how Darwin's reading strategies reordered the products of his culture.}
\label{fig:culturalsurprise}
\end{figure}
While many studies see scientific innovations as following large-scale cultural trends \citep{Sun2013}, individuals can also be understood as ahead of their time, pursuing connections and ideas before they are recognized by the culture as a whole \citep{Johnson2010,Bliss2014}. By ordering Darwin's readings by publication date, rather than reading date, we see how the culture gradually accumulates and assimilates content. We then compare how the culture produced these texts to how Darwin, in his reading, consumed them.

Whether surprise is higher in the reading order or the publication order is a substantive empirical question. There are good reasons to imagine that the reading order will be lower in text-to-text and text-to-past surprise than the publication order---but equally plausible reasons for the opposite.

Consider, first, the idea that the reading order surprise is the lower of the two. Informally, this would suggest that reader has been at least partially successful at reordering the large number of texts into a more systematic sequence, finding similarities between chronologically distant texts and making sense of a disordered, distributed discovery process. A reader like Darwin, particularly early in his career when benefiting from more senior mentors, might be expected to attempt to do something similar.

However, equally reasonable arguments work for the opposite direction, where reading order is more surprising than the publication order. In this case, the reader is ``remixing'' the products of the past, sampling texts from different periods in such as way as to juxtapose thematically distant readings. While society accumulates themes gradually, in an exploitation regime, the reader is in an exploration regime, bringing them into unexpected contact.

While both possibilities appear \emph{a priori} plausible, the data decisively favor the second. Figure~\ref{fig:culturalsurprise} shows the text-to-text and text-to-past cumulative surprise for Darwin's reading order (solid line) compared to the publication date order (dashed line). Since volumes are published and read at different times, the $x$-axis is now ordinal (\emph{i.e.}, by position in the reading or publication sequence), rather than temporal (\emph{i.e.}, by date read or published). This allows us to compare his reading order to the publication order independent of time.

Compared to Darwin's reading practices, cultural production has lower rates of surprise. While cumulative text-to-text surprise for Darwin often shows either flat or positive (above-null text-to-text surprise) slope, the publication order path is less explorative in both text-to-text and text-to-past cases. These findings, both at high levels of statistical significance ($p\ll10^{-3}$), provide strong evidence for the remixing hypothesis.

\subsection{Strategy Shifts between Biographically Significant Epochs}

Between 1837 and 1860, Darwin's three major intellectual projects are reflected in his publication history. First, he began assembling his research journals on the geology and zoology from the voyage of the \emph{HMS Beagle}. The last of these volumes was published in 1846. A second epoch can be dated from 1 October 1846 when, while assembling the last of his \emph{Beagle} notes, Darwin discovered a gap in the taxonomic literature concerning the living and fossil \emph{cirripedia} (or barnacles) \citep{DAR158}. After a period of intense work, he published four volumes on the taxon from 1851 to 1854. A final epoch begins with his journal entry on 9 September 1854, marking the day he began sorting his notes for a major work on species \citep{DAR158}. The revolutionary \emph{Origin of Species} was published on 24 November 1859. These dates define three intellectual epochs: (1) from the beginning of records in 1837 to 30 September 1846; (2) from 1 October 1846 to 8 September 1854; and (3) from 9 September 1854 to the end of records in 1860.

We use the text-to-text and text-to-past models to characterize the exploration and exploitation of Darwin's reading behavior in each epoch. In instances where Darwin's average KL-divergence is above the null (more positive), Darwin is more exploratory. In instances where Darwin's average KL-divergence is below the null (more negative), Darwin is more exploitative. The degree to which he is in either mode is shown by the magnitude of the number. Table~\ref{table-darwin-epochs} shows these values.

Darwin's three biographical epochs are characterized by major shifts in both text-to-text and text-to-past surprise. Darwin begins, in epoch one, in an exploitation mode in both text-to-text and text-to-past. His turn to the barnacles in 1846 is marked by a shift from exploitation towards exploration at the text-to-past level (global shift to new area), and an intensification of his exploitation strategy at the local, text-to-text level (increased focus in this new area). In the third epoch, when Darwin ``began sorting his notes for Species Theory'' \citep{DAR158}, text-to-past remains in the exploration mode; text-to-text now shifts to exploration as well.
 
\begin{table}[th!]
\begin{center}
\begin{tabular}{c|ccc}
           & \emph{Beagle} writings & Barnacles  & Synthesis \\
Start date & 2 October 1836 & 1 October 1846 & 9 September 1854 \\ \hline 
Text-to-text & -0.68 & -0.96 & 0.32 \\
Text-to-past & -0.09 & -0.06 & 0.26 \\
\end{tabular}
\end{center}
\caption{{\bf Information-theoretic correlates of biographically significant events}. In this first table, we measure the relative surprise of Darwin's reading order by reference to dates derived from qualitative biographical work. The first major epoch of Darwin's intellectual life identified in this fashion corresponds to his post-\emph{Beagle} work, when his readings were mostly in natural history and geology. Both text-to-text and text-to-past surprise remain low---a regime of simultaneous local and global exploitation. The second epoch, when Darwin turns to a study of barnacles, shows an increase in text-to-past surprise (new topics; exploration) coupled with a decrease in text-to-text surprise (smaller jumps within these new topics; exploitation). The third epoch, when Darwin begins to collect his notes for his ``Species Theory'', is characterized by a rise in both text-to-text and text-to-past surprise. Now Darwin is neither repeatedly returning to well-covered topics (as in epoch one), nor turning his attention to a new, but narrow, range (as in epoch two), but rather ranging widely over new, previously understudied topics.}
\label{table-darwin-epochs}
\vspace{1cm}
\end{table}

\subsection{Unsupervised Detection of Strategy Shifts}

In addition to using Darwin's personally-specified epochs, we use a Bayesian model (Bayesian Epoch Estimation [BEE], see Methods) to estimate epoch breaks from text-to-text and text-to-past surprise alone. This process determines inflection points for Darwin's behavior without reference to outside biographical facts, allowing us to determine the extent to which the intellectual epochs identified by traditional, qualitative scholarship align with purely information-theoretic features of his reading.

For text-to-text surprise, we find the boundary at 16 September 1854 (log-likelihood relative to not having a boundary: $\Delta \mathcal{L} = 2.61$; 14 times as likely as not having a boundary). This is within 1 week of his journal entry on 9 September 1854 marking the start of his synthesis.  For text-to-past surprise, we find the boundary at 27 May 1846 ($\Delta \mathcal{L} = 6.17$; 479 times as likely). The difference between the automatically-selected date and his recorded start date on 1 October 1846 suggests the need for further investigation of what our results suggest was a period of relative uncertainty for Darwin about his research plans. 

The exploration-exploitation characteristics of these epochs are shown in Table~\ref{table-auto-epochs}. The close proximity of these automatically-detected breaks and the biographically significant epochs of the previous section confirm the central role of information-theoretic surprise in tracing the evolution of Darwin's search strategies. 

Both the text-to-text and text-to-past models make highly simplifying assumptions about the nature of Darwin's reading. The text-to-text case makes the most severe assumption of all: that Darwin's reading choices are conditional solely on the book just read. If Darwin's reading choices are strongly influenced by longer term memory (as seems likely), and it is these patterns which define the true epoch boundaries, it is natural that evidence for epoch boundaries in the text-to-text BEE model is weaker than the text-to-past case. In addition, our BEE makes the simplifying assumption that successive surprise values are independent draws from the distribution associated with that epoch.

\begin{table}[t]
\begin{center}
\begin{tabular}{c|ccc}
           & \emph{Beagle} writings & Barnacles  & Synthesis \\
Start date & 2 October 1836 & 27 May 1846 & 16 September 1854 \\ \hline 
Text-to-text & -0.78 & -0.76 & 0.21 \\
Text-to-past & -0.11 & -0.02 & 0.24 \\
\end{tabular}
\end{center}

\caption{{\bf Biographically significant events are detectable by unsupervised learning}.
Even in the absence of qualitative information about Darwin's life, our Bayesian model, based only on text-to-text and text-to-past surprise measurements finds---with only slight differences---the three historically-noted epochs of Table~\ref{table-darwin-epochs}: (1) from the start of our records in 1837 until text-to-past surprise changes from exploitation to exploration in Spring 1846, (2) from Spring 1846 until text-to-text surprise changes from exploitation to exploration in Autumn 1856, and (3) from Autumn 1856 to the end of our data, when both (local) text-to-text and (global) text-to-past selection behaviors are in the exploration state. The automatically-selected and biographical epochs agree on these characterizations, with small variance in the second epoch's start date.}
\label{table-auto-epochs}
\end{table}

An AIC analysis (Section~\ref{sec:aic}) further emphasizes the relative strength of evidence for text-to-text and text-to-past epoch boundaries. Evidence for boundaries in the text-to-text BEE model is naturally weaker than the text-to-past case. The text-to-text model assumes decisions are made solely by the last read text, as opposed to longer term memory, which seems more plausible. In addition, our text-to-past BEE makes the simplifying assumption that successive surprise values are independent draws from the distribution associated with that epoch. However, text-to-past results are intrinsically correlated, so this assumption only approximates the text-to-past case.

The results of our AIC analysis are shown in Table~\ref{table:aic}.

\begin{table}[h]
\begin{center}
\begin{tabular}{r|c|c|c|c}
& Breaks & k & AIC & relative $\mathcal{L}_{AIC}$ \\ \hline
Null T2T &  & 2 & 3911.61 & $1.0$ \\
1-epoch T2T & 16 Sep 1854 & 5 & 3912.38 & $0.68$ \\ 
2-epoch T2T & 28 Dec 1847 \& 16 Sep 1854 & 8 & 3914.67 & $0.21$ \\ \hline
Null P2T &  & 2 & 2035.18 & $1.0$ \\
1-epoch P2T & 27 May 1846 & 5 & 2028.83 & $23.82$ \\ 
2-epoch P2T & 8 Feb 1849 \& 4 Sep 1854 & 8 & 2021.93 & $750.70$ \\ \hline
\end{tabular}
\end{center}
\caption{{\bf AIC Model Selection}. The likelihood for each division in the 2-epoch model is shown in Figure~\ref{fig:epoch-estimation}. The AIC of the independent selection for a 2-epoch model is also shown.  Note that the AIC for text-to-past selects for epoch breaks, but not for text-to-text (see Results).}
\label{table:aic}
\end{table}

\begin{figure}
    \centering
    \includegraphics{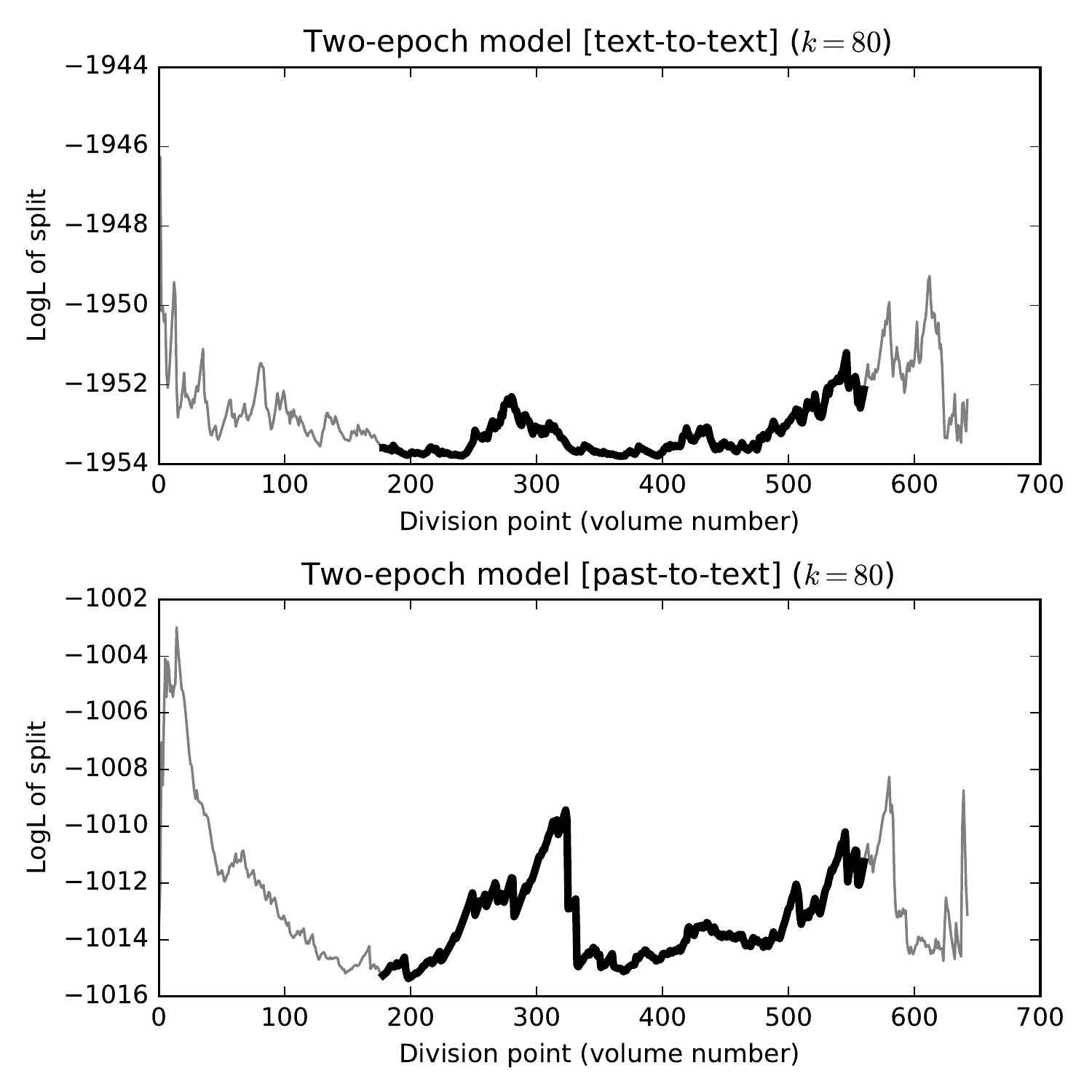}
    \caption{{\bf Two-epoch model likelihoods}. Fisher maximum-likelihood estimation for a 2-epoch BEE model over the text-to-text and text-to-past $k=80$ models of 647 of Darwin's readings. The darker line indicates the window of the 5-year minimum epoch length. Note the phase transition at the 325th volume in the text-to-past case (bottom) and the 548th volume in the text-to-text case (top). Note also that the text-to-past case comes close to transition at the 548th volume as well, indicating the strength of the transition to exploration in the third epoch on both local and global scales.}
    \label{fig:epoch-estimation}
\end{figure}{}

\section{Discussion}
Models of cultural change often understand innovation as a multi-level combinatoric process, in which bundles of ideas are subject to cultural processes analogous to natural selection \citep{Jacob1977,Wagner2014}. These evolutionary analogies typically consider change at the population level, as new ideas are created, spread, and modified by the crowd. A variety of recent studies covering conceptual formation in science, technology, and the humanities have taken this population-level perspective, including work on the recombination of patents \citep{Youn2014}, novelties \citep{Tria2014}, and citations \citep{Garfield1979}. Sociological studies of scientific practice have investigated how disciplines \citep{Sun2013} or ``communities of practice'' \citep{Bettencourt2015} are formed.

The mechanisms driving cultural innovation at the population-level cannot, however, be fully understood without taking into account the cognitive processes that operate at the level of individual scientists. We have taken a step towards modeling these individual-level processes by studying the information foraging behavior of one preeminent scientist, using an information-theoretic framework applied to probabilistic topic models of his reading behavior. The information-theoretic measure we use to quantify surprise, KL divergence, connects both analytically and empirically to linguistics \citep{COGS:COGS12167,hale2001probabilistic,Levy20081126,light2002statistical,COGS:COGS1267,resnik1993selection} and visual search \citep{Demberg2008193,itti2009bayesian}. The LDA topic models that generate this information space also have cognitive correlates \citep{Griffiths2007}.

Our methods allow us to zoom in on Darwin's individual-level process to identify major epochs in his reading strategies. Over time, Darwin shifts towards increasing exploration as he prepares to write the \emph{Origin}. Interestingly, the overall order we find empirically---exploitation then exploration---is in contrast to many of predictions derived from mathematical accounts of how optimal agents navigate the exploration-exploitation dilemma~\citep{gittins1979bandit}. These generally predict that individuals begin with exploration (see, \emph{e.g.}, \citet{Berger-Tal2014}), shifting later to exploitation as they gain information about the environment. 

If Darwin's early exploitation was guided by an earlier exploration phase prior to 1837, then our results starting with exploitation are consistent with these earlier mathematical results. Alternatively, it may be the case that the early phase of exploitation was necessary for Darwin to gain sufficient abilities or confidence to explore in a reliable fashion later. Finally, it is worth noting that some evidence in favor of these standard models can still be found in the short-term switch towards greater exploitation at the text-to-text surprise from the first to the second epoch. This suggests that these foraging behavior models may be useful at shorter timescales in an individual's life. We return to this notion in Chapter~\ref{sec:darwin-extension}, when we consider books Darwin \emph{could} have read prior to the publication of \emph{Descent of Man}.

Our use of Charles Darwin allows us to validate our methods by reference to the extensive qualitative literature on his intellectual life. Having presented a general information-theoretic framework for describing the exploitation-exploration trade-off, we can now look at other searchers to see if other strategies exist for managing the exploitation-exploration trade-offs. Expanding these results beyond the Darwin test case will be essential to providing new empirical constraints on theories of how individuals explore the cultures of their time (see Chapter~\ref{chapter:extensions}).

Our method also allows us to compare the individual and the collective. We have found, in particular, that Darwin followed a path through the texts that was more exploratory than the order in which the culture produced them. Our work reveals an important distinction between these two levels of analysis; underneath gradual cultural changes are the long leaps and exploration comprising an individual's consumption, combination, and synthesis. 

Our cognitive analysis of these records builds upon decades of archival scholarship and innovations in the digital humanities. Darwin's industry extends beyond the bounds of the data we use here, however. During the \emph{Beagle} voyage, he kept a library of 180 to 275 titles \citep{DCPBeagle}. His retirement library contains 1,484 titles \citep{rutherford1908}. Darwin's handwritten marginalia in 743 of these books is currently being digitized by the Biodiversity Heritage Library. This retirement library contains many texts not included in our study of his reading records, which only last through 1860. We examine some of these later readings in Chapter~\ref{chapter:extensions}. Finally, an extensive network of correspondents also contributed to Darwin's knowledge. The Darwin Correspondence Project\footnote{\url{https://www.darwinproject.ac.uk/}} contains over 15,000 letters to and from Darwin before 1869. A complete understanding of his information foraging will necessarily seek to understand this separate social process. %

Darwin's sustained engagement with the products of his culture is remarkable. He averaged one book every ten days for twenty-three years, including works of fiction and foreign-language texts which are not part of the present analysis. For some months in our data, Darwin appears to be reading one book every two days, a fact even he was astonished by:
\begin{quote}
When I see the list of books of all kinds which I read and abstracted, including whole series of Journals and Transactions, I am surprised at my industry. \\ \small{--- \emph{Autobiography of Charles Darwin}, p. 119\nocite{DarwinAutobiography}.}
\end{quote}

\noindent Darwin not only consumed information, it consumed him. In the words of Herbert Simon: ``what information consumes is rather obvious: it consumes the attention of its recipients'' (\citeyear{Simon1971b}). Even the most ambitious individuals must confront and manage the limits of their own biology in allocating attention.

Standard theories for how individuals balance the exploration-exploitation tradeoff draw on classic work in the statistical sciences~\citep{gittins1979bandit} and machine-learning~\citep{Thrun92c}, and often focus on determining mathematically optimal strategies for different environments~\citep{Cohen2007}. Our work provides both new tools for the study of how individuals in the real world approach these problems, and new results on an exemplar individual. 

\section{Conclusion}
Charles Darwin's well-documented reading choices show evidence of both exploration and exploitation of the products of his culture. Rather than follow a pure surprise-minimization strategy, Darwin moves from exploitation to exploration, at both the local and global level, in ways that correlate with biographically-significant intellectual epochs in his career. These switches can be detected with an unsupervised Bayesian model. Darwin's path through the books he read is significantly more exploratory than the culture's production of them.

To what extent the patterns we identify in Darwin, and his relationship to culture as a whole, hold for other scientists in other eras is an open question. The development of an individual is in part the history of what they choose to read, and it is natural to ask what patterns these choices have in common. 

The methods we have developed and tested here represent the first application of topic modeling and cognitively-validated measures, such as KL divergence, to a single individual. These domain general methods can be used to study the information foraging patterns of any individual for whom appropriate records exist, and to look for common patterns across both time and culture. Even though we present but a single case study, our experimental design and null models validate the observations as not just measurements, but significantly above chance that periods of exploration and exploitation existed in Darwin's readings. These measures can now be extended to other figures.

These results also advance the state of the art in information foraging. The exploration/exploitation results from analysis of KL divergence are not merely a measure, but integrate a model of reading behavior, relative to a null over permutations of the books read. The text-to-text and past-to-text measures define an upper and lower bound on a forager's behavior. This lays a foundation for future work on serial position effects and determining an appropriate decision window for reading decisions.
\chapter{Case Study: Darwin's Writing Behavior}
\label{chapter:writings}

The past chapter examined Darwin's reading notebooks and how he organized the cultural mileu through differing phases of exploration and exploitation through information-theoretic measures over topic models. In this chapter, we examine the fruit of that foraging expedition---\emph{The Origin of Species}---and how its development observed in the writing of intermediate drafts is related to his reading behavior. While the previous chapter focused on macro-level trends in topic space, this chapter examines micro-level observations of specific topics in specific texts. These micro-level observations are shown to do more than state the obvious, but advance argumentation in the history of science. In terms of information foraging theory, the previous chapter assumed no goal-directed behavior, yet in our measure of relative surprise we were able to automatically extract epoch breaks that coincided with biographical changes in goals. In this chapter, we look at some of those writings and expand our notion of information foraging by showing how reading occurs relative to a fixed point, such as the draft of a text.

Charles Darwin left a rich written record of his scientific work and practice. Despite the vast amount of scholarship that has been devoted to this record, historians of science remain uncertain of the answers to basic questions in Darwin's intellectual development. One such question concerns the delay from the early 1840s, when Darwin first drafted his theory of evolution, until 1859, when the first edition of \emph{The Origin of Species} was published. Close reading of his letters and manuscripts has led historians to different conclusions about the reasons for ``Darwin's delay'', ranging from fear of persecution, concerns about his credibility among biologists, and interruptions caused by illness and other life events \citep{Gruber1974,Browne2006,VanWyhe2007,Quammen2007,richards2015,Buchanan2017}. A second, related, question concerns the degree to which Alfred Russel Wallace's essay, which Darwin received in 1858, anticipated Darwin's ideas. Historians have disagreed about the extent to which Wallace deserves to be credited as co-discoverer of the theory of evolution by natural selection \citep{Merton1957,VanWyhe2013,Costa2014}. In this chapter, we apply computational methods to augment the kinds of evidence to which historians usually appeal in these disputes. 

We base our analyses on novel information-theoretic methods that allow us to compare the distributions of topics among Darwin's writings (the two essay sketches and \emph{The Origin} itself) and Wallace's essay. Topic models, derived by a machine learning run on the books that Darwin read over the course of his intellectual life, provide the context for these comparisons. These models define  a ``topic space'', within which we can situate Darwin's and Wallace's writing. We then apply quantitative measures of the distances between the writings, and represent directional changes through time. Darwin facilitated the possibility of taking this approach by making a note of each book he read or wanted to read from 1837, when he returned to England aboard the \emph{HMS Beagle}, until 1860, just after publication of \emph{The Origin} \citep{vorzimmer1977}. These reading notes, along with the written manuscripts, together provide information about decades of individual scientific practice. The textual nature of these records make them particularly amenable to computational analysis, especially given that it is now possible to supplement them with digitized full-text versions of many of the volumes Darwin read.

In this chapter we extend our use of these reading models to supply quantitative evidence about the relationships among the four pieces of writing by Darwin and the one by Wallace. This provides new answers to two basic historical puzzles, previously treated qualitatively by historians of science, namely:

\begin{enumerate}
\item the mythology of ``Darwin's Delay,'' \emph{i.e.}, that despite completing an extensive draft in 1844, Darwin waited until 1859 to publish \emph{The Origin of Species} due to external pressures; and
\item the relationship between Darwin and Wallace's contemporaneous theories, especially in light of their joint presentation.
\end{enumerate}

The quantitative evidence induced from the models supplements but does not replace the evidence provided by traditional methods. This work does not replace close reading with algorithmic analysis. Rather, it expands the number of methods and range of evidence which historians use to understand the scientific record, similar to the interpretive approach that has been advocated by others for the digital humanities \citep{Rockwell2016}.

\section{Dataset Curation}
The reading notebook corpus is identical to that in Chapter~\ref{sec:notebook-corpus}.

In addition to the base corpus and model, we used four additional digitized texts:

\begin{itemize}
\item \emph{The Origin of Species}, 1st edition, drawn from Project Gutenberg.
\item The \emph{1842 Sketch}, drawn from the Project Gutenberg text of \emph{The Foundations of the Origin of Species}.
\item The \emph{1844 Essay}, drawn from the Project Gutenberg text of \emph{The Foundations of the Origin of Species}.
\item Wallace's ``On the tendency of varieties to depart infinitely from the original type'', which was co-published with an excerpt of Darwin's 1844 Essay on 30 June 1858, drawn from \emph{Darwin Online}.
\end{itemize}

\section{Methods}
We trained an additional topic model over the reading notebooks corpus at $k=200$ topics to decrease the granularity of the topic models as we examined specific texts.

For each of the four texts listed above, we employed query sampling (Chapter~\ref{sec:query-sampling}) to project the texts into topic space. We performed 1,000 samples per text. We used the average topic distribution of the samples to determine distances to each of the readings.

\subsection{The Topics in Darwin's Readings}
As mentioned above, we trained an additional 200-topic model to increase the granularity of topics for the interpretive work in this chapter. The topics identified in the 200-topic model range from very generic to quite specific and interpretable. Sometimes the gist of a topic can be assessed by looking at the ten or so words assigned the highest probability, although it is important to recognize that these words do not fully represent the topic \citep{Schmidt2012}. Topics are more than their top ten words; indeed, in the 200-topic model examined here, it  takes 300-600 words to reach 50 percent of the probability mass of the topic. In this model, ``\texttt{T1}'' labels the topic identified as the most represented in the corpus as a whole (\texttt{T2} is the next most represented, etc.). \texttt{T1} assigns the highest probabilities to these ten words: theory, physical, facts, philosophy, phenomena, views, argument, study, conditions, changes. These ten words suggest an abstract level of theorizing about science and philosophy. However, it requires 530 words to comprise 50 percent of the  probability mass of \texttt{T1}.  Manually inspecting this longer list confirms that the topic is primarily weighted towards the sorts of  generically abstract words found among the top ten. This list also contains ``philosophy'' and four other words sharing the same ``philosoph-'' stem, as well as the words ``religious'' and ``theological''.

Simply knowing that \texttt{T1} is highest represented in the reading corpus as a whole doesn't tell us much about that corpus, however. The strength of topic modeling lies in the comparisons it enables among a set of documents, in contrast to simply counting the particular words used in a document. Using \texttt{T1} to retrieve the top three books in which it is most highly concentrated, returns three philosophy of science books: Powell's \emph{Essays on the Spirit of Inductive Philosophy}, Lewes's commentary on Comte's \emph{Philosophy of the Sciences}, and Whewell's \emph{Astronomy and General Physics Considered with Reference to Natural Theology}. 

\begin{figure}
\begin{center}
\includegraphics[width=\textwidth]{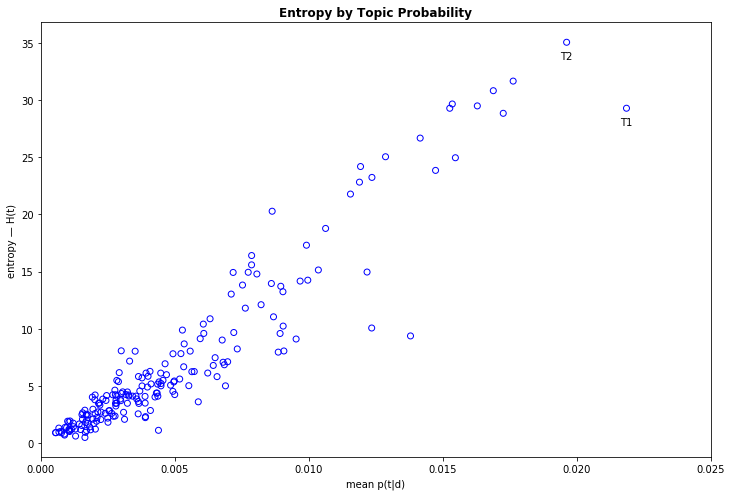}
\end{center}
\caption{\textbf{Mean probability and entropy of topics}. This scatter plot shows the mean probability of topics across all documents (X-axis) and the Shannon entropy across documents (Y-axis) of all topics in the 200-topic model. Topics in the upper right appear frequently and in many texts. The majority of the topics appear with very low probability and are unevenly distributed across the texts.}
\label{fig:topic-entropy}
\end{figure}

One would like to know how characteristic \texttt{T1} is of each book Darwin read---in other words, were a plurality of the books dealing with high-level theoretical issues, or was this kind of discussion relatively concentrated in a small subset of the books? This question can be addressed via the notion of the Shannon entropy of the topic across all documents: a topic that is more evenly represented across the documents will have a higher entropy.  \texttt{T1} is the right-most point in Figure~\ref{fig:topic-entropy} and it has one of the highest entropy values, meaning that it was assigned by the model to many of the volumes that Darwin read. 

The topic with the second most weight in the corpus as a whole (\texttt{T2}) is distributed has higher entropy and thus is more broadly distributed than \texttt{T1}. Its  ten highest probability words are somewhat generic words that  seem diagnostic of the type of writing rather than specific content: circumstance, prove, observation, observe, render, sufficiently, acquainted, admit, naturally, rendered.  The books retrieved using \texttt{T2} are dominated by tracts discussing domestication: Anderson's six-volume work \emph{Recreations in Agriculture, Natural History, Arts and Miscellaneous}, Sebright's \emph{The Art of Improving the Breeds of Domestic Animals}, and Pallas's \emph{An Account of the Different Kinds of Sheep Found in the Russian Dominions and Among the Tartar Hordes of Asia}. \texttt{T2} has higher Shannon entropy, however, than \texttt{T1}, indicating that it is more uniformly distributed throughout the corpus than \texttt{T1}, and the next 10 most probable words (useful, proof, fully, discover, facts, powerful, derived, perceive, individual, obtain) support the idea that \texttt{T2} tracks a  mode of writing about empirical results rather than towards a specific subject matter. Manual inspection of the 562 words comprising 50 percent of \texttt{T2}'s probability mass, confirms that the topic is more oriented toward terms surrounding observation and experiment than the more abstract, theory-oriented \texttt{T1}.

As can be seen in Figure~\ref{fig:topic-entropy}, the majority of topics discovered by the model are relatively low probability and low entropy in the corpus as a whole, indicating that they are sparsely distributed among the books. Because such topics are highly discriminative among the books, they prove very useful both for retrieving those books and for the kind of analytical/interpretative work we are engaged in here.  For example, \texttt{T49}---whose ten highest probability words are pigeons, eggs, feathers, fowls, hen, cock, pigeon, breed, fowl, hens---is of obvious relevance for understanding the sources contributing to \emph{The Origin}. 

While the observations we have made so far about the distribution of topics among Darwin's readings might seem relatively obvious to historians (or even lay readers) of Darwin, they nonetheless confirm that the modeling methods are automatically finding meaningful relationships among the texts. The power of using computational methods is that they can detect patterns that may be invisible to readers, but which are nonetheless interesting, such as the large-scale changes in reading selections we reported in Chapter~\ref{chapter:readings}.

\subsection{Sampling the Writings with Query Sampling}
Historical interpretation is often challenged by the discovery of new documents. Query sampling is a method for assigning a mixture of topics from a prior model to documents not in the original training set. This process is also known as ``fitting'' a model. In the present study, we fit Darwin's writings to the model trained on the books he read. In query sampling, topics are probabilistically assigned to the words in the new document and the assignments are revised iteratively until they stabilize using the same method used to train the original model. Because of the random starting point, running the sampling process multiple times leads to different topic distributions for the same text, although the results tend to cluster reliably in ways that we go on to describe.

\begin{figure}
\begin{center}
\includegraphics[width=\textwidth]{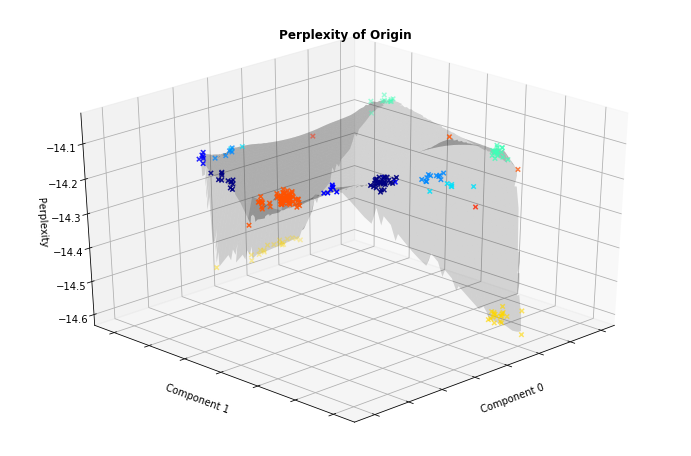}
\end{center}
\caption{\textbf{Perplexity landscape of 1,100 samples of \emph{The Origin of Species}}. The colored Xs correspond to points falling in the eight clusters represented in Figure~\ref{fig:origin-clusters}. The principal components are derived from the dimensional reduction process which allows the 200 dimensions of the topic model to be represented two-dimensionally. Point heights are relative to perplexity, a measure of model fitness. The grey manifold is a surface fitted to the perplexity values, indicating local maxima and minima of model fitness. Principal components are sometimes interpretable, although it is not important for our present purposes that they be interpreted here.}
\label{fig:landscape}
\end{figure}

The approach taken in query sampling is similar to starting in a random location in a biological fitness landscape and hill-climbing towards a peak. Different starting points lead to different peaks (and, even when ascending the same peak, different runs may end up slightly higher or lower on that peak, due to slightly different topic mixtures). Nevertheless, if the exercise is repeated multiple times, an overall picture of the landscape emerges from repeated encounters with the same peaks. The counterpart to fitness in our analysis is \emph{perplexity}, an information theoretic measure of how well the current state of the word-topic distribution matches the observed distribution of words in the text. A visual representation of this fitness landscape for \emph{The Origin} is in Figure~\ref{fig:landscape}.

This variability in outputs is something to be understood and harnessed, not feared. It can support a crucial aspect of the humanities: the interpretation of texts. For any work, there is not claimed to be a single ``correct'' interpretation but rather a set of interpretations in dialog with one another. Digital methods augment existing debates in the humanities by providing different ways of looking at the text \citep{Rockwell2016,Allen2017}. Each fitness peak in the topic landscape can be treated as a way to interpret the text. These interpretations are encoded as a particular distribution of topics.

\begin{figure}
\begin{center}
\includegraphics[width=\textwidth]{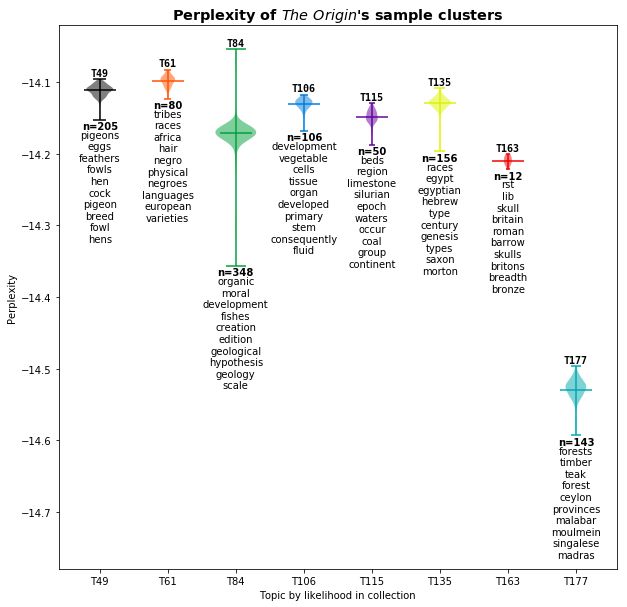}
\end{center}
\caption{\textbf{Cluster analysis of \emph{The Origin of Species}}.  This ``violin'' plot shows the distribution of perplexity (fit to the document) by topic cluster for \emph{The Origin of Species}. The number below each cluster shows the number of samples classified in that group, and the surface area of the bulging part of each violin is proportional to this number. The horizontal line in the center of each violin shows the median perplexity, while the vertical lines span the outliers in each cluster.}
\label{fig:origin-clusters}
\end{figure}

We  approach the diversity of the sampled results by applying a clustering algorithm to the topic distributions, using the silhouette method to choose the number of clusters \citep{Rousseeuw1987}. For \emph{The Origin}, this method detects eight clusters, shown in Figure~\ref{fig:origin-clusters}. Each cluster has a different highest-probability topic, which characterizes the primary interpretation of the text for that cluster. Inspection of the topics reveals that they are immediately applicable to \emph{The Origin}. For example, pigeons (\texttt{T49}) provide a significant example for Darwin. Likewise, the dominant topic of the largest cluster (\texttt{T84}) captures some key areas of theoretical concern with `'`development'', ``creation'', ``geological'', ``organic'' among the top ten highest probability words.

Because the topics fit to \emph{The Origin} by query sampling are derived from the model of the readings, some of the words that have a high probability for a topic in the readings are likely not to appear in \emph{The Origin} at all. For example, the second most likely word in \texttt{T84}, ``moral'' does not appear in the first edition of \emph{The Origin}. Likewise, some of the geographic terms prominent in \texttt{T177} do not appear in the book.  Indeed, \texttt{T177} (with terms related to forests and South Asian geography and culture) presents a somewhat idiosyncratic view of \emph{The Origin}. The statistical ``perplexity'' of this cluster with respect to the text confirms a relatively poor fit. Nonetheless, the assignment of \texttt{T177} is grounded both in Darwin's reading of Falconer's \emph{Report on the teak forests of the Tenasserim provinces} in 1853, and in his writing. Falconer is mentioned six times in the first edition of \emph{The Origin}, and related issues are discussed in passages such as this, from chapter 5: 

\begin{quote}[W]e have evidence, in the case of some few plants, of their becoming, to a certain extent, naturally habituated to different temperatures, or becoming acclimatised: ... trees growing at different heights on the Himalaya, were found in this country to possess different constitutional powers of resisting cold. Mr. Thwaites informs me that he has observed similar facts in Ceylon.
\end{quote}
\texttt{T177}, like other clusters featuring geographical and ethnographic terms (\texttt{T61}, \texttt{T135}, \texttt{T163}), highlights how Darwin's own travels, his correspondence with other travelers, and his reading of their published accounts expanded the global range of his evidence.

\section{Darwin's Delay}
Darwin's synthesis of \emph{The Origin} began well before he started organizing his notes in 1854, with two private essays written in 1842 and 1844\footnote{These two essays were published posthumously as \emph{The Foundations of The Origin of Species} \citep{Foundations}.}. It is a historical curiosity that he would wait until 1859 to publish his work, especially as immediately after finishing the second essay he wrote to his wife, Emma, with an addendum to his will concerning publication instructions should he die before finishing his work \citep{DCP761}. This period has become known as ``Darwin's Delay'' \citep{VanWyhe2007}. Theories about its primary cause include general fear of persecution \citep{Gruber1974,Quammen2007}, the anonymous 1844 publication of \emph{Vestiges of the Natural History of Creation} \citep{Chambers1844} highlighting gaps in Darwin's argument \citep{Browne1995}, and extended illness \citep{Gruber1974,VanWyhe2007,richards2015}. We provide evidence for another motivation for the delay: Darwin wanted more time to gather evidence and develop his argument, a view supported by \citet{VanWyhe2007}, \citet{richards2015}, and \citet{Buchanan2017}.

We use KL-divergence to trace the increase in cognitive surprise through Darwin's written presentations of his theory. We looked at this two ways: one in comparison to Darwin's reading order, and the other in comparison to the cultural accumulation of knowledge signified by publication dates of the volumes read.

\begin{figure}
\begin{center}
\includegraphics[width=\textwidth]{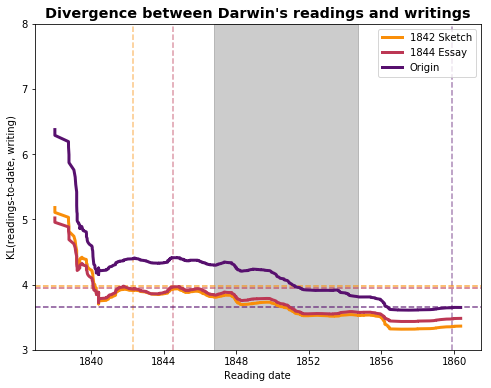}
\end{center}
\caption{\textbf{KL divergence between Darwin's readings and writings by reading order}. Vertical dashed lines indicate date of publication. \emph{The Origin} diverges more from the readings-to-date than either of the two previous drafts at all time points. However, each successive draft diverges less from the readings-to-date at the time of writing. The curves been smoothed by averaging over 5 years of readings.  The grey rectangle shows the duration of the barnacle project.}
\label{fig:doc-divergence-history}
\end{figure}

Figure~\ref{fig:doc-divergence-history} shows that with respect to what Darwin had read at any given time (position along the $x$-axis), \emph{The Origin} is more divergent than either of the earlier essays, and that the 1844 essay is more divergent from the readings than the 1842 version. Interestingly, however, the 1842 and 1844 essays are both more divergent from Darwin's readings at their respective times of writing, than the \emph{Origin} is from the readings he had completed by 1859. This computational evidence supports the claim that Darwin's continued reading during the period between 1844 and 1859 was materially relevant to what he eventually wrote.

This period includes Darwin's barnacle project. However, the barnacle project was not undertaken specifically to tackle the species problem \citep{Buchanan2017}.  In Figure~\ref{fig:doc-divergence-history}, the duration of the barnacle project is shown by the grey rectangle. The period 1851-1854 shows little change in the divergence between Darwin's readings and writings.  This emphasizes the degree to which the barnacle project, or at least the reading Darwin engaged in during the period of his most intense attention to that project when the four volumes were being published, was independent of the species theory.  The surprise of \emph{The Origin} relative to Darwin's readings begins to decrease in 1854 when he begins organizing his notes. All three texts decrease sharply in 1856, around the time he commits pen to paper to his ``large work on species'' on 14 May 1856 \citep{DAR158}. The specific decrease in divergence occurs when Darwin reads \emph{Types of Mankind} \citep{Nott1854}, which catalogs differences among different human races. This is due to a heavy contribution of \texttt{T-135} from that text, being one of the few volumes containing the words `races' and `types' extensively.

\subsection{Publication Order}
\begin{figure}
\begin{center}
\includegraphics[width=\textwidth]{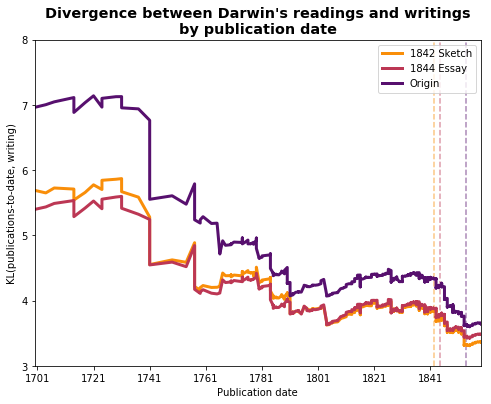}
\end{center}
\caption{\textbf{KL divergence between Darwin's readings and writings by publication order}. Vertical dashed lines indicate date of publication. \emph{The Origin} diverges more from the publications-to-date than either of the two previous drafts at all time points. }
\label{fig:doc-divergence-pub}
\end{figure}

Figure~\ref{fig:doc-divergence-pub} shows the divergence of Darwin's writings with respect to the order his readings were published, reflecting the cultural order of production. Here we see that \emph{The Origin} diverges more than either of the earlier essays. The 1842 essay is more divergent than the 1844 essay throughout most of the 18th century, although this could just be a consequence of undersampling of publications from those earlier years in Darwin's readings.

The decreasing divergence from Darwin's writings to the books published at the time of writing shows that Darwin was following the scientific debate, and indeed incorporating those ideas into his writings. However, that the 1842 essay has the lowest surprise in 1859 also shows that the culture Darwin was engaging with was also getting closer to his earlier proposals, perhaps indicating a higher likelihood of accepting his fully-formed ideas in \emph{The Origin}.

\section{Wallace and Darwin}

\begin{figure}
\begin{center}
\includegraphics[width=0.5\textwidth]{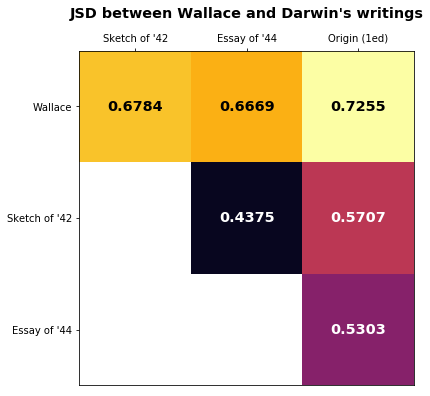}
\end{center}
\caption{\textbf{Similarity between Darwin's writings and Wallace's essay}. This heatmap shows the Jensen-Shannon Distance between Darwin's various drafts, \emph{The Origin of Species}, and Wallace's contemporaneous essay. The Essay of '44 and Sketch of '42 are the closest two texts.}
\label{fig:doc-sims}
\end{figure}

\begin{figure}
\begin{center}
\includegraphics[width=0.5\textwidth]{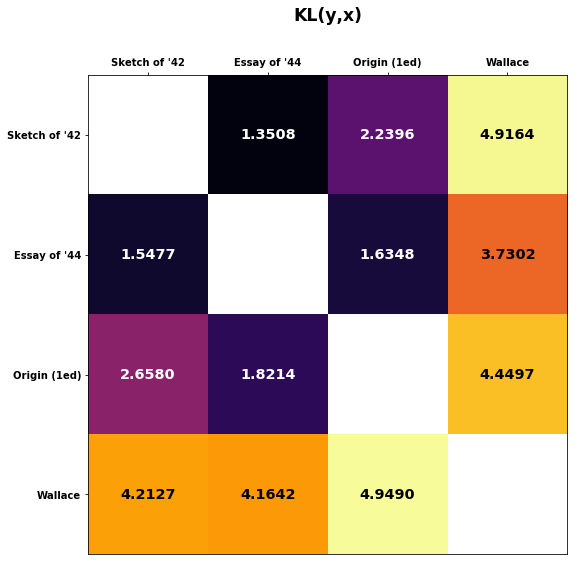}
\end{center}
\caption{\textbf{Divergences between Darwin's writings and Wallace's essay}. This heatmap shows the asymmetric Kullback-Leibler divergence between Darwin's essay drafts, \emph{The Origin of Species}, and Wallace's contemporaneous essay. 
}
\label{fig:doc-divergence}
\end{figure}

Regardless of the primary cause of Darwin's delay, his sudden rush to publication is often attributed to the co-discovery of natural selection by  Wallace, whose own essay ``On the tendency of varieties to depart infinitely from the original type'' was co-published with an excerpt of Darwin's 1844 essay on 30 June 1858 \citep{DarwinWallace1858}.  Darwin had received Wallace's manuscript less than two weeks earlier, on 18 June 1858. At that point, Darwin had already been organizing his notes for \emph{The Origin} for four years. Writing to Lyell, Darwin remarked on the impressive similarity of Wallace's essay to his earlier work:
\begin{quote}
I never saw a more striking coincidence. If Wallace had my M.S. sketch written out in 1842 he could not have made a better short abstract! \citep{DCP2285}
\end{quote}
We take Darwin's remark as both praising Wallace's work and emphasizing how much further his own ideas had developed by 1858. Darwin's observation indicated not just similarity between their ideas, but a \emph{specific} similarity to his 1842 description of natural selection. Applying the JSD measure, the models partially capture Darwin's observation: Wallace's work is more similar to the 1842 and 1844 essays than to \emph{The Origin}. However, it is marginally more similar to the 1844 essay. Darwin's mention of the 1842 sketch may be interpreted as a generic reference to the earlier period, or may reflect that Darwin was not letting on just how far Wallace had come.

To investigate further, we examine the asymmetries in the KL divergences among these texts (Figure~\ref{fig:doc-divergence}). We compare two scenarios: reading Wallace's text after encountering Darwin's writings (last row, figure~\ref{fig:doc-divergence}) and reading Darwin's text after encountering Wallace's manuscript (last column, figure~\ref{fig:doc-divergence}). Wallace's text encloses the Sketch of 1842, but by the time of the 1844 essay, Darwin had already integrated many of the topics that would be later encountered in Wallace's essay, and the KL asymmetry between Darwin and Wallace has reversed direction from the earlier works:  Darwin's writing had begun to enclose Wallace's. The notion of enclosure thus provides another way to interpret Darwin's claim that Wallace had written a near perfect abstract of his 1842 essay: abstracts should enclose their targets. Although the symmetrical JSD metric indicates that Wallace's essay is closest to Darwin's 1844 essay, the asymmetrical KL measure indicates that by 1844, Darwin had already begun to go beyond the range of topics covered by Wallace; thus Wallace's essay ought not be regarded as an abstract of his 1844 work.

The final comparison between Wallace's essay and \emph{The Origin} highlights how much farther Darwin had extended the theory's scope. Wallace immediately recognized the accomplishment and was  generous with his praise for Darwin,  writing to Bates in December 1860:

\begin{quote}
I know not how, or to whom, to express fully my admiration of Darwin’s book. To him it would seem  flattery, to others self-praise; but I do honestly believe that with however much patience I had worked and experimented on the subject, I could never have approached the completeness of his book, its vast accumulation of evidence, its overwhelming argument, and its admirable tone and spirit. \citep{Wallace-Bates1860}
\end{quote}

\section{Conclusion}
Charles Darwin, despite the enormity of his contributions to biology, was generous in his willingness to credit others as sources of ideas, but also clear-minded concerning how far further he had managed to take the arguments than others \citep{Merton1957}. His evidence base was both directly derived from his experiences aboard \emph{HMS Beagle} and indirectly derived from his voluminous reading of the works of others. The works he read provided a source of both theoretical ideas and observations supporting his theory, but they also presented challenges that he knew he must address. As \citet{Lennox2018} emphasizes, Darwin signals his concern about ``a crowd of difficulties ... so grave that to this day I can never reflect on them without being staggered'' \citep{Origin1e}.  Lennox argues that \emph{The Origin} consequently has an unusual structure: After outlining the theory in the first four chapters, Darwin then uses the  middle chapters of the book to deal with this crowd of difficulties, before turning to present the positive evidence for the theory.

In this chapter we have shown how a computational model of Darwin's non-fiction readings can be used to provide evidence concerning the development of the written version of his theory, particularly with regards to the lengthy time between the early 1840s and 1859.  Darwin's delay was not just a matter of biding his time until the ideas became acceptable (or he became unafraid of any possible damage to his reputation). Our model helps confirm that the argument for natural selection was not merely awaiting an opportune moment for publication after writing the 1844 draft: Darwin's reading both during and after the barnacle period had measurable impact on what he wrote. Additionally, the books being published by others reduced the gap between Darwin's conception of natural selection and culture-writ-large. While his receipt of Wallace's essay emphasized the urgency of completing \emph{The Origin}, Darwin recognized (and our models help confirm) that Wallace was still well behind him in his thinking---as Wallace himself recognized once he had \emph{The Origin} in hand. Our approach cannot speak directly to Darwin's intent in the delay, but does show the relevance of Darwin's activity to his work on species.

The previous chapter established a model for investigating exploration-exploitation behavior while avoiding any goal-directed behavior. This chapter introduced the notion of out-of-sample documents for an information space. These writings could be seen as driving an information foraging goal. We performed an analysis comparing the reading behavior over time to the topic distributions of the writings. This showed that Darwin's overall reading trajectory reduced the divergence from his writings. This was observed even though the reading model had no conception of the goal-directed behavior. If we did not see the divergence decrease, skepticism about the fitness of the topic model as a representation of the information environment would have been warranted. Instead, these findings validate our choices for operationalizing information foraging theory.

Finally, we hope to have shown the application of an information retrieval and information foraging framework to intellectual history opens up new questions and does more than demonstrate the obvious, thus establishing that the application of computational methods has much to contribute to the history of science.
\chapter{Extension: Darwin's Later Reading and Writing}
\label{chapter:extensions}
\label{sec:darwin-extension}

In chapters \ref{chapter:readings} and \ref{chapter:writings}, we examined an idealized dataset: Charles Darwin's readings and writings. His reading notebooks provided day-level granularity of what he was reading, and even included the adjacent possible of ``books to be read.'' His writings captured various iterations of his theory of natural selection and modification by descent at various stages of maturity, and were also explicitly dated. However, not all thinkers record their scientific process so meticulously. This chapter explores an extension of our model using less-precise data.

Darwin's reading notebooks stopped in 1860, but he did not stop reading or acquiring new books. The collection of books he owned in 1882 has become known as the ``Darwin Library''. The physical collection was donated to the Botany School at Cambridge University by his son Francis, and subsequently cataloged. These physical artifacts have been digitized by the Biodiversity Heritage Library\footnote{https://www.biodiversitylibrary.org/collection/darwinlibrary}. They include Darwin's marginalia - a strong indicator that he engaged with the book. Other indicators of Darwin's readings include references in his publications, borrowing records from Edinburgh University Library and the London Lending Library, and sales catalogs of his elder relatives.

In chapter~\ref{chapter:writings} we discussed Darwin's Delay, the 17-year-gap between his 1842 draft and publication of \emph{The Origin of Species}. While we showed that Darwin used the time to incorporate materials that were materially relevant to \emph{The Origin}, he never publicly published his thoughts on human evolution until 1871 in \emph{The Descent of Man}. Here, we examine Darwin's second delay in publishing on human evolution.


\section{Dataset Curation}
\begin{figure}
    \centering
    \includegraphics[width=0.75\textwidth]{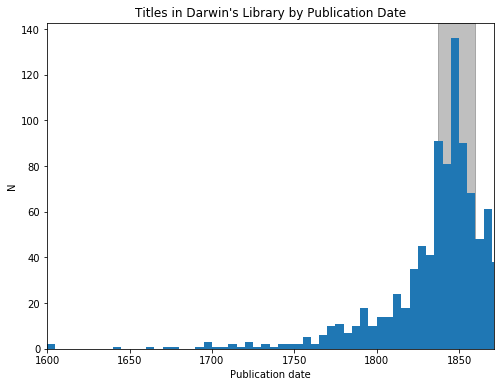}
    \caption{{\bf The number of pre-1871 books in Darwin's Library by publication date.} The shaded region indicates the time period when Darwin maintained his reading notebooks. That most of the books in the Library come from this time period indicates the significance of it to Darwin's scholarship.}
    \label{fig:publication-dates}
\end{figure}

First, I digitized the Rutherford Catalog of the Darwin Library into a parsed bibliography. Then, I filtered the records to only those before 1871 (1011 records). This inclusive criteria captures many volumes would have been available to Darwin, had he sought them at the time of publication, even though he may have acquired them later. I further filtered these records with the same selection criteria as in the previous two chapters: English-language non-fiction, excluding journals (430 volumes). Of these volumes, 155 were already in the reading notebooks collection. I located 265 new volumes for inclusion in the dataset, for a total of 912 volumes (see Figure~\ref{fig:publication-dates} for the distribution of publication dates).

The \texttt{Library-1871} corpus was stoplisted with the NLTK English-language stoplist, plus frequently-occurring terms until 75\% of the corpus was removed, and low-frequency terms until 5\% of the corpus was removed. This matched the preparation of the original \texttt{Reading-Notebook} corpus. I trained LDA topic models at $k=60$ and $k=200$ topics for 500 iterations, matching the studies in chapters \ref{chapter:readings} and \ref{chapter:writings}, respectively.

\section{Validation}
\begin{figure}
    \centering
    \includegraphics[width=\textwidth]{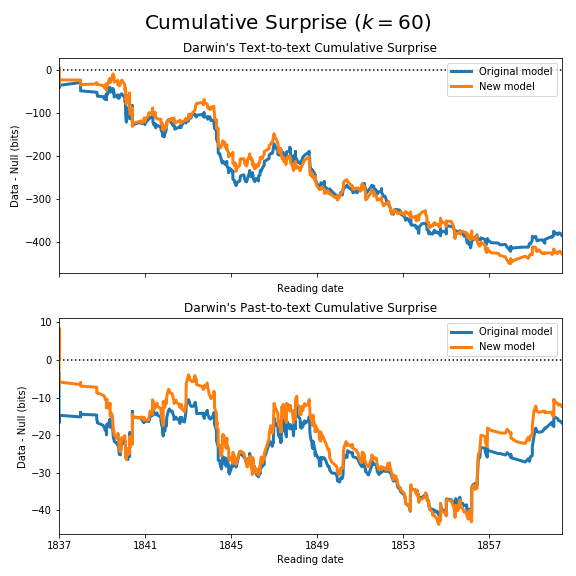}
    \caption{{\bf Validation of Darwin's reading notebook experiments with new topic models}. The blue lines indicate the model from Figure~\ref{fig:kl_over_time}. The orange lines indicate the exact same readings but with topic distributions drawn from a new model trained on the original corpus of 646 texts, plus an additional 130 volumes from the Darwin Library. Note that the original model and new model trend in largely the same directions. The BEE also finds similar epoch breaks.}
    \label{fig:validation}
\end{figure}

To validate the new models, I reran the text-to-text and past-to-text measures from the reading notebook study using the subsection of \texttt{Library-1871} shared by both corpora. The results are in Figure~\ref{fig:validation}. Both measures show similar trends to the findings in Figures~\ref{fig:kl_over_time} and\ref{fig:culturalsurprise}.

\begin{figure}
    \centering
    \includegraphics[width=\textwidth]{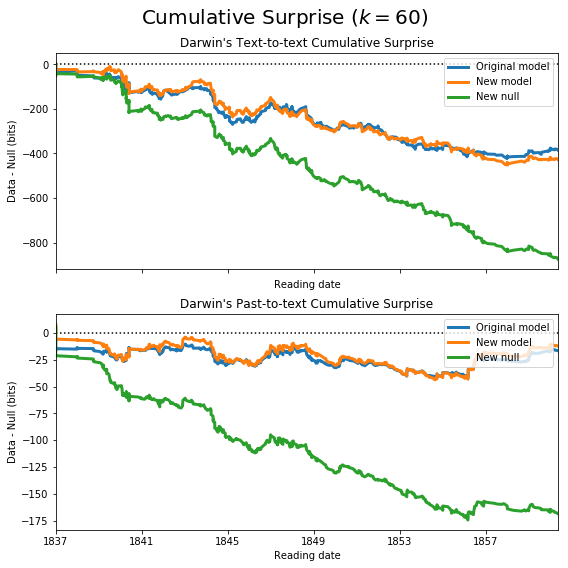}
    \caption{{\bf Verification of null model using additional volumes}. The new null (bottom line) includes an additional 130 volumes. It indicates that the additional volumes further removed correlations between random volumes, i.e., the additional volumes demonstrated a stronger correlation among Darwin's readings.}
    \label{fig:new-null}
\end{figure}

As the new model contains an additional 130 volumes published before publication of \emph{The Origin of Speices}, we revise the null comparison for his readings by adding these additional volumes (see Figure~\ref{fig:new-null}). This revised null shows that the cumulative surprise of what Darwin chose to read was even less surprising than under the original null---\emph{i.e.}, the books Darwin chose to read were more correlated than just picking a random book off the shelf, even with a larger shelf. It also indicates that the additional books in the Darwin Library further diversified his knowledge base. If the new books had instead duplicated existing knowledge, we could have seen a \emph{weaker} null, as each book at random would be more likely to be correlated.

The epoch analysis with the new model and new null reveals a shift in both division points. The first division moves from May 1846 to December 1847, while the second moves from September 1854 to August 1854. In the first case, this pushes the start of the behavioral change from prior to starting the barnacles project in October 1846 to a full-year into the project. In the second case, this shows a shift in reading just prior to starting to organize his notes, as opposed to just after.

\section{\emph{The Descent of Man}}
In Chapter~\ref{chapter:readings}, we showed that during the 17 year gap between the first draft and publication of \emph{The Origin of Species} Darwin read books that were materially relevant to his work. This was shown by demonstrating that the readings-to-writing KL divergence decreased over time. If he was reading works unrelated to \emph{The Origin}, this measure would have increased. 

While we do not have an order for the books read after May 1860, we can compare the divergence between the \emph{Descent} and the topic distribution of the 647 volumes in our reading notebook corpus and between the \emph{Descent} and the topic distribution of the 265 volumes in Darwin's Library that were not in the notebooks.  Since he did not read these volumes prior to the end of the reading notebooks, he must have read them after (if at all). 

\begin{figure}
    \centering
    \includegraphics[width=\textwidth]{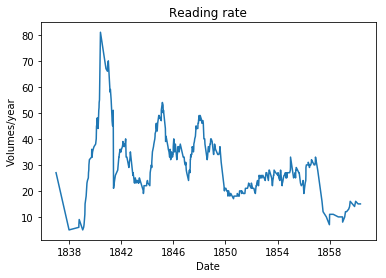}
    \caption{{\bf Darwin's reading rate}. The number of books Darwin read in a given 1-year period. The average rate was 33 books per year before he started writing the \emph{The Origin of Species}, but slowed down dramatically after that point to 15 books per year.}
    \label{fig:reading-rate}
\end{figure}

The comparison to the end of the notebooks and the new acquisitions allows us to sew Darwin's yearly reading rate had slowed to 15 volumes/year by the end of the notebook period (see Figure~\ref{fig:reading-rate}), but until beginning work on the \emph{Origin}, he read an average of 33 books/year. It is possible then that he could have consumed all of these new volumes in his library in the 11 intervening years between \emph{The Origin} and \emph{Descent of Man}.

To get a set of topic distributions for \emph{Descent of Man}, I ran 1,000 query samples (see Section~\ref{sec:query-sampling}) under the $k=200$ model over the Project Gutenberg edition of the text. For each sample, I took the KL divergences to each corpus and reported the averages. Table~\ref{table:descent-divergences} shows the divergences between the \emph{Descent of Man} and three collections: the books he reported read in the reading notebooks, the pre-1871 Darwin Library volumes not in the reading notebooks, and the combined collection. \emph{Descent} is closest to the volumes not in the reading notebooks, and closer to all readings than to those in the reading notebooks. This indicates that while Darwin's early readings informed his theories on human evolution, the delay was again caused by a search for materially-relevant evidence.

\begin{table}[]
    \centering
    \begin{tabular}{c|c}
        Corpus & KL divergence to \emph{Descent}\\ \hline
        Reading Notebooks & 3.283  \\ 
        Darwin's Library & 2.807  \\
        All Volumes & 2.928
    \end{tabular}
    \caption{{\bf KL divergence between two subcorpora and \emph{The Descent of Man}}. \emph{Descent of Man} is closest to the volumes not in the reading notebooks, and closer to the entire corpus than just the reading notebooks. This indicates that Darwin's continued information foraging (i.e., reading) impacted \emph{The Descent of Man} more than earlier readings.}
    \label{table:descent-divergences}
\end{table}

\section{Revisions of \emph{The Origin of Species}}
\emph{The Origin of Species} had six revisions. In chapters~\ref{chapter:readings} and \ref{chapter:writings}, we focused our studies on the first edition, as it was contemporaneous with the reading notebooks. The second edition was published January 7, 1860, and added material to address religious objections. The third edition was published in April 1861, adding an introductory appendix addressing the controversy surrounding the book. The fourth edition was published in 1866, and the fifth edition in 1869 added the phrase ``survival of the fittest''. The sixth and final edition was published on February 19, 1872, which for all practical purposes means it was compiled in 1871, making our corpus an accurate reflection of the texts which may have influenced the book. For a visualization of the changes in the six editions see \citet{Fry2009}.

For each edition, 100 query samples were trained under the $k=200$ model, with comparisons to the average of the KL between the text and the various corpora made in the same manner as for \emph{Descent}. Rather than comparing to all texts in the corpus for each volume, I compare to all books in the corpus published at the time of the revisions publication. The results are summarized in Table~\ref{table:origin}.

\begin{table}[]
    \centering
    \begin{tabular}{lc|ccc}
        Edition & Publication & All at publication & Notebooks & Library \\ \hline
        Origin (1e) & 1859-11-24 & 3.855 & 3.825 & 2.817 \\
        Origin (2e) & 1860-01-07 & 3.706 & 3.771 & 2.793 \\
        Origin (3e) & 1861-04-00 & 3.769 & 3.849 & 2.781 \\
        Origin (4e) & 1866-00-00 & 3.593 & 3.829 & 2.750 \\
        Origin (5e) & 1869-02-10 & 3.390 & 3.790 & 2.708 \\
        Origin (6e) & 1872-02-19 & 3.123 & 3.665 & 2.585
    \end{tabular}
    \caption{{\bf Comparison of each edition of \emph{The Origin of Species}}. This table shows the divergence from each edition of \emph{The Origin} to the books in Darwin's Library and Reading Notebooks that were published by the time that edition of \emph{The Origin} was published. We also show the divergence to the books referenced in the Notebooks, which is exclusively pre-1860 texts, and to the Library. Across the board, the books in the Library are less divergent than the books in the Reading Notebooks.}
    \label{table:origin}
\end{table}

The query sampling process seems to detect that Darwin's revisions were not minor changes, but rather substantive contributions that incorporated new material. This can be seen in the decreasing divergence to the new Library collection with each successive edition. Curiously, the original notebook corpus has no temporal correlation to the different editions, and the sixth edition is closest to the older documents. Part of this may be due to Darwin's inclusion of an introductory appendix which addresses criticism, which may be found in the earlier documents. Although Darwin is often citing these earlier works negatively, even discussion of them causes his works to share discourse with them. Closer reading is necessary to determine the roots of this phenomenon.

\section{Conclusion}
This extension of the case study in chapters \ref{chapter:readings} and \ref{chapter:writings} shows how materials without a reading date can be incorporated with a reading corpus to hypothesize about the relationship of the texts an author read to an author's publications. We also showed how revisions can cause a text to come closer to the discourse of previously-published works as criticism is addressed.

As a whole, the Darwin case study with this extension provides descriptive measures and models of an individual's reading and writing behavior over the course of 34 years. However, it does not provide a generative model which predicts \emph{what} text is read next, \emph{when} it will be read, or \emph{if} it will be read at all. Additional datasets, such as Darwin's ``Books to be read'' list, could aid in forming a generative model to answer these questions. Information foraging theory revolves around deciding which source of information to allocate attention. Here we show how our results hold in expanded information environments. Developing a generative model would enable us to predict what books were read next and fill the gaps between 1860 and 1871 with a text-to-text comparison. Further discussion of a generative model is found in Chapter~\ref{chapter:future}: Future Work. 

\chapter{Preliminary Work: Thomas Jefferson's Correspondence and Retirement Library}
\label{chapter:jefferson}

In this chapter, I present a new case study with very different metadata from the Darwin case study. Our aim here is to show how a study could be designed that incorporates personal library holdings, which are only guaranteed to have publication dates and possibly an acquisition date, alongside correspondence.

Thomas Jefferson was a crucial figure of the American Enlightenment. As his tombstone reads: ``Author of the Declaration of American Independence, of the Statute of Virginia for Religious Freedom, and Father of the University of Virginia'', his impact on American thought is indisputable.  In addition to these roles, he served as minister to France, first Secretary of State, second Vice President of the United States, and third President of the United States. Jefferson established the Library of Congress during his presidency, and later sold his personal library to Congress after the first Library of Congress was burned by the British in the War of 1812.

\begin{figure}
    \centering
    \includegraphics[width=.75\textwidth]{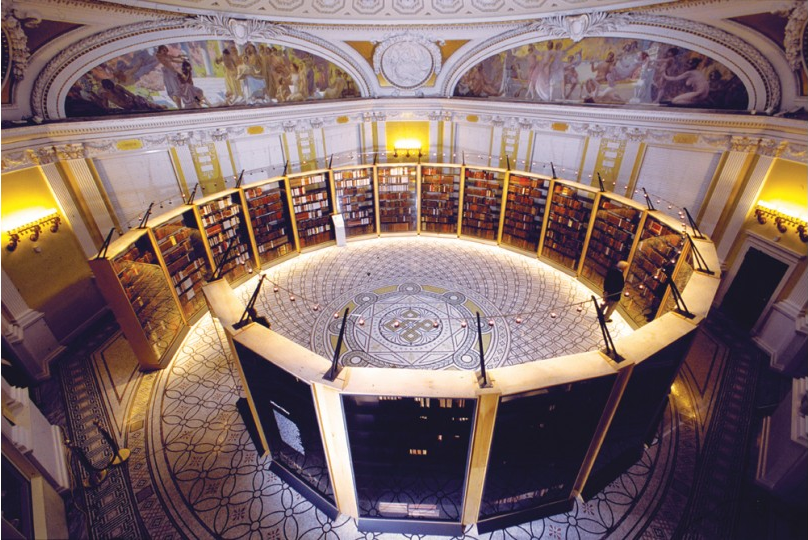}
    \caption{\textbf{Reproduction of Jefferson's Library at the Library of Congress.} On August 24, 1814, British troops burned Washington D.C., including the United States Capitol, which housed the Library of Congress. Thomas Jefferson's personal library was sold to Congress for \$23,950 in 1815 to rebuild the collection. That library is shown here, reconstructed in its original catalog order and using items from the original sale when possible.  Photo: Library of Congress.}
    \label{fig:my_label}
\end{figure}{}

Jefferson was also meticulous about preserving his correspondence, inventing a ``polygraph'' machine which used two quills to write a copy of a letter simultaneously with the first. The most famous correspondence was that of Jefferson and John Adams, the second President of the United States. The two men had become estranged by partisan politics in the elections of 1796 and 1800, but reconciled in 1816, beginning a correspondence reflecting on the origins of their nation. This correspondence that lasted until they died on the same day -- July 4, 1826, the 50th anniversary of the Declaration of Independence. Adams's last words were ``Jefferson lives,'' which was in fact false, as Jefferson died 8 hours before Adams.

Hypocritically, the author of ``that all men are created equal'' was a slaveholder. He was an avowed white supremacist, arguing that black Americans should be deported, and while fathering his own mixed-race children with his slave Sally Hemmings \citep{foster_jefferson_1998}, he was a firm opponent of miscegenation. Despite these contradictions, Jefferson was often said to be ``ahead of his time'' in many regards: religious tolerance, meteorology, public education, and exploration. Even on subjects of slavery, Jefferson's own words are said to be ``light-years ahead of [his time]'' \cite{coates_thomas_2012}, clearly articulating the moral wrongs of the practice.

In this extension, we examine his correspondence in the context of his retirement library. This study shows how the text-to-text and past-to-text measures can be used with non-reading data to answer historical questions about how an individual interacted with their culture. We compare topic distributions of each letter in Jefferson's correspondence to the cumulative topic distribution of all books at each publication date to determine at what point in the published record Jefferson's letters are closest. This measure does not allow us to place Jefferson ``light-years'' ahead of his time, as he could not have bought a book for his retirement library published after his death in 1826, it does, however, allow us to place him in the context of his own time. By finding his letters were closer to books published after they were written, we can show how he shaped his culture. 

\section{Dataset Curation}
\begin{figure}
    \centering
    \includegraphics[width=\textwidth]{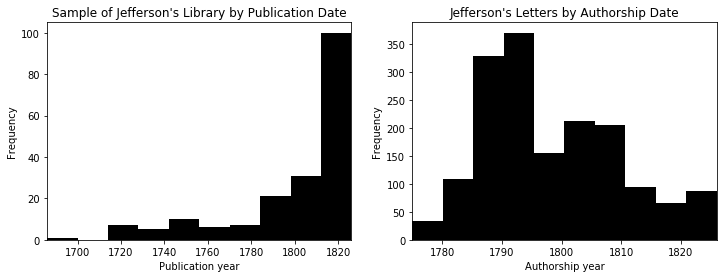}
    \caption{\textbf{Temporal distribution of Jefferson's Retirement Library and Letters}. The left side shows the distribution of publication years in Jefferson's Retirement Library. Note that most books were published during his retirement, \emph{i.e.}, after 1809. The right side shows the distribution of authorship years in our sample of 1,731 letters. Note that there are fewer letters in this sample from retirement.}
    \label{fig:jefferson-dates}
\end{figure}

With the assistance of several undergraduate students, we curated two corpora. The Retirement Library consists of 413 volumes representing 189 bibliographic records (as in Darwin's collection, many books were published as multiple volumes in the 18th and 19th centuries). The Letters of Thomas Jefferson corpus consists of 1,465 letters from Jefferson and 266 letters to Jefferson (1,731 total letters). These were collected from The Library of Congress's \emph{Thomas Jefferson Papers, 1606 to 1827} collection\footnote{\url{https://www.loc.gov/collections/thomas-jefferson-papers/about-this-collection/}} and \emph{The Memoirs, Correspondence, and Miscellanies, from the Papers Of Thomas Jefferson}, a posthumous 1829 volume edited by his grandson, Thomas Jefferson Randolph. The publication and authorship distribution of these books and letters is shown in Figure~\ref{fig:jefferson-dates}.

The corpus was stoplisted with the NLTK English-language stoplist, plus frequently-occurring terms until 65\% of the corpus was removed, and low-frequency terms until 5\% of the corpus was removed. I trained LDA topic models at $k=60$ and $k=200$ topics for 500 iterations.

\section{``Ahead of his time''}
Jefferson was often said to be ``ahead of his time,'' but to what degree was this true? We use his retirement library as a proxy for the broader culture Jefferson was interacting with, while his letters are a proxy for his own contributions to that culture. Figure~\ref{fig:jefferson-dates} shows that Jefferson continued to at least purchase new works, with his retirement library consisting largely of texts purchased in his retirement.

\begin{figure}
    \centering
    \includegraphics[width=.75\textwidth]{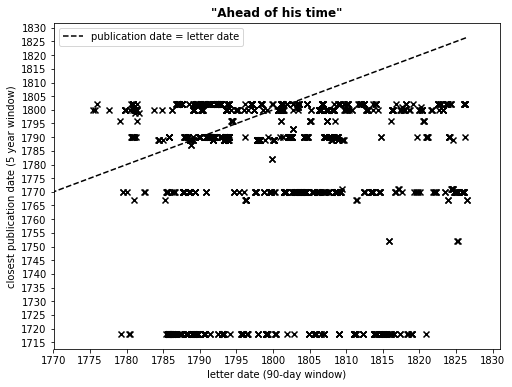}
    \caption{\textbf{Date of least-divergent 5-year publication window to 90-day correspondence window.} We find that by and large, Jefferson was closest to the volumes published in 1802. An expanded corpus of books going past Jefferson's death may be more interesting for our question of whether he was ``ahead of his time.''}
    \label{fig:jefferson-time-analysis}
\end{figure}

We analyzed the KL divergence of a sliding 90-day window of Jefferson's letters to 5-year publication windows of books in his library. For each window, we select the lowest KL-divergence publication window to indicate ``the time'' Jefferson was ``of'' at any given moment. These results are shown in Figure~\ref{fig:jefferson-time-analysis}. Until Jefferson's first presidency, he was closer to books that were yet-to-be-published, \emph{i.e.}, above the dashed diagonal line. After that point, the closest publication date falls below the date of authorship, despite his purchasing of more books published in recent years.

One account of this data is that until Jefferson became President, he was, in fact, ahead of his time. After having his opportunity to leave a direct mark on the executive, the rest of his retirement was spent defending this legacy in letters, often returning to earlier texts. This would match what we see in the revisions of \emph{The Origin} in the previous chapter. Successive editions were closer to the earlier texts because Darwin was directly addressing criticisms raised in them. There is some biographical evidence for this as well, as the Jefferson-Adams letters reflcted upon America's founding and their respective, divergent, presidencies.

\section{Future Work}
One open question is whether the BEE model would find biographical correlates in the automatically-detected epoch divisions of Jefferson's letters. In particular, we would expect the following time periods to be revealed:
\begin{enumerate}
    \item 1775-1784 --- Revolutionary
    \item 1784-1800 --- Statesman
    \item 1801-1809 --- President
    \item 1809-1826 --- Retirement
\end{enumerate}

\noindent These four periods represent the major times reflected in the letters and clear breaks in tone of the conversations. Jefferson the Diplomat wrote in a different modality than Jefferson the President.

Generating a reading model to cover Jefferson's library catalogs is more challenging than extending Darwin's reading lists. For Darwin, we have partial reading data from his reading notebooks. Jefferson's correspondence does not have such day-by-day records. However, he did maintain purchase records in his \emph{Commonplace Book} \citep{wilson2014jefferson}. These purchase records could give a lower bound for when the soonest he could have read a particular volume and help induce a reading history from this data.

Furthermore, there are many challenges with using a model containing both letters and books in the same corpus. LDA is not adept at handling such wide variance in document lengths, especially with a flat prior, as discussed in Chapter~\ref{chapter:topic-modeling}. A future improvement with these variable-length documents could be introducing an asymmetric prior. Training a model on only the book data, then query sampling the letter data may produce better results as well.

Our exploration of whether Jefferson was ``ahead of his time'' largely revealed that Jefferson was a man of 1802, during his first term in office. This may show a different mode of foraging behavior. While Jefferson's writing was closer to books that were yet-to-be-published until his Presidency, the similarity of his output to earlier works may reflect the final stage of \citet{Berger-Tal2014}'s model of information foraging: information exploitation. Early efforts at examining Darwin's retirement library showed similar trends of getting closer to earlier works, as shown in the previous chapter.

\blankpage
\chapter{Preliminary Work: Citation Networks}

In this extension to our model, we induce reading histories from a citation database to generalize the case studies of Darwin and Jefferson to a population-level analysis. This is done by inducing reading histories from a large citation database---The Clarivate Web of Science. We use the citation network to collect papers written by a particular author or authorship team, and infer what they read from their citation behavior.

Modern scientific citation practice was established by the time Darwin did his scholarship. The advent of citation databases, however, had not~\citep{Garfield1979}. I make some simplifying assumptions about citation practice in order to leverage citation databases for this study. First, I assume that a citation indicates that an article has been read. This assumption may be rather spurious, as \citet{simkin_read_2002} found that nearly 80\% of citations are ``unread,'' measured by the persistence of misprints (errors) in the citation across multiple papers. This assumption greatly simplifies our exploratory analysis without necessitating a model of citation motivations \citep{erikson_taxonomy_2014}. Multi-author papers also confound this citation---which author read a particular citation? We narrow our initial case study to neuroscience, a discipline in which the publication norms presume the first author to have done the majority of the writing (and presumably, the reading), while the last author is the principal investigator of the project. We track first-last author dyads as a single author in our study.

Neuroscience is a particularly interesting area to do a citation analysis on because citation meta-analyses are becoming acceptable in the domain to provide not only domain-relevant information retrieval, but can make novel discoveries in neuroscience. For example, by studying semantic networks \citet{Richardet2015} were able to extract brain connectivity from the neuroscience literature. This type of technique is particularly useful for cognitive neuroscience, where researchers must frequently cross the ``translation problem'' in interdisciplinary research  \citep{Demarest2015}, bridging the gap between cognitive systems research and the biological sciences, often bringing two disjoint literatures together. The current state of the art is \textit{NeuroSynth}~\citep{Yarkoni2011}. It uses text mining and image processing together to extract correlations between anatomical and functional data. The current version of the software provides automated meta-analysis of 3,107 terms across 11,406 brain studies. An addition to NeuroSynth enabled correlation with specific alleles, giving a genetic component as well \citep{Fox2014}. A recent addition to NeuroSynth used topic modeling to identify regions of brain activity using MNI brain-coordinate space \citep{rubin_decoding_2017}. \citet{Alhazmi2018} built upon \emph{NeuroSynth} to find correlations in the literature that go beyond shared vocabulary.

Citations formally emerged in the 19th century with the advent of the \emph{Proceedings of the Royal Society} and other scientific institutions. While scientists have almost always made references, prior practice was to attribute individuals rather than precisely dated texts. This is because scientific practice was largely based on personal communications prior to the 1800s. This was seen in Darwin's work and documented through the Darwin Correspondence Project. The previous section's work relating Jefferson's library to his letters tackles these earlier attribution issues more thoroughly. However, early papers did not have the same quantity of citations as they do now. In 1910, a sample of articles only found 1.5 dated citations per article. After this, the quantity of citations dramatically increased  \citet{bazerman_shaping_1988}.

\section{Dataset Curation}
To perform an analysis of discipline-wide foraging patterns, we induced a reading list for each author based on the articles they cited in their publications. To determine the relevant authors, we queried the Web of Science for 23 journals ranked in the top 30 of the \emph{Scimago Neuroscience Rankings} (see Figure~\ref{table:journals}) \footnote{\url{https://www.scimagojr.com/journalrank.php?category=2801}}. We purposefully excluded papers in mega-journals, such as \emph{Nature} and \emph{Science} (see explanation in Section \ref{sec:exclude-nature}). The 23 journals we included contained in 173,682 articles in the 2016 Web of Science release. Of these, 158,359 articles contained citations to 1,681,364 articles. 128,198 of those cited articles were other articles in the sample. See Table~\ref{table:sample-description} for more details on the sample.

\begin{table}[b]
\begin{center}
\begin{tabular}{llr}
\textbf{ISSN} & \textbf{Journal} & \textbf{\# articles} \\ \hline
0006-8950 & Brain & 9,850 \\
0166-2236 & Trends in Neurosciences & 4,851 \\
0898-929X & Journal of Cognitive Neuroscience & 7,865 \\
0022-3077 & Journal of neurophysiology & 19,666 \\
0301-0511 & Biological Psychology & 4,163 \\
0896-0267 & Brain topography & 825 \\
0256-7040 & Child's nervous system & 6,221\\
0270-6474 & The Journal of Neuroscience & 34,348 \\
0896-6273 & Neuron & 9,936 \\
0010-9452 & Cortex & 3,812\\
0953-816X & EJN, European Journal of Neuroscience & 17,830 \\
0014-4819 & Experimental Brain Research & 14,737 \\
0028-3932 & Neuropsychologia & 8,407 \\
0147-006X & Annual Review of Neuroscience & 829 \\
1047-3211 & Cerebral Cortex & 4,367 \\
1053-8119 & Neuroimage & 14,785 \\
1065-9471 & Human brain mapping & 3,246 \\
1097-6256 & Nature Neuroscience & 5,375 \\
1471-003X & Nature Reviews Neuroscience & 1,229 \\
1758-8928 & Cognitive Neuroscience & 286 \\
1863-2653 & Brain Structure \& function & 1,046 \\  \hline
& \textbf{Total} & 173,682
\end{tabular}
\end{center}
\caption{\textbf{Neuroscience journals used in citation study.} We selected 23 journals based on the top 30 of the \emph{Scimago Neuroscience Journals} ranking, excluding mega-journals, such as \emph{Nature} and \emph{Science}. These journals contained 173,682 articles in the Web of Science.}
\label{table:journals}
\end{table}

\begin{table}[h!]
\begin{center}
\begin{tabular}{ll}
\# first-last author dyads & 106,726 \\ \hline 
\# articles (all) & 173,682 \\
\# articles w/citations & 158,359 \\ 
\# citations & 7,680,675 \\
\# citations (within sample) & 2,124,097 \\
\# cited articles (all) & 1,691,364 \\ 
\# cited articles (within sample) & 128,198 \\
\end{tabular}
\end{center}

\caption{\textbf{Description of sample used in neuroscience citation study.} This article counts show the number of abstracts used for the study. The citation counts show that each article, on average, cites 48.50 articles. Each article is cited, on average, 4.54 times. Articles from the sample are cited by other articles in the sample an average of 16.56 times, indicating that neuroscientists tend to cite other neuroscientists at a higher rate. This fact is not surprising, but serves as an important validation of the selection.}
\label{table:sample-description}
\end{table}

After obtaining the citation dataset, we derived the list of unique first author individuals and first-last author dyads from the articles. For each dyad and first-author individual, we inferred a reading history by noting the date of the first publication that cited an article. This became the reading date for that article.

\section{Dataset Limitations}
Several of the modeling decisions above introduced limtations to our dataset. The most severe of which are sample completeness and author disambiguation, discussed below.

\subsection{Sample Completeness}
\label{sec:exclude-nature}
Mega-journals, such as \emph{Nature}, \emph{Science}, the \emph{Proceedings of the National Academy of Sciences (PNAS)}, and \emph{PLoS Computational Biology}, may include material far outside the realm of neuroscience, but nonetheless publish high-quality, significant, and even seminal papers in neuroscience. Those journals are not included in our sample because of the large quantity of extraneous material we would receive. Coming up with a way to use keywords to filter the abstracts to neuroscientifically-relevant articles would improve the quality of our dataset. An alternative to keywords may be to construct the core neuroscience-journal dataset and then include all articles in these high-prestige mega-journals which were cited by or which cite articles in our journal list. A third option would be to use topic models to assess the domain-relevance of articles from non-core journals  \citep{Chuang2013}.

We also may not be scraping all relevant journals - we barely scrape psychology or cognitive science with this list. The UCSD Map of Science contains a listing of 1,198 journals in the ``Psychiatry and Brain Research'' category \citep{Borner2012}. This may be another seed to include more references in our database. Another way to select journals would be to use the ``Neuroscience'' category in the \emph{Journal Citation Report}\footnote{\url{https://clarivate.com/products/journal-citation-reports/}}.

\subsection{Author Disambiguation}
The Web of Science dataset does not provide unique identifiers for each author to indicate whether an author of the same name is in fact the same person across different articles. Instead, the Web of Science provides authors in the format of ``LastName, FirstInitial'' (ex. ``Murdock, J''). Fortunately, even with such a scarcity of data, initials-based disambiguation has been shown to be 97\% accurate at matching authors together \citep{Milojevic2013}, especially in a discipline-constrained dataset like the neuroscience set we examined. There are state-of-the-art disambiguation schemes that can achieve a 99\% accuracy, but at the cost of simplicity \citep{Torvik2009}. I make no effort in this project to further disambiguate beyond initials.

\section{Methods}
For this study, I trained several topic models with differing numbers of topics over the corpus of 128,129 article abstracts, reporting here for $k=250$. Then, for each authorship dyad, I inferred the papers they read from the citation graph. Since we do not know the order in which papers were read, I assign the ``reading date`` for each cited paper as the publication year of the citing article. Then I calculate the KL divergence from the papers read to articles written for each year. Dates are normalized for each other, such that year 0 represents an author's first publication.

\section{Results}
\begin{figure}
    \centering
    \includegraphics[width=.75\textwidth]{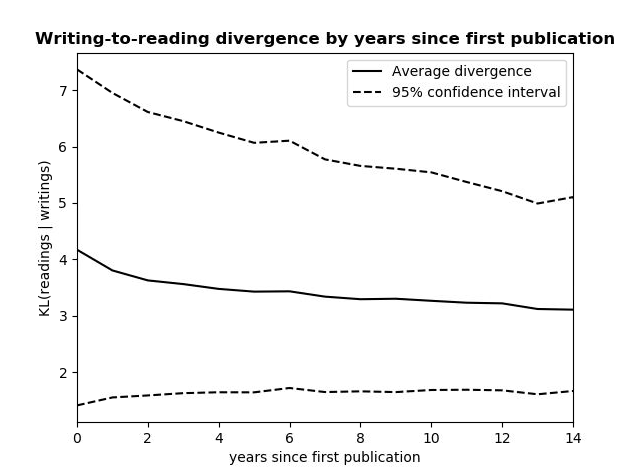}
    \caption{\textbf{Cumulative citation-to-abstract divergence for 106,726 authors.} Over time, the population's divergence between the abstracts an author wrote against the abstracts of articles cited decreases. This suggests a pattern of overall exploitation among published neuroscientists. The lines above and below show the 95\% confidence intervals.}
    \label{fig:cumulative-divergence-hyperbrain}
\end{figure}

 Figure~\ref{fig:cumulative-divergence-hyperbrain} shows the average with the 95\% confidence interval, finding that over the course of an author's career, the KL divergence between abstracts of articles an author wrote and the abstracts of the articles they cited decreases.

\section{Future Work}
This preliminary analysis shows that over the course of the average author's career, the divergence between the abstracts of an author's writings and the abstracts of the papers they cited decreases over time, suggesting that most authors in neuroscience move toward exploitation of existing knowledge throughout their career. Doing an analysis to see if there are different distributions in this pattern, perhaps scientists that undergo different exploration-exploitation cycles, could reveal key insights into the sociology of science. Furthermore, the vast quantity of data could enable the creation of a generative model for citations \citep{Kucuktunc2014}. This could also be seen as a generative model for reading selections, a critical gap in our study of Darwin's reading behavior.

However, there are other experiments on information foraging that should be carried out with this data. The initial results having a wide confidence interval demands more scrutiny. There also is no null model for each author - I do not permute their readings. Creating a null-relative model is crucial for examining these behaviors.

We also abstract away the temporal nature of the data. However, scientists behavior may be temporally correlated. During moments of paradigm shift \citep{Kuhn1970}, authors may have a higher divergence from previous articles as they begin citing a new body of work. During periods of ``normal'' science, we would expect to see continued exploitation by most authors.

The current study does not look at population effects or individual differences in any way. One impact of inducing reading histories is that we can look at a much broader population of information foragers. Perhaps there are foraging differences between highly-cited authors and the rest of the population, leading to more productive scientific output. There also may be changing norms around citation practices. With the advent of citation databases and automated notifications to authors when their articles are cited, citation behavior may be a signal to other authors of the relevance of their work \citep{erikson_taxonomy_2014}. This signal is a clear example of information scent.

The sophistication of the reading date determination could also be improved by examining co-citation data. As an article is cited by more of an article's citations, it is more likely that the article was read after those citing articles were read and pointed the author to it. This posits that co-citation is another type of information scent. Again, changes in technology may change that assumption, as Google Scholar and other databases make it possible to start with an article and see which newer articles cite it.

Finally, while the results here are largely inconclusive and much future work remains, the methodology of inducing reading histories from citation networks provides a gateway for expanded studies of information foraging in a naturally-occurring dataset.
\chapter{Conclusions and Future Work}
\label{chapter:future}

The operationalization of an information environment with topic models, use of naturally-occurring datasets, introduction of new applications for KL divergence, and development of null models of knowledge acquisition (reading) and knowledge creation (writing) all represent significant advances in the state of the art in the logic of discovery, information foraging, and cultural analytics. The case studies of Charles Darwin's readings, writings, and later works demonstrate different aspects of knowledge acquisition and its relationship to knowledge creation. It also shows how datasets with varying degrees of precision can be combined. The notebooks have much more granularity than his retirement library catalog, but the two in tandem are able to extend a 23 year dataset to 34 years, with strong tests of the null models. The preliminary study of Thomas Jefferson's correspondence and retirement library shows how non-book data, such as letters, can be used with this class of models. Additionally, this data has no reading data. The initial results draw attention to modeling concerns when mixing book-length and letter-length texts in the same corpus. Finally, the preliminary work on citation networks and abstracts shows a method for inducing reading histories based on the texts that are cited in an authors works.

In this conclusion, we address several open problems. Some concerns are methodological, such as the identification of influence, behavior of parallelized topic model implementations and the inclusion of multi-lingual corpora. Others are more data expansions: various other datasets of Darwin's scientific career and of general-population readers. Finally, there are a few modeling concerns. Our model of information foraging does not take into account either the diet or the patch model of optimal foraging theory. The reading models, while able to detect behavior shifts, are unable to make predictive assessments of what text should be read next, so we address that by proposing a generative model.

\section{Identifying Influence}
Throughout this dissertation, I have not made claims of influence. Influence is a strong claim: it necessitates that an artifact was encountered, interacted with (\emph{i.e.}, read), and then affected a future action. The problem is proving the causality between one action and the next.  A weak claim of influence is that reading a book and then reading another book in the same area (lowering text-to-text divergence in our model) shows influence of one book on the next reading decision. Stronger claims of influence are that a particular artifact changed the output of a creative act. For example, Darwin's reading of Malthus's essay on population changed his framing of \emph{The Origin}. There is a folk psychology of influence, often seen in literary critiques or music studies. Capitalizing on these notions of influence coluld be a fruitful path for research, and likely the subject of another dissertation.

However, this notion of influence, and the varying degrees of influence claims, has not been well-defined. One way of stepping around influence is to claim that a particular work is ``resonant'' with another \citep{barron2018individuals}. This aspect removes the causal implications of influence. The work presented in chapter~\ref{chapter:writings} on successive text drafts makes an attempt at interpreting the direction of influence between two texts, exploiting the asymmetries of KL divergence to show which text ``encloses'' another.

In complexity science, quantifying causality has a long history. The ``Granger causality'' is a measure of the degree to which one time series analysis can predict another \citep{granger_investigating_1969}. It is based on two principles: 1) The cause happens prior to its effect. 2) The cause has unique information about the future values of its effect \citep{granger_testing_1980}. $X$ ``Granger-causes'' $Y$, if the past values of both $X$ and $Y$ predict the values of $Y$ more than the past values of $Y$ alone. This has led to extensive studies on information theoretic measures quantifying the mutual information of $X$ and $Y$ \citep{hlavackova-schindler_causality_2007,abdul_razak_quantifying_2014,sugihara_detecting_2012}.

There are two philosophical issues involved in discussions of influence: 1) the \emph{post hoc} fallacy and 2) necessary and sufficient conditions. The \emph{post hoc} fallacy is short for the Latin phrase \emph{post hoc ergo propter hoc} (``after this, therefore because of this''). The simple temporal order of one thing before another does not immediately mean it is a causal force for it. For example, the rooster's crow is correlated with the sunrise, but does not cause it. In Granger causality, it is the second principle that addresses this fallacy---the cause has unique information about the future values of its effect. For an overview of philosophical research done on the \emph{post hoc} fallacy, see \citet{pinto_post_2001}. The causality in influence relations also raise an issue of whether a particular influence was necessary or sufficient for the action it was said to influence. In almost all cases, influence is only sufficient, although it is interesting that both Darwin and Wallace found Mathus's essay to be their most significant influence. For an overview of philosophical research on necessary and sufficient conditions, see \citet{brennan_necessary_2017}.

\section{Topic Modeling Concerns}
There are several methodological concerns with topic modeling that remain for future work. I address three below: serial position effects, multi-lingual topic models, and parallelism.

\subsection{Serial Position Effects in LDA}
Rather implicitly, topic models have functioned as a model of semantic memory over the course of a lifetime. While there are some indications that LDA models similar aspects of human judgements on word similarity and polysemy \citep{Griffiths2007}, there are no explorations of how a topic model may reflect the serial position effect. The topic model itself is a bag of words model, so should exhibit no biases with respect to document and word order \citep{Blei2003}. However, query sampling, or other techniques for adding new documents to a previously trained corpus may show either primacy or recency effects. We suspect that models for which the word-topic assignments are \emph{not} allowed to change with training will show  a primacy effect, while models that do allow the word-topic assignments to vary will show a recency effect. Experimentation will help operationalize the use of LDA as a proxy for semantic memory.

\subsection{Multi-lingual Topic Models}
In our studies we focused primarily upon the English-language texts. As the word distribution of non-English texts is often totally disjoint from English texts, the topics representing them were usually entirely disjoint. Due to this, we removed non-English texts from our corpora, preferring to do monolingual analyses. However, Darwin read professionally in at least English, French, and German. One challenge is sharing semantics across languages. A study using topic models over comparable documents from different languages (for example, using the Wikipedia articles from different languages on the same subject) found it possible to generate aligned, multilingual topics \citep{mimno2009polylingual}. \citet{Boyd-Graber2014} were able to find similar results without requiring use of a parallel corpus. Multi-lingualism was a fact of life for scientists in the 18th century, so coming up with models that reflect this is crucial for a complete historical understanding.

\subsection{Parallelization of Topic Models}
Our parallel implementation of topic modeling splits the corpus into $N$ subcorpora of approximately equal size. Each subcorpora has its own Gibbs sampler. This technique is known, formally, as ``hogwild Gibbs sampling'' \citep{johnson_analyzing_2013}. This algorithm is essentially a Map-Reduce operation \citep{newman_distributed_2008}. Each sampler operates independently (map) and then results are coordinated at the end (reduce). The results of hogwild Gibbs estimates results within a standard deviation of a sequential algorithm for all variables in a graphical model \citep{daskalakis_hogwild!-gibbs_2018}. However, the trade-off between time saved from a large number of independently-operating Gibbs samplers versus the accuracy of those samples is relatively unexplored.

\section{Other Darwin Datasets}
In this dissertation we studied Darwin's reading notebooks (and the texts referenced by the notebooks), two drafts of \emph{On the Origin of Species}, six editions of \emph{The Origin}, and \emph{The Descent of Man}. Darwin read and wrote more books than these collections:

The books that Darwin had with him on the Beagle are well-known\footnote{For one edition of the Beagle library see: \url{http://darwin-online.org.uk/BeagleLibrary/Beagle_Library_Introduction.htm}}. These books are of particular relevance for the creation of the 5 volumes in \emph{The Zoology of the Voyage of the Beagle} and the \emph{Journal of researches into the geology and natural history of the various countries visited by H.M.S. Beagle}. These would further the studies in chapters \ref{chapter:readings}, \ref{chapter:writings}, and \ref{chapter:extensions}.

One other shortcoming of our study: we focused upon the non-fiction reading of Darwin. Digital humanists in particular would be interested in a study of the fiction books Darwin read. These volumes were previously identified as part of the reading notebooks study in chapter \ref{chapter:readings}, but not included in the study, as they often were anomalies compared to the topic distributions of the other works.

\section{Other Reading Histories}
Reading records exist not only for elites, like Charles Darwin, but also for the general public. The maintenance of personal reading diaries is widespread, as evidenced by the more than 30,000 records in the UK Reading Experience Database (1450--1945),\footnote{\url{http://www.open.ac.uk/Arts/reading/}} and the 50 million registered users of Goodreads.\footnote{\url{https://www.goodreads.com/about/us} (accessed 2016 July 14)} Further studies will be accelerated by advances in information retrieval techniques to find the full text for each entry in these reading diaries. These large-scale surveys can be complemented by in-depth studies of the ``commonplace books'' left by historical figures. These books record quotes, readings, and interactions that may become useful in their later intellectual life, and include Marcus Aurelius~\citep{marcus}, Francis Bacon~\citep{Bacon1883}, John Locke~\citep{Locke1706}, and Thomas Jefferson~\citep{wilson2014jefferson}.

\section{Patch and Diet Models}
Our models of information foraging theory borrow concepts from optimal foraging theory, but do not utilize the models of OFT.

There are two primary models: the patch model and the diet model. In the patch model, environmental structure determines foraging behavior \citep{Charnov1976}. The diet model weights the energy cost of pursuing different types of resources, each having different energy gains and resource distributions \citep{Stephens1986}. This dissertation leaves negotiation of the patch model and diet model in the information foraging context to future work.

Our models of information foraging did not account for energy costs in any manner, assuming that attention was already managed given our historical datasets. Additionally, we did not examine the sparsity of the overall LDA space. By examining the data sparsity we could easily apply the patch model to see if the semantic geometry of the topic space was affecting reading behavior. As for better incorporating the diet model, there is a clear correlation between document length and amount of time necessary to read it. This budgeting of attention could be incorporated to a future version of this model.

\section{Generative Model}
More generally, the ignorance of the diet and patch models speaks to a bigger hole in our methodology. While the case studies were validated through null model testing and experimental design, we did not then create a generative model for reading decisions. Predicting \emph{what} text is read next, \emph{when} it will be read, or \emph{if} it will be read at all could further advance the study of knowledge acquisition.  Darwin's ``Books to be read'' list \citep{vorzimmer1977}, could aid in forming a generative model to answer these questions. A generative model would enable us to predict what books were read next and fill the gaps in reading history from his retirement library from 1860 to Darwin's death. Furthermore, it would be immensely helpful in inducing reading histories for figures that did not record them, such as Jefferson.

\section{Conclusion}
Even with these limitations, tying together notions of bounded cognition, information foraging, and information theory without embedding any particular goal-directed behavioral assumptions is a huge advance in the state of the art. Our studies on the timescale of years and decades in an individual goes beyond prior work in cognitive science. It is our hope that readers can leap from our first impressions of the open research problems in this section into a fruitful investigation.

\printbibliography[heading=bibintoc,title={References}]

\phantomsection
\cftlocalchange{toc}{400pt}{0cm}
\cftaddtitleline{toc}{chapter}{Curriculum Vita}{}


\section*{Supplementary material}

# Jaimie Murdock, Ph.D.

**E-Mail**  research@jamram.net                    **Homepage**  http://jamram.net/

## EDUCATION

*Doctor of Philosophy*, Informatics and Cognitive Science                    June 2019
Indiana University, Bloomington, IN
Dissertation: Topic Modeling the Reading and Writing Behavior of Information Foragers
Committee: Colin Allen (co-chair), Staša Milojević (co-chair), Michael Jones, Peter Todd

*Bachelor of Science*, Computer Science (with honors)                    December 2010
Indiana University, Bloomington, IN
Concentrations: Artificial Intelligence and Programming Languages
Founder, Computer Science Club

*Bachelor of Science*, Cognitive Science                    December 2010
Indiana University, Bloomington, IN
Co-chair, Midwest Undergratuate Cognitive Science Conference (MUCSC)
President, Student Organization for Cogntive Science (SOCS)

## POSITIONS HELD

*Cornell University, Ithaca, NY (remote)*                    January 2018 – present
Web Applications Developer                    `https://arXiv.org`

*Internet Archive, San Francisco, CA (remote)*                    March – December 2017
Web Crawl Engineer                    `https://archive.org`

*Indiana University, Bloomington, IN*                    August 2010 – August 2012
Visiting Research Associate                    `https://inphoproject.org`

## PUBLICATIONS

### Journal Articles

[3] Jaimie Murdock, Colin Allen, and Simon DeDeo. "Exploration and Exploitation of Victorian Science in Darwin's Reading Notebooks". *Cognition* 159 (2017), pp. 117–126. DOI: `10.1016/j.cognition.2016.11.012`. arXiv: `1509.07175`.

[4] Jared Lorince, Sam Zorowitz, Jaimie Murdock, and Peter Todd. "Wisdom of the Few? "Supertaggers" in Collaborative Tagging Systems". *Journal of Web Science* 1.1 (2015). DOI: `10.1561/106.00000002`. arXiv: `1502.02777`.


## Book Chapters

[5] Jaimie Murdock, Cameron Buckner, and Colin Allen. "Evaluating Dynamic Ontologies". In: *Knowledge Discovery, Knowledge Engineering, and Knowledge Management: Second International Joint Conference, IC3K 2010, Valencia, Spain, October 25-28, 2010, Revised Selected Papers*. Ed. by Ana Fred, Jan L. G. Dietz, Kecheng Liu, and Joaquim Filipe. Vol. 272. Communications in Computer and Information Science. Springer-Verlag, 2013, pp. 258–275. DOI: `10.1007/978-3-642-29764-9_18`.

[6] Jaimie Murdock. "Computational Discovery". In: *The Dynamics of Science: Computational Frontiers in History and Philosophy of Science*. Ed. by Grant Ramsay and Andreas De Block. University of Pittsburgh Press, accepted.

[7] Jaimie Murdock, Colin Allen, and Simon DeDeo. "Darwin's Semantic Voyage". In: *The Dynamics of Science: Computational Frontiers in History and Philosophy of Science*. Ed. by Grant Ramsay and Andreas De Block. University of Pittsburgh Press, accepted.


## Conference Papers

[8] Jaimie Murdock, Colin Allen, and Simon DeDeo. "Quantitative and Qualitative Approaches to the Development of Darwin's Origin of Species". *Current Research in Digital History* 1 (2018). DOI: `10.31835/crdh.2018.14`. arXiv: `1802.09944`.

[9] Jaimie Murdock, Jacob Jett, Tim Cole, Yu Ma, J. Stephen Downie, and Beth Plale. "Towards Publishing Secure Capsule-Based Analysis". *2017 ACM/IEEE Joint Conference on Digital Libraries (JCDL)*. Toronto, Ontario, Canada, June 2017. DOI: `10.1109/JCDL.2017.7991585`.

[10] Jared Lorince, Sam Zorowitz, Jaimie Murdock, and Peter Todd. ""Supertagger" behavior in building folksonomies". *WebSci '14: Proceedings of the 2014 ACM Conference on Web Science*. Bloomington, Indiana, USA, June 2014, pp. 129–138. DOI: `10.1145/2615569.2615686`.

[11] Jaimie Murdock, Cameron Buckner, and Colin Allen. "Containing the Semantic Explosion". *Proceedings of the WWW2012 conference workshop PhiloWeb 2012: "Web and Philosophy, Why and What For?"* Ed. by Alexandre Monnin, Harry Halpin, and Leslie Carr. Vol. 859. CEUR Workshop Proceedings. Apr. 2012.

[12] Jaimie Murdock and Colin Allen. "InPhO for All: Why APIs Matter". *Journal of the Chicago Colloquium on Digital Humanities and Computer Science* 1.3 (Nov. 2011).

[13] Jaimie Murdock and Larry S. Yaeger. "Identifying Species by Genetic Clustering". *Advances in Artificial Life: 20th Anniversary Edition - Back to the Origins of Alife, ECAL 2011*. Paris, France: MIT Press, Aug. 2011, pp. 564–572.

[14]  Jaimie Murdock, Cameron Buckner, and Colin Allen. "Two Methods for Evaluating Dynamic Ontologies". *Proceedings of the 2nd International Conference on Knowledge Engineering and Ontology Development (KEOD)*. Ed. by Ana Fred and Joaquim Filipe. Vol. 272. Revised and expanded in [5]. Valencia, Spain: Springer-Verlag, Oct. 2010, pp. 110–122. DOI: `10.5220/0003101601100122`.


## Conference Abstracts

[15]  James Pustejovsky, Marc Verhagen, Keongmin Rim, Yu Ma, Liang Ran, Samitha Liyanage, Jaimie Murdock, Robert H. McDonald, and Beth Plale. "Enhancing Access to Digital Media: The Language Application Grid in the HTRC Data Capsule". *PEARC '17: Proceedings of the Practice and Experience in Advanced Research Computing 2017 on Sustainability, Success and Impact*. Poster. New Orleans, Louisiana, USA, July 2017, 60:1–60:3. DOI: `10.1145/3093338.3104171`.

[16]  Jaimie Murdock and Colin Allen. "Visualization Techniques for Topic Model Checking". *AAAI-15: Proceedings of the Twenty-Ninth AAAI Conference on Artificial Intelligence*. Demo of `https://hypershelf.org`. Jan. 2015, pp. 4284–4285.

[17]  Jaimie Murdock, Jiaan Zeng, and Robert H. McDonald. "Topic Exploration with the HTRC Data Capsule for Non-Consumptive Research". *JCDL '15: Proceedings of the 15th ACM/IEEE-CS Joint Conference on Digital Libraries*. Tutorial. Knoxville, Tennessee, USA, June 2015, p. 295. DOI: `10.1145/2756406.2756929`.

[18]  Timo Sztyler, Jakob Huber, Jan Noessner, Jaimie Murdock, Colin Allen, and Mathias Niepert. "LODE: Linking Digital Humanities Content to the Web of Data". *JCDL '14: Proceedings of the 14th ACM/IEEE-CS Joint Conference on Digital Libraries*. Demo. London, United Kingdom, Sept. 2014, pp. 423–424.

[19]  Jaimie Murdock, Robert Light, Colin Allen, and Katy Börner. "Mapping the Intersection of Science and Philosophy". *JCDL '13, Proceedings of the 13th ACM/IEEE-CS Joint Conference on Digital Libraries*. Poster. Indianapolis, Indiana, USA, July 2013, pp. 405–406. DOI: `10.1145/2467696.2467777`.

[20]  Jaimie Murdock and Larry S. Yaeger. "Genetic Clustering for the Identification of Species". *GECCO '11. Proceedings of the 13th Annual Conference Companion on Genetic and Evolutionary Computation*. Poster. Expanded in [13]. Dublin, Ireland, July 2011, pp. 29–30. DOI: `10.1145/2001858.2001875`.

[21]  Mathias Niepert, Cameron Buckner, Jaimie Murdock, and Colin Allen. "InPhO: A System for Collaboratively Populating and Extending a Dynamic Ontology". *JCDL '08: Proceedings of the 8th ACM/IEEE-CS Joint Conference on Digital Libraries*. Pittsburgh, Pennsylvania, USA, June 2008, p. 429. DOI: `10.1145/1378889.1378976`.


## Miscellaneous

[22]  Jaimie Murdock, Jiaan Zeng, and Colin Allen. *Towards Evaluation of Cultural-scale Claims in Light of Topic Model Sampling Effects*. Advanced Collaborative Support Technical Report. HathiTrust Research Center, June 2016. arXiv: `1512.05004`.

## TEACHING

### Indiana University, Bloomington, IN

*Foundations in Science and Mathematics*
Introduction to Computer Science                                    June 2016
Introduction to Computer Science                                    June 2015

*Department of History and Philosophy of Science and Medicine*
HPSC-X420/X521 Computational Methods for HPS            January - May 2016

*Cognitive Science Program*
COGS-Q260 Programming for the Cognitive and Info. Sciences     January - March 2016
COGS-Q320 Computation in the Cognitive and Info. Sciences        March - May 2016
COGS-Q260 Programming for the Cognitive and Info. Sciences     January - March 2015
COGS-Q320 Computation in the Cognitive and Info. Sciences        March - May 2015

*School of Informatics, Computing, and Engineering*
INFO-I399 Research Methods                                August - December 2013
CSCI-Y390 Independent Study                                 January - May 2013
CSCI-C343 Data Structures                                 August - December 2012
CSCI-C343 Data Structures                                 August - December 2010

## PRESENTATIONS

*Current Research in Digital History 2018*                        March 17, 2018
The Development of Darwin's Origin of Species

*Joint Conference on Digital Libraries 2017*                       June 20, 2017
Towards Publishing Secure Capsule-based Analysis

*HTRC User Group Meeting*                                         March 2, 2017
HTRC Data Capsule - faster, bigger, better

*Philosophy of Science Association 2016 Meeting*                November 4, 2016
Darwin's Semantic Voyage

*Guest lecture in ILS-Z657 Digital Humanities*                  November 3, 2016
From Texts to InPhO (and back!)

*International Conference on Thinking 2016*  August 4, 2016
The Cognitive Science of Darwin's Delay

*HASTAC 2016 Workshop*  May 13, 2016
From Texts to InPhO (and back!)

*HASTAC 2016*  May 12, 2016
Darwin's Semantic Voyage

*Maryland Insitute for Technology in the Humanities (MITH) Staff Talk*  Apirl 19, 2016
Darwin's Semantic Voyage

*IU CCCP Lab Meeting*  April 13, 2016
From Texts to InPhO (and back!)

*IU Intelligent and Interactive Systems Seminar*  March 23, 2016
Darwin's Semantic Voyage

*Pestilli Research Laboratory*  March 4, 2016
Topic Modeling: A Practical Introduction

*IU Speech and Hearing Science Graduate Seminar*  February 26, 2016
Visualizing Scientific Data

*Guest Lecture in HPSC-X521 Digital Methods for HPS*  February 18, 2016
Topic Modeling: A Practical Introduction

*IU CCCP Lab Meeting*  February 10, 2016
Topic Modeling: A Practical Introduction

*HathiTrust Research Center User Group*  November 11, 2015
Charles Darwin, Information Foraging, and Non-consumptive Research

*Student Organization for Cognitive Science*  October 27, 2015
Darwin's Semantic Voyage

*CLingDing: IU Computational Linguistics Discussion Group*  October 23, 2015
Darwin's Semantic Voyage

*2015 Conference on Complex Systems (CCS'15)*  October 2, 2015
Darwin's Semantic Voyage

*Guest Lecture in STAT-S675 Statistical Learning*  September 11, 2015
Exploration and Exploitation of Darwin's Reading Notebooks

*International Center for Jefferson Studies Fellows' Forum*  June 4, 2015
Polymathic and Polygraphic: Thomas Jefferson's Mind

*Interactive Cognitive Systems Open House*  April 17, 2015
Visualization Techniques for LDA

*Goldstone, Landy, Todd Lab*  April 10, 2015
Topic-Driven Foraging

*IU SoIC Informatics PhD Un-poster Session*  April 8, 2015
Darwin's Semantic Voyage

*APA Pacific Division Invited Symposia*  April 4, 2015
Comparing Topic Models of Digital Philosophy Resources

*HathiTrust Research Center Uncamp*  March 30, 2015
Topic Modeling with the HTRC Data Capsule

*Network Science Talks @ IU*                                              March 9, 2015
Topics Over Time: Into Darwin's Mind

*IU Cognitive Computing Lab*                                          November 5, 2014
Making Pretty Pictures w/Topic Models

*HathiTrust Research Center Uncamp*                            September 9, 2013
InPhO and HathiTrust: Digging for Philosophy in the Sciences

*International Association for Computing and Philosophy Annual Meeting*   July 17, 2013
Simulating Expertise through Subcorpus Modeling

*Berkeley Insitute of Design Lunch Series*                          August 7, 2012
A Tale of Two Encyclopedias

*WWW2012 Philoweb Workshop*                                         April 17, 2012
Containing the Semantic Explosion

*IU Computer Science Club Tech Talk*                            November 17, 2011
Practical Parallelism

*IU Student Organization for Cognitive Science*                November 15, 2011
Categorization

*Int. Conf. on Knowledge Engineering and Ontology Development*   October 26, 2010
Two Methods for Evaluating Dynamic Ontologies

## AWARDS

Fellow, International Center for Jefferson Studies (ICJS)                    May 2015
Provost's Award for Undergraduate Research and Creative Activity          April 2010
Cognitive Science Program Outstanding Contribution Award                  April 2009
Hutton Honors College Arthur Metz Scholar                                 April 2007

## CONSULTING

University of Pittsburgh                                    January 2019 - Present
The Internet Philosophy Ontology Project                     `https://inphoproject.org`

University of Victoria                                        April 2015 - August 2017
Linked Modernisms                                            `http://linkedmods.uvic.ca`

Stanford University                                               June - August 2014
Stanford Encyclopedia of Philosophy                        `https://plato.stanford.edu`

University of Alabama                                      October 2010 - August 2011
Syriac Reference Portal                                            `https://syriaca.org`

Logic Pedallers for McGraw-Hill                             October 2008 - May 2009
Power of Logic Web Tutor                                    `http://www.poweroflogic.com`

## MENTORING

*Graduate:* Aishwarya Iyer, Adithya Nagaraj Tirumale, Rohit Zawar

*Undergraduate:* Evan Boggs, Tim Downey, Hannah Eli, Alex Frost, Lauren Gaynor, Bradi Heaberlin, Ram Iyer, Ryan Penelton, Wesley Pettyjohn, Jessie Pusateri, Loran Saggu, Kirtan Sakariya, Jesse Squires, Justin Stamets, Aish Thamba, Samuel Waggoner, Catherine Xu, Sara Zhang, Sam Zorowitz


\bookmark[dest={Jaimie_Murdock_CV.pdf.1}]{Curriculum Vitae}

\end{document}